\documentclass[10pt,twocolumn,letterpaper]{article}

\usepackage[pagenumbers]{cvpr} % To force page numbers, e.g. for an arXiv version

\usepackage{bm}
\usepackage{graphicx}
\usepackage{amsmath}
\usepackage{amssymb}
\usepackage{booktabs}
\usepackage{multirow}
\usepackage[dvipsnames]{xcolor}

\makeatletter
\@namedef{ver@everyshi.sty}{}
\makeatother

\usepackage{tikz}
\usepackage{tcolorbox}

\usepackage{algorithmic}
\usepackage{algorithm}
\usepackage{pifont}
\usepackage{listings}
\usepackage{amsmath,amsfonts,bm}

\def\eqref#1{equation~\ref{#1}}
\def\1{\bm{1}}

\DeclareMathAlphabet{\mathsfit}{\encodingdefault}{\sfdefault}{m}{sl}
\SetMathAlphabet{\mathsfit}{bold}{\encodingdefault}{\sfdefault}{bx}{n}

\definecolor{royalblue}{RGB}{65,105,225}
\definecolor{better}{RGB}{0, 165, 156}
\definecolor{tab_blue}{RGB}{31, 119, 180}
\definecolor{tab_orange}{RGB}{255, 127, 14}
\definecolor{tab_green}{RGB}{44, 160, 44}
\definecolor{tab_red}{RGB}{214, 39, 40}

\usepackage[pagebackref,breaklinks=true,colorlinks,urlcolor=magenta,citecolor=royalblue,bookmarks=false]{hyperref}

\usepackage[capitalize]{cleveref}
\crefname{section}{Sec.}{Secs.}
\Crefname{section}{Section}{Sections}
\Crefname{table}{Table}{Tables}
\crefname{table}{Tab.}{Tabs.}

\def\ourblock{RobustResBlock}
\def\ourscale{RobustScaling}
\def\ournetwork{RobustResNet}

\def\etal{{\it et al.}}

\begin{document}

\title{Revisiting Residual Networks for Adversarial Robustness:\\An Architectural Perspective}

\author{Shihua Huang$^{1}$\quad\quad Zhichao Lu$^{2}$\thanks{\emph{Corresponding author}}\quad\quad Kalyanmoy Deb$^{1}$\quad\quad Vishnu Naresh Boddeti$^{1}$\\
{\normalsize $^{1}$ Michigan State University\quad\quad $^{2}$ Sun Yat-sen University} \\
{\tt\small \{shihuahuang95, luzhichaocn\}@gmail.com\quad\quad \{kdeb, vishnu\}@msu.edu}
}
\maketitle

%%%%%%%%% ABSTRACT
\begin{abstract}
Efforts to improve the adversarial robustness of convolutional neural networks have primarily focused on developing more effective adversarial training methods. 
In contrast, little attention was devoted to analyzing the role of architectural elements (such as topology, depth, and width) on adversarial robustness. 
This paper seeks to bridge this gap and present a holistic study on the impact of architectural design on adversarial robustness. 
We focus on residual networks and consider architecture design at the block level, i.e., topology, kernel size, activation, and normalization, as well as at the network scaling level, i.e., depth and width of each block in the network. 
In both cases, we first derive insights through systematic ablative experiments. Then we design a robust residual block, dubbed \ourblock{}, and a compound scaling rule, dubbed \ourscale{}, to distribute depth and width at the desired FLOP count.
Finally, we combine \ourblock{} and \ourscale{} and present a portfolio of adversarially robust residual networks, \ournetwork{}s, spanning a broad spectrum of model capacities. 
Experimental validation across multiple datasets and adversarial attacks demonstrate that \ournetwork{}s consistently outperform both the standard WRNs and other existing robust architectures, achieving state-of-the-art AutoAttack robust accuracy of 61.1\% without additional data and 63.7\% with 500K external data while being $2\times$ more compact in terms of parameters. Code is available at \url{ https://github.com/zhichao-lu/robust-residual-network}
\end{abstract}
\section{Introduction}

\begin{figure}[ht]
    \begin{subfigure}[b]{0.277\textwidth}
    \centering
    \includegraphics[trim={0, 0, 0, 0}, clip, width=\textwidth]{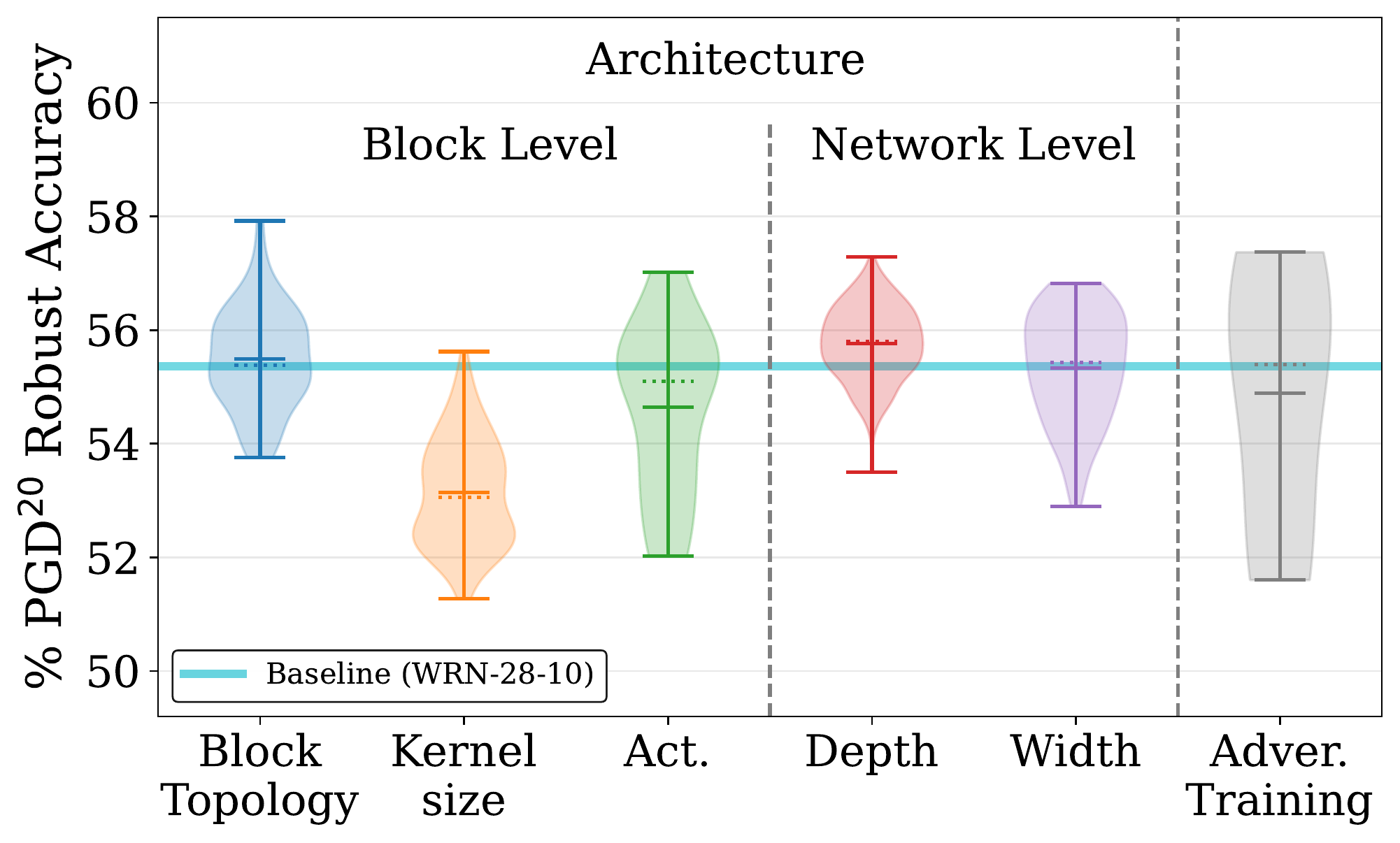}
    \end{subfigure}\hfill
    \begin{subfigure}[b]{0.2\textwidth}
    \centering
    \includegraphics[trim={0, 0, 0, 0}, clip, width=\textwidth]{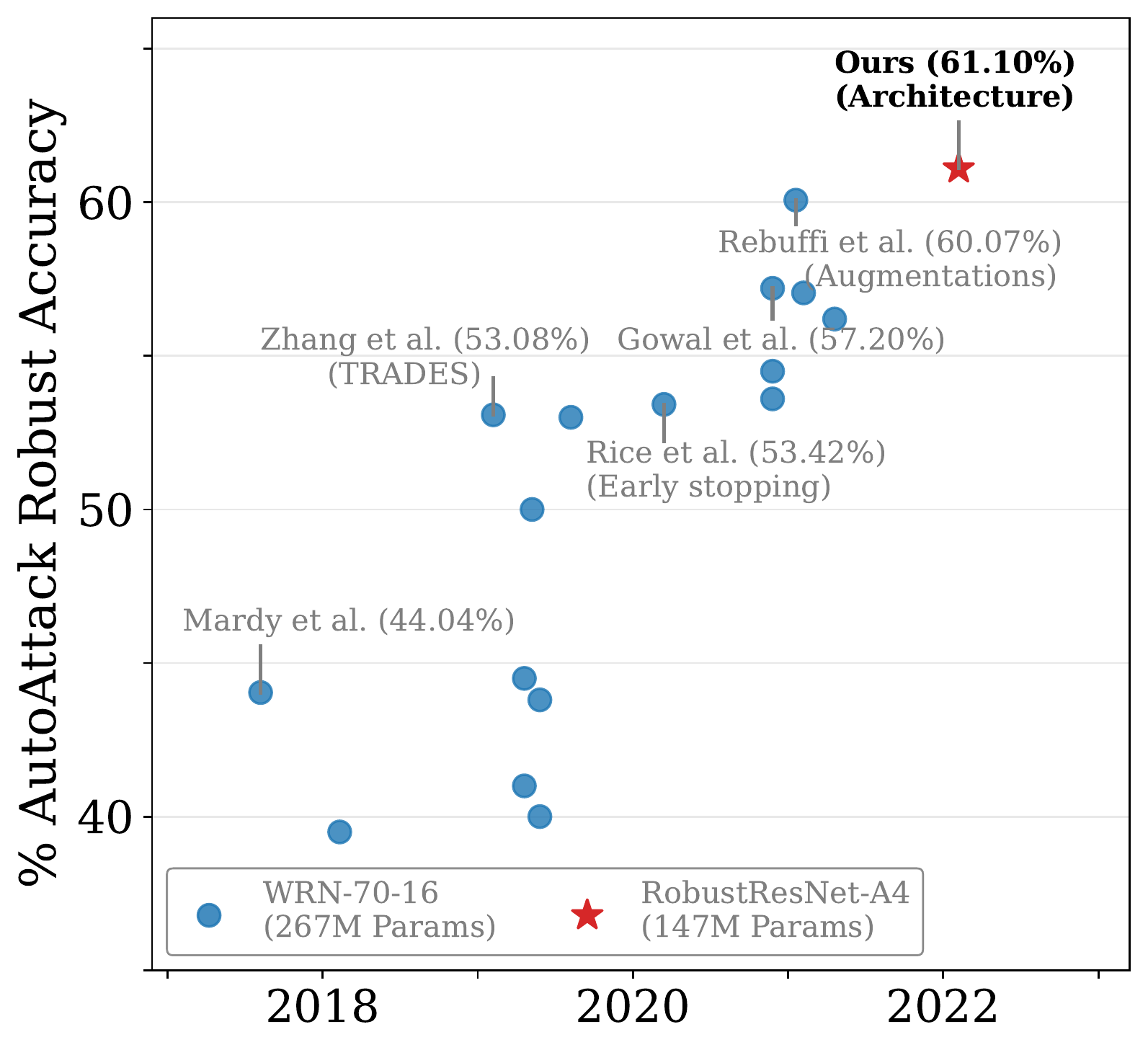}
    \end{subfigure}
    \caption{\textbf{(\emph{L}) Impact of architectural components} on adversarial robustness on CIFAR-10, relative to that of adversarial training methods. The variations of each component are elaborated in \S\ref{sec:method}. \textbf{(\emph{R}) Progress of SotA robust accuracy against AutoAttack} \emph{without additional data} on CIFAR-10 with $\ell_{\infty}$ perturbations of $\epsilon=8/255$ chronologically. We show that innovation in architecture (this paper) can improve SotA robust accuracy while simultaneously being almost $2\times$ more compact. Zoom in for details.\label{fig:teaser}}
    \vspace{-0.5cm}
\end{figure}

Robustness to adversarial attacks is critical for practical deployments of deep neural networks. Current research on defenses against such attacks has primarily focused on developing better adversarial training (AT) methods~\cite{madry2018towards, zhang2019theoretically, wang2019improving, shafahi2019adversarial, wong2020fast}. These techniques and the insights derived from them have primarily been developed by fixing the architecture of the network, typically variants of wide residual networks (WRNs)~\cite{zagoruyko2016wide}. While a significant body of knowledge exists on designing effective neural networks for vision tasks under standard empirical risk minimization (ERM) training, i.e., traditional learning without inner optimization needed in AT, limited attention has been devoted to studying the role of architectural components on adversarial robustness. However, as we preview in Figure~\ref{fig:teaser}, architectural components can impact adversarial robustness as much as, if not more than, different AT methods. As such, there is a large void in practitioners' toolboxes for designing architectures with better adversarial robustness properties.

The primary goal of this paper is to bridge this knowledge gap by (i) \emph{systematically studying the contribution of architectural components to adversarial robustness}, (ii) \emph{identify critical design choices that aid adversarial robustness}, and (iii) finally \emph{construct a new adversarially robust network that can serve as a baseline and test bed for studying adversarial robustness}. We adopt an empirical approach and conduct an extensive amount of carefully designed experiments to realize this goal.

We start from the well-founded observation that networks with residual connections exhibit more robustness to adversarial attacks \cite{cazenavette2021architectural}, and thus, consider the family of \emph{residual networks}. Then we systematically assess the two main aspects of architecture design, block structure and network scaling, and \emph{adversarially train and evaluate more than 1200 networks}. For \emph{block structure}, we consider the choice of layers, connections among layers, types of residual connections, activation, etc. For \emph{network scaling}, we consider the width, depth, and interplay between them. 
To ensure the generality of the experimental observations, we evaluate them on three different datasets and against four adversarial attacks. 
To ensure the reliability of the empirical observations, we repeat each experiment multiple times with different random seeds.
Based on our empirical observations, we identify architectural design principles that significantly improve the adversarial robustness of residual networks.
Specifically, we make the following new observations:
\begin{enumerate}
    \setlength\itemsep{-0.1em}
    \item[\ding{182}] Placing activation functions before convolutional layers (i.e., pre-activation) is, in general, more beneficial with adversarial training, as opposed to post-activation used in standard ERM training. And sometimes, it can critically affect block structures such as the \emph{basic block} used in WRNs. (\S\ref{sec:block_topology}, Figure~\ref{fig:abl_topology_c10_basic} - \ref{fig:abl_topology_c10_inverted})
    
    \item[\ding{183}] Bottleneck block improves adversarial robustness over the de-facto basic block used in WRNs. In addition, both aggregated and hierarchical convolutions derived under standard ERM training lead to improvements under adversarial training. (\S\ref{sec:block_topology}, Figure~\ref{fig:abl_topology_c10_cw40} and \ref{fig:abl_aggre_hier_c10}).
    
    \item[\ding{184}] A straightforward application of SE~\cite{senet} degrades adversarial robustness. Note that this is unlike in standard ERM training, where SE consistently improves performance across most vision tasks when incorporated into residual networks  (\S\ref{sec:block_topology}, Figure~\ref{fig:abl_se}). 

    \item[\ding{186}] The performance of smooth activation functions is critically dependent on adversarial training (AT) settings and datasets. In particular, removing BN affine parameters from weight decay is crucial for the effectiveness of smooth activation functions under AT. (\S\ref{sec:act_norm})
    
    \item[\ding{185}] Under the same FLOPs, \emph{deep and narrow} residual networks are adversarially more robust than \emph{wide and shallow} networks. Specifically, the optimal ratio between depth and width is $7:3$. (\S\ref{sec:compund_scale})
    
    \item[\ding{187}] In summary, architectural design contributes significantly to adversarial robustness, particularly the block topology and network scaling factors.
\end{enumerate}

\noindent Building upon the insights above, we make the following contributions:

\begin{itemize}
\setlength\itemsep{-0.1em}
\item We propose a simple yet effective SE variant, dubbed \emph{residual SE}, for adversarial training. Empirically, we demonstrate that it leads to consistent improvements in the adversarial robustness of residual networks across multiple datasets, attacks, and model capacities.

\item We propose \emph{\ourblock{}}, a novel residual block topology for adversarial robustness. It consistently outperforms the de-facto basic block in WRNs by $\sim3\%$ robust accuracy across multiple datasets, attacks, and model capacities.

\item We present \emph{\ourscale{}}, the first compound scaling rule to efficiently scale both network depth and width for adversarial robustness. Technically, \ourscale{} can scale any architecture (e.g., ResNets, VGGs, DenseNets, etc.). Experimentally, we demonstrate that \ourscale{} is highly effective in scaling WRNs, where the scaled models yield consistent $\sim2\%$ improvements on robust accuracy while being $\sim2\times$ more compact in terms of learnable parameters over standard WRNs (e.g., WRN-28-10, WRNs-70-16).

\item We present a new family of residual networks, dubbed \emph{\ournetwork{}s}, achieving state-of-the-art AutoAttack \cite{croce2020reliable} robust accuracy of 61.1\% without generated or external data and 63.7\% with 500K external data while being $2\times$ more compact in terms of parameters.
\end{itemize}
% -------------------------------------------------------------------------------------------
\begin{figure*}[t]
    \centering
    \begin{subfigure}[b]{0.845\textwidth}
    \centering
    \includegraphics[width=\textwidth]{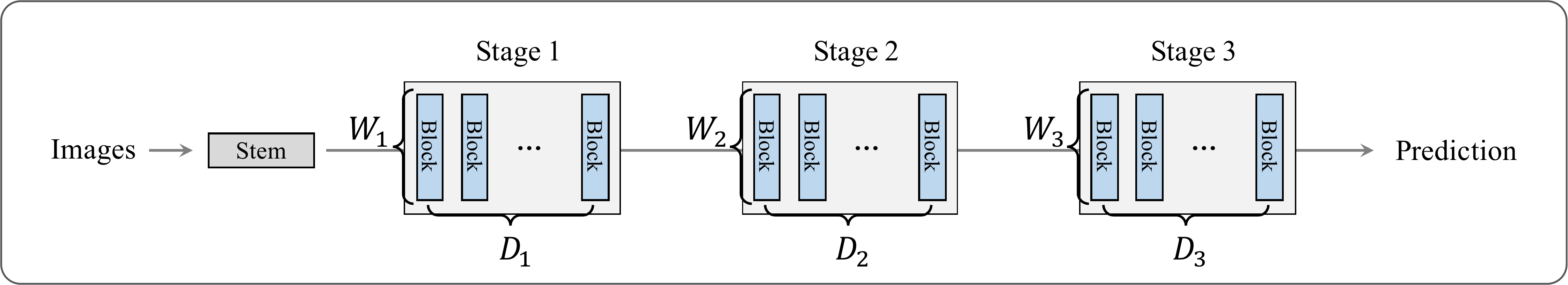}
    \end{subfigure}
    \begin{subfigure}[b]{0.85\textwidth}
    \centering
    \includegraphics[width=\textwidth]{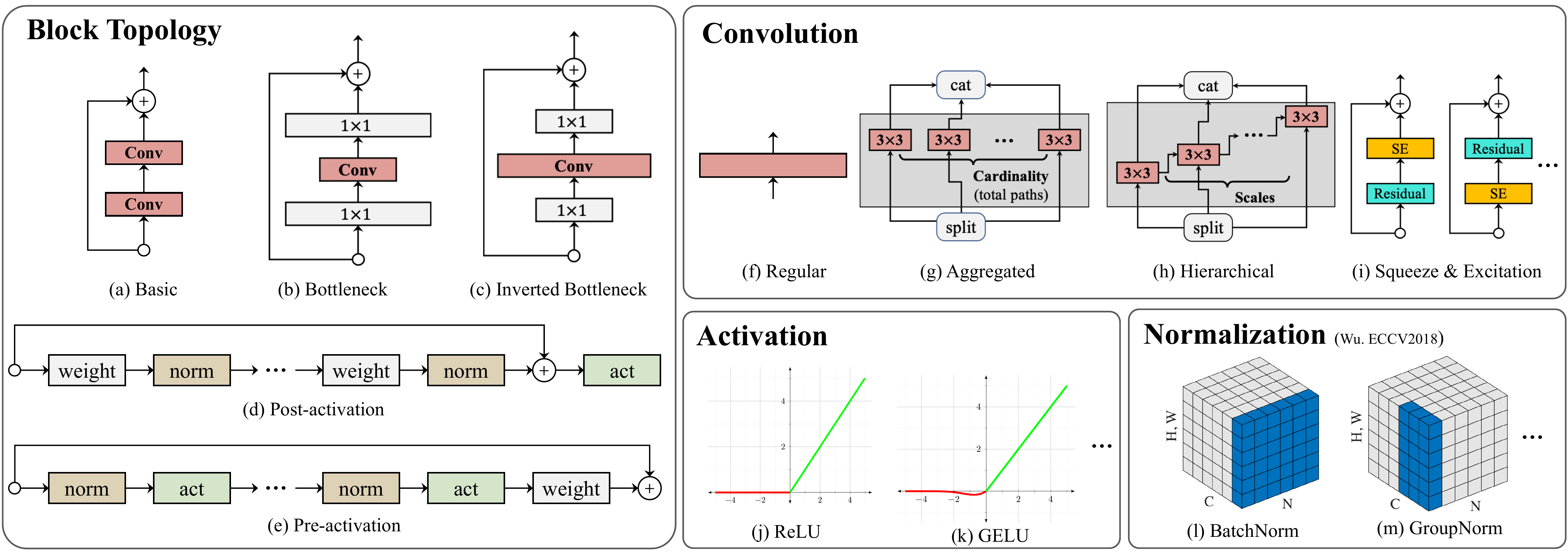}
    \end{subfigure}
    \vspace{-10pt}
    \caption{Overview of the architecture components we considered for adversarial robustness: at the network scaling level (\emph{Top}), the network has three stages, each with multiple blocks controlled by scaling parameters, i.e., depth and width; at the block level (\emph{Bottom}), we explore variants of residual blocks and their components including convolution, activation, kernel size, normalization, etc.\label{fig:overview}}
    \vspace{-12pt}
\end{figure*}
% -------------------------------------------------------------------------------------------

\section{Background and Related Work\label{sec:related_work}}
This section provides a brief overview of related work. An extended description of the relation to existing work is provided in Appendix \S\ref{sec:app_related_work}. 

\vspace{2pt}
\noindent\textbf{Adversarial white-box attacks.} Since the first demonstration that high-performant DNNs are vulnerable to small perturbations in inputs (a.k.a. adversarial examples) \cite{szegedy2013intriguing}, a plethora of efforts have been devoted to crafting stronger adversarial examples (AEs) -- fast gradient sign method (FGSM) \cite{goodfellow2014explaining} is one of the earliest methods that applies a single gradient step to generate AEs; projected gradient descent (PGD) \cite{madry2018towards} is a widely studied method that performs well in most cases while being computationally efficient; Carlini \& Wagner (CW) \cite{carlini2017towards} introduced an alternative loss that exhibits strong attack performance; AutoAttack (AA) \cite{croce2020reliable} is an aggregated attack formed from an ensemble of four complementary attacks.

\vspace{2pt}
\noindent\textbf{Adversarial Training as a Defense.} Adversarial training (AT) has emerged as one of the most effective ways to guard against adversarial attacks. The basic idea of AT is to leverage AEs during the training process of a DNN model. Early work on AT~\cite{madry2018towards} used inputs perturbed by PGD for training. Since then, AT techniques have been extended in multiple directions -- customized loss functions to balance the trade-off between natural and robust accuracy \cite{zhang2019theoretically} or making use of misclassified natural examples \cite{Wang2020Improving}; advanced AT procedures such as early stopping to prevent \emph{robust overfitting} \cite{rice2020overfitting} and weight ensembling \cite{izmailov2018averaging,chen2021robust,wang2022selfensemble}; more diverse data for training by generative modeling \cite{gowal2021improving,sehwag2022robust} or data augmentation \cite{rebuffi2021data}. 

\vspace{2pt}
\noindent\textbf{Robust Architecture.} A few attempts have been made to explore the impact of architectural components on adversarial robustness. From a block structure point of view, (1) Cazenavette \etal{} showed that residual connections significantly aid adversarial robustness~\cite{cazenavette2021architectural}; (2) Xie \etal{} showed that smooth activation functions lead to better adversarial robustness on ImageNet \cite{xie2020smooth}, with a similar observation by Pang \etal{} on CIFAR-10 with ResNet-18 \cite{pang2020bag}; (3) Dai \etal{} identified that parameterized activation functions have better robustness properties \cite{dai2022parameterizing}. However, neither of these studies verified their corresponding observations across different model capacities and datasets.

From a network's scaling factors point of view, the prevailing convention favors wide networks, i.e., using WRNs instead of ResNets (RNs) \cite{zhang2019theoretically,wang2021convergence}. However, we argue that there is no clear consensus on the impact and optimal configurations of scaling factors for adversarial robustness. More specifically, (1) Xie \etal{} hinted that compound scaling with a simple strategy would produce a more robust model than scaling up a single dimension \cite{xie2020smooth}; (2) Gowal \etal{} found that deeper models perform better \cite{gowal2020uncovering}; (3) Huang \etal{} studied the impact of network scaling factors and showed that reducing the capacity of the last stage leads to better adversarial robustness \cite{huang2021exploring}; (4) Mok \etal{} claimed that there is no clear relationship between the width and the depth of architecture and its robustness \cite{mok2021advrush}; (5) Zhu \etal{} showed that width helps robustness in the over-parameterized regime, but depth can help only under certain initialization \cite{zhu2022robustness}. However, none of these studies provided a way to \emph{simultaneously} scale depth and width.

To summarize, unlike this paper, none of the aforementioned prior works holistically study the impact of architectural components, i.e., block structure \emph{and} network scaling, on adversarial robustness.
\section{Preliminaries\label{sec:preliminaries}}

In this section, we describe the experimental setup in terms of the adopted architectural skeleton and the details on training and evaluating the networks against adversarial attacks.

\iffalse
{\color{red}
\begin{table}[ht]
    \centering
    \begin{tabular}{c|c}
        Standard ERM Training &  \\
        Standard Adversarial Training & \\
        Advanced Adversarial Training & \\
    \end{tabular}
    \caption{Caption}
    \label{tab:my_label}
\end{table}
}
\fi

\vspace{2pt}
\noindent\textbf{Architecture Skeleton:} Figure~\ref{fig:overview} shows the skeleton of the network that we consider. It comprises a stem (i.e., a single $3\times3$ convolution) and three processing stages. Each stage is made up of a varying number of convolutional blocks. The first block in stages two and three uses a stride of two to downsample the feature sizes by half. We denote the depth (i.e., number of blocks) and width (in terms of widening factors) of $i\textit{-}th$ stage by $D_{i}$ and $W_{i}$, respectively. Unless otherwise specified, we use $3\times3$ convolution, ReLU activation, and batch normalization as the default operations. We study the effect of the block structure (variants of residual blocks) and the network scaling (configurations of [$D_{1}, D_{2}, D_{3}$] and [$W_{1}, W_{2}, W_{3}$]) on the network's adversarial robustness, within this architectural skeleton.

\vspace{2pt}
\noindent\textbf{Datasets:} We evaluate adversarial robustness on three datasets, CIFAR-10, CIFAR-100 and Tiny-ImageNet.

\vspace{2pt}
\noindent\textbf{Training:} We employ two training strategies in this paper: 
\begin{itemize}
\setlength\itemsep{-0.2em}
    \item \underline{Baseline adversarial training (BAT)}:
    For \emph{outer minimization}, following prior work, models are adversarially trained for 100 epochs with a batch size of 128 using SGD with an initial learning rate of 0.1 (which is multiplied by 0.1 at 75$^{\rm{th}}$ and 90$^{\rm{th}}$ epochs), momentum 0.9, and weight decay $2\times10^{-4}$. We consider three different adversarial losses, SAT \cite{madry2018towards}, TRADES~\cite{zhang2019theoretically} ($\gamma=6$), and MART~\cite{wang2019improving} ($\lambda=5$). For \emph{inner maximization}, we use 10 and 7 steps of PGD with a step-size of $\alpha=2/255$ for CIFAR-10/-100 and Tiny-ImageNet, respectively. The maximum perturbation strength is set to $\epsilon=8/255$ to constrain the $\ell_{\infty}$-norm. \emph{Unless otherwise specified, TRADES is our default baseline}.
    \item \underline{Advanced adversarial training (AAT)}:  
    For \emph{outer minimization}, following \cite{gowal2020uncovering,rebuffi2021data,gowal2021improving}, models are adversarially trained with TRADES~\cite{zhang2019theoretically} for 400 epochs using SGD with Nesterov momentum and weight averaging \cite{izmailov2018averaging}. We use a batch size of 512 and an initial learning rate of 0.2, which is decayed by a factor of 10 two-thirds-of-the-way through training. The decay rate of weight averaging is 0.999. 
    For \emph{inner maximization}, we follow the same procedure as in the case of baseline AT except for the step-size $\alpha$, which is set to 0.1 and decreased to 0.01 after five steps. 
\end{itemize}

\vspace{2pt}
\noindent\textbf{Evaluation:}
We consider multiple attacks for evaluating adversarial robustness including, FGSM~\cite{goodfellow2014explaining}, 20-step PGD (PGD$^{20}$)~\cite{madry2018towards}, 40-step CW (CW$^{40}$)~\cite{carlini2017towards}, and AutoAttack (AA)~\cite{croce2020reliable} with the same perturbation constraint $\epsilon=8/255$. We repeat each experiment multiple times and compute the mean performance {to account for noise in evaluating adversarial attacks}. In all results, we show the mean and standard deviation across the repetitions using markers and shaded regions, respectively.
\section{Design of Adversarially Robust ResNets \label{sec:method}}
We decompose and study the architectural design of adversarially robust residual networks at the block (i.e., block topology and components) and network (i.e., scaling factors such as depth and width) level.

\subsection{Impact of Block-level Design \label{sec:block}}
Designing a block involves choosing its topology, type of convolution, activation and normalization, and kernel size. We examine these elements independently through controlled experiments and, based on our observations, propose a novel residual block, dubbed \emph{\ourblock{}}. A preview of \ourblock{} is provided below.
% -------------------------------------------------------------------------------------
\begin{figure}
    \begin{tcolorbox}[title=Summary of our Robust Residual Block]
        \noindent Building upon the empirical evidence from \S\ref{sec:block_topology} - \S\ref{sec:act_norm}, we propose a new residual block design, dubbed \emph{\ourblock{}}, to substitute the basic block in architectures designed for adversarial robustness.
        \begin{tikzpicture}
            \node[text width=13cm] (a) {
            \noindent\emph{-- Block Topology:} Bottleneck block with\\
            pre-activation, hierarchically aggregated \\
            convolution, and our residual SE (\S\ref{sec:block_topology}).
            
            \vspace{2pt}
            \noindent\emph{-- Activation:} ReLU (\S\ref{sec:act_norm}).
            
            \vspace{2pt}
            \noindent\emph{-- Normalization:} BatchNorm (Appendix \S\ref{sec:app_se}).};
            \node[below of=a, node distance=3.3cm] (z) {};
            \node[left of=z, node distance=3.0cm] (b) {\includegraphics[width=0.95\textwidth]{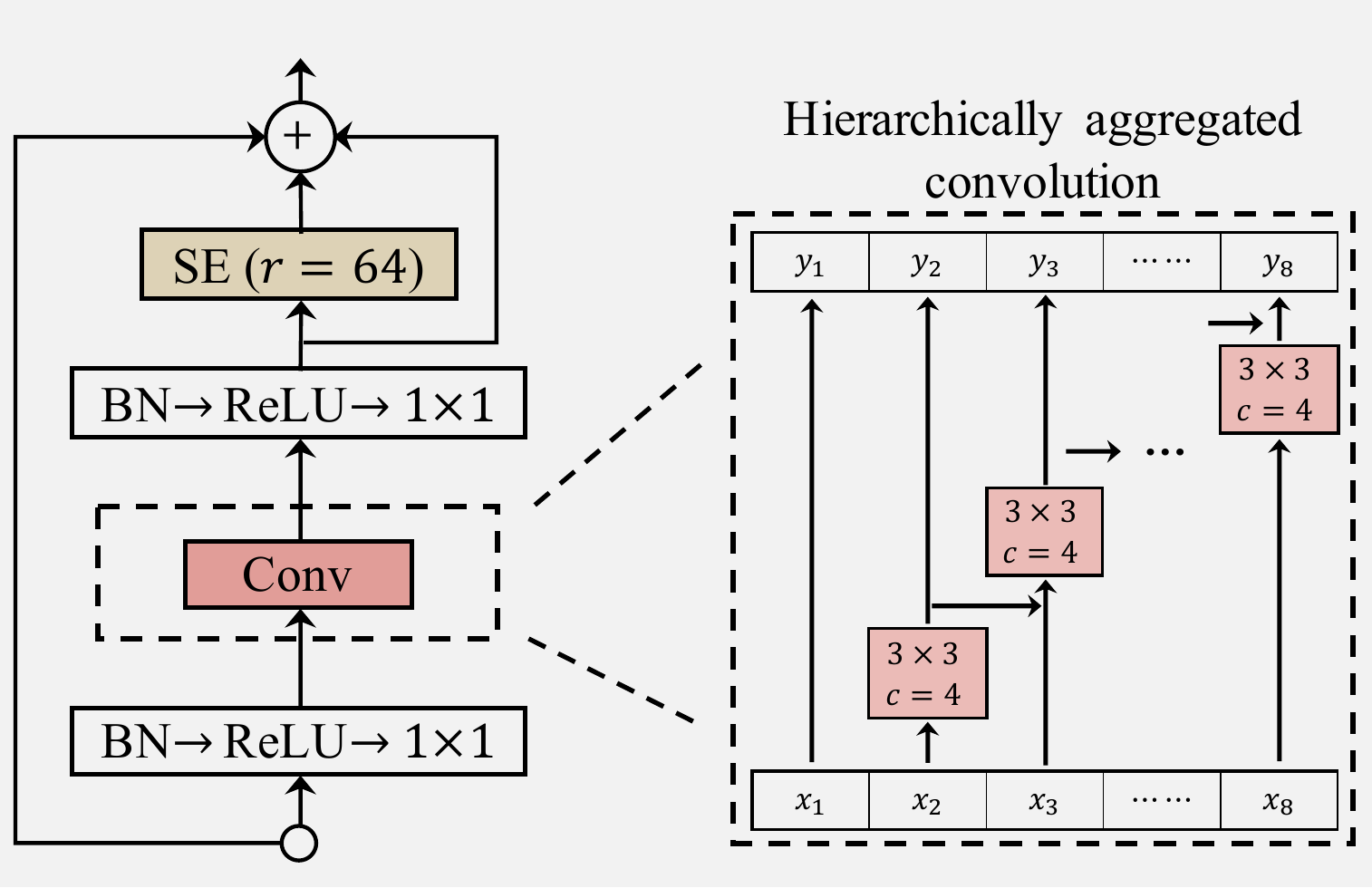}};
            \node[below of=z, node distance=4.6cm] (x) {};
            \node[left of=x, node distance=3.0cm] (c) {\includegraphics[width=0.95\textwidth]{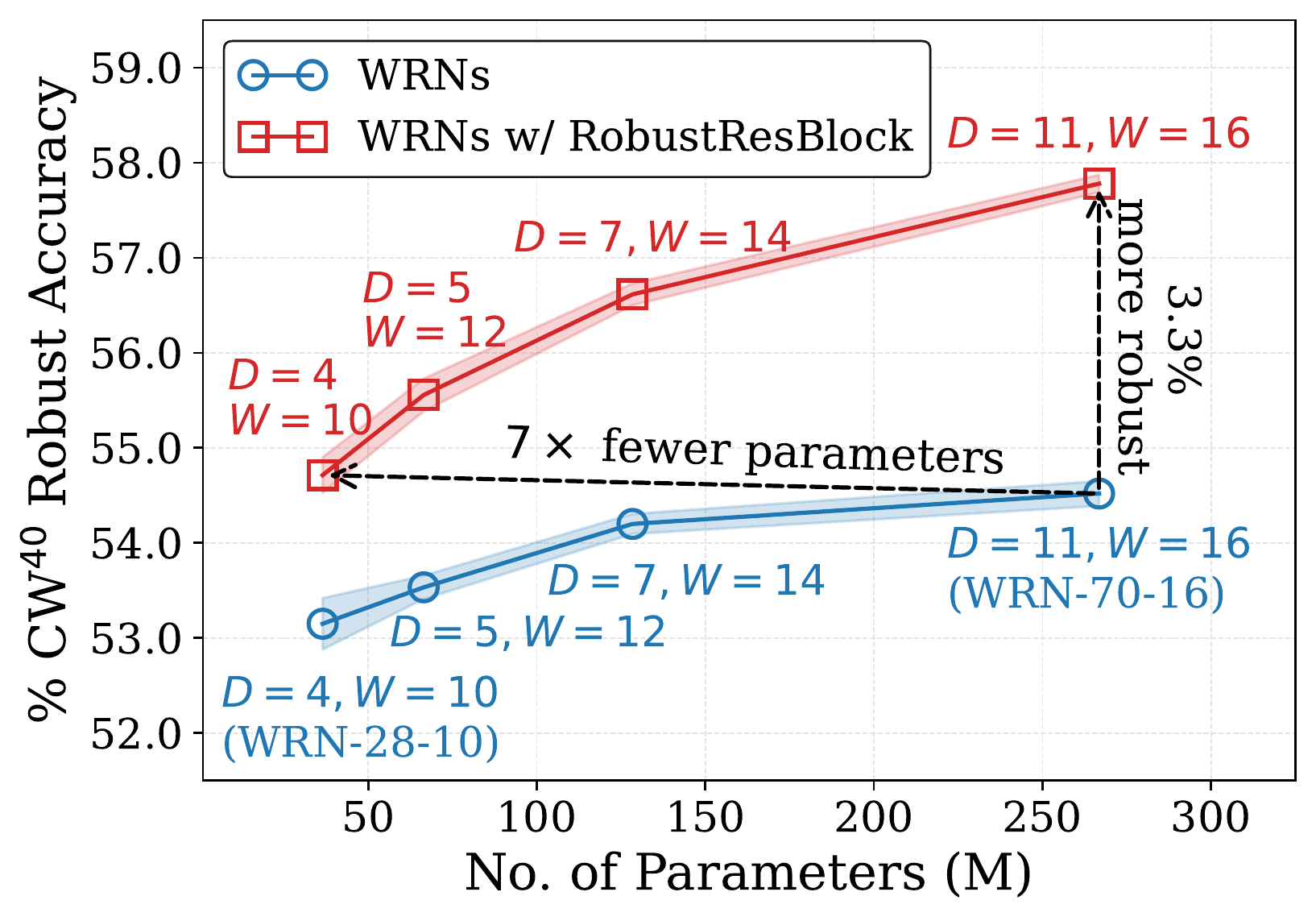}};
        \end{tikzpicture}
    \end{tcolorbox}
\end{figure}
% -------------------------------------------------------------------------------------

\subsubsection{Block Topology\label{sec:block_topology}}
\noindent\textbf{Residual Topology:} Figure~\ref{fig:overview} (a, b, c) shows the primary variants of residual blocks in the literature, namely, basic~\cite{he2016deep}, bottleneck~\cite{he2016deep}, and inverted bottleneck~\cite{mobilenetv2}. Among them, the basic block is the de-facto choice for studying adversarial robustness \cite{zhang2019theoretically, Wang2020Improving,rebuffi2021data,gowal2021improving}. Surprisingly, the bottleneck and inverted bottleneck blocks have rarely been employed for adversarial robustness, despite their well-established effectiveness under standard ERM training for image classification, object detection, etc. \cite{wightman2021resnet,tan2019efficientnet}. Therefore, we revisit these residual blocks in the context of adversarial robustness. And for each block, we consider two variants (post-activation \cite{he2016deep} and pre-activation \cite{he2016identity}) corresponding to the placement of activation functions before and after a convolution ({see Figure~\ref{fig:overview} (d, e) for an illustration}). Moreover, we consider models of different capacities by varying the stage-wise depth $D_{i\in\{1,2,3\}}$ and width $W_{i\in\{1,2,3\}}$ among \{4, 5, 7, 11\} and \{10, 12, 14, 16\}, respectively.

% -------------------------------------------------------------------------------------
\begin{figure}[t]
    \begin{subfigure}[b]{0.23\textwidth}
    \centering
    \includegraphics[width=0.95\textwidth]{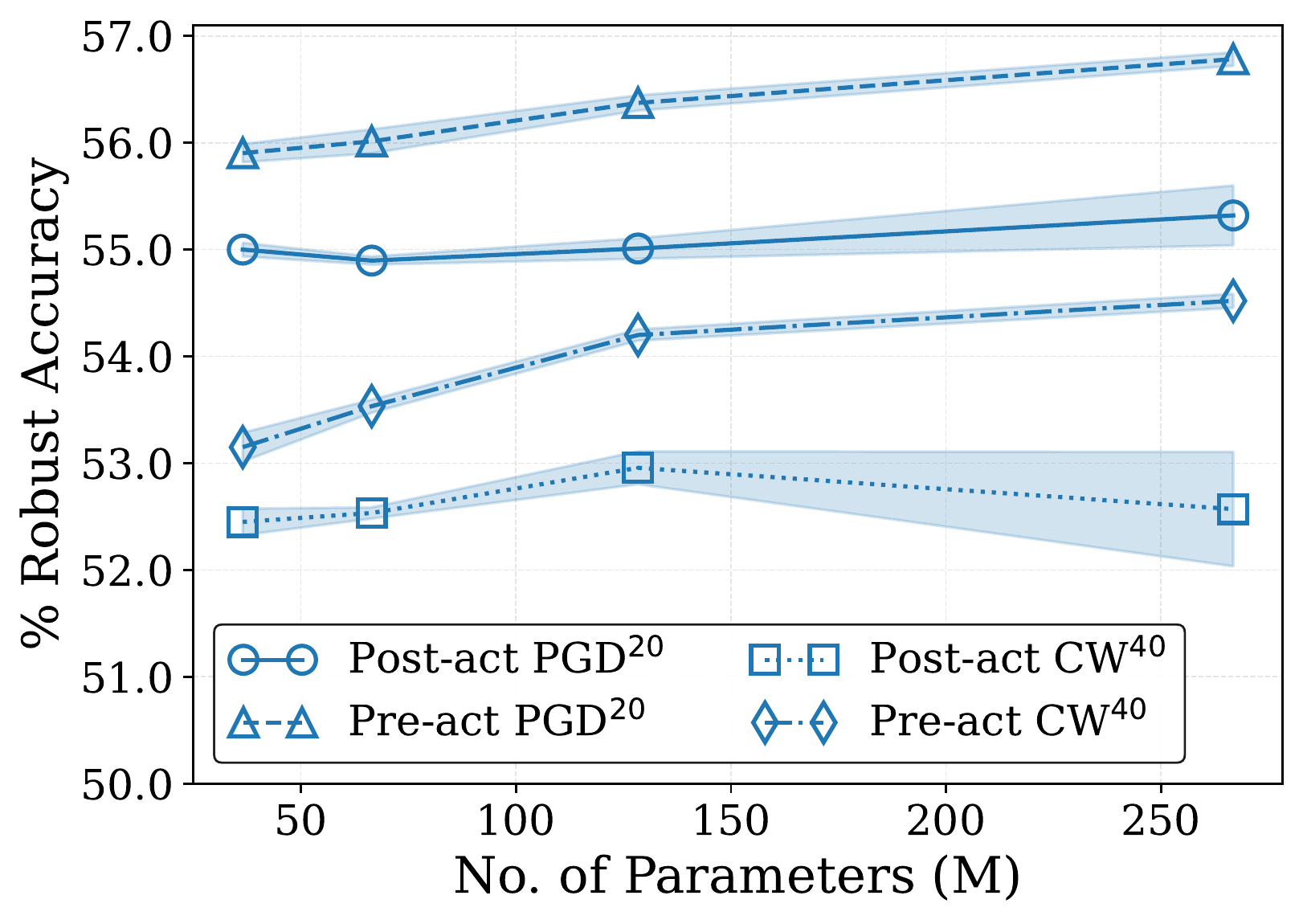}
    \caption{\scriptsize Basic \label{fig:abl_topology_c10_basic}}
    \end{subfigure}\hfill
    \begin{subfigure}[b]{0.23\textwidth}
    \centering
    \includegraphics[width=0.95\textwidth]{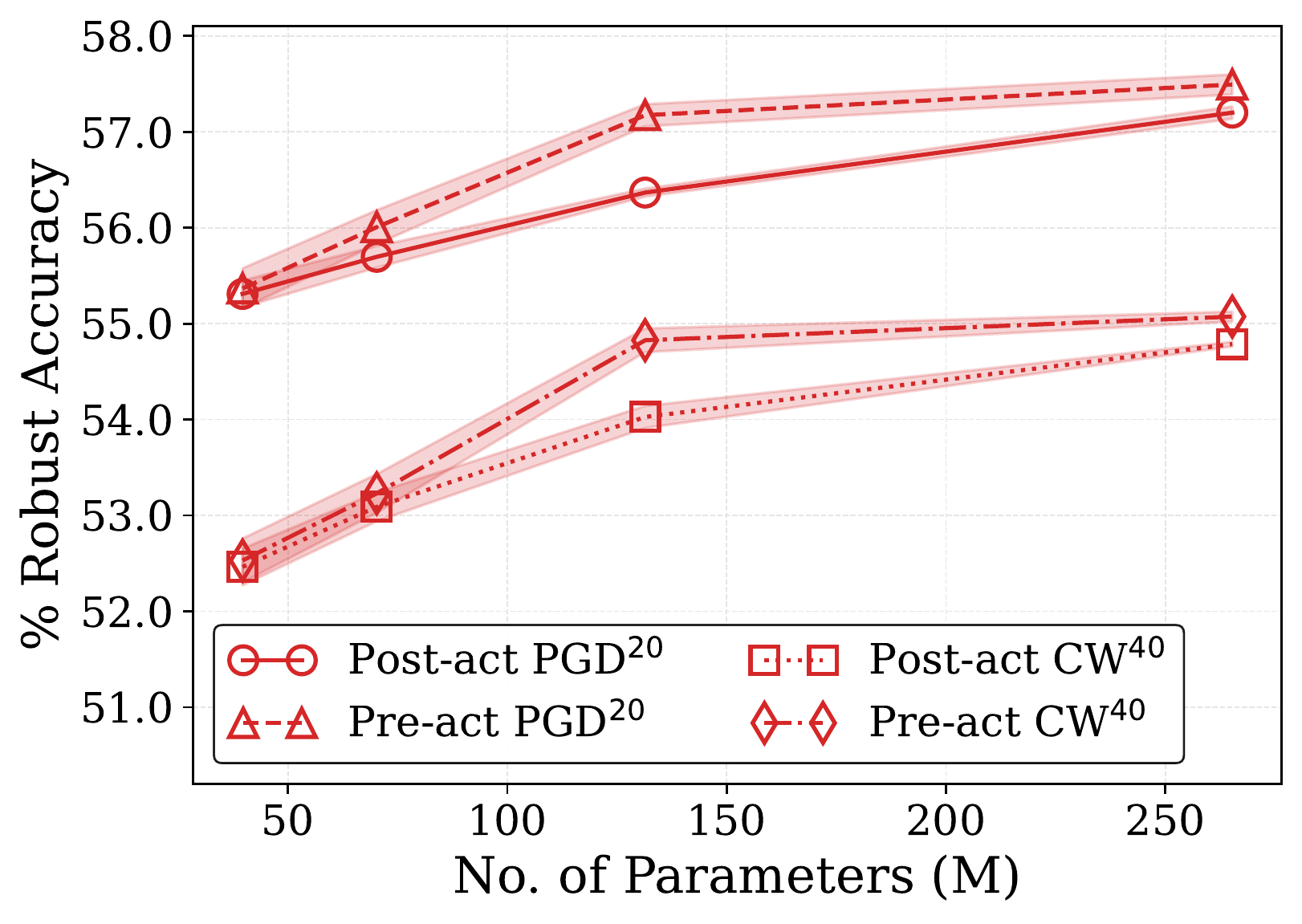}
    \caption{\scriptsize Bottleneck \label{fig:abl_topology_c10_bottleneck}}
    \end{subfigure}\\
    \begin{subfigure}[b]{0.23\textwidth}
    \centering
    \includegraphics[width=0.95\textwidth]{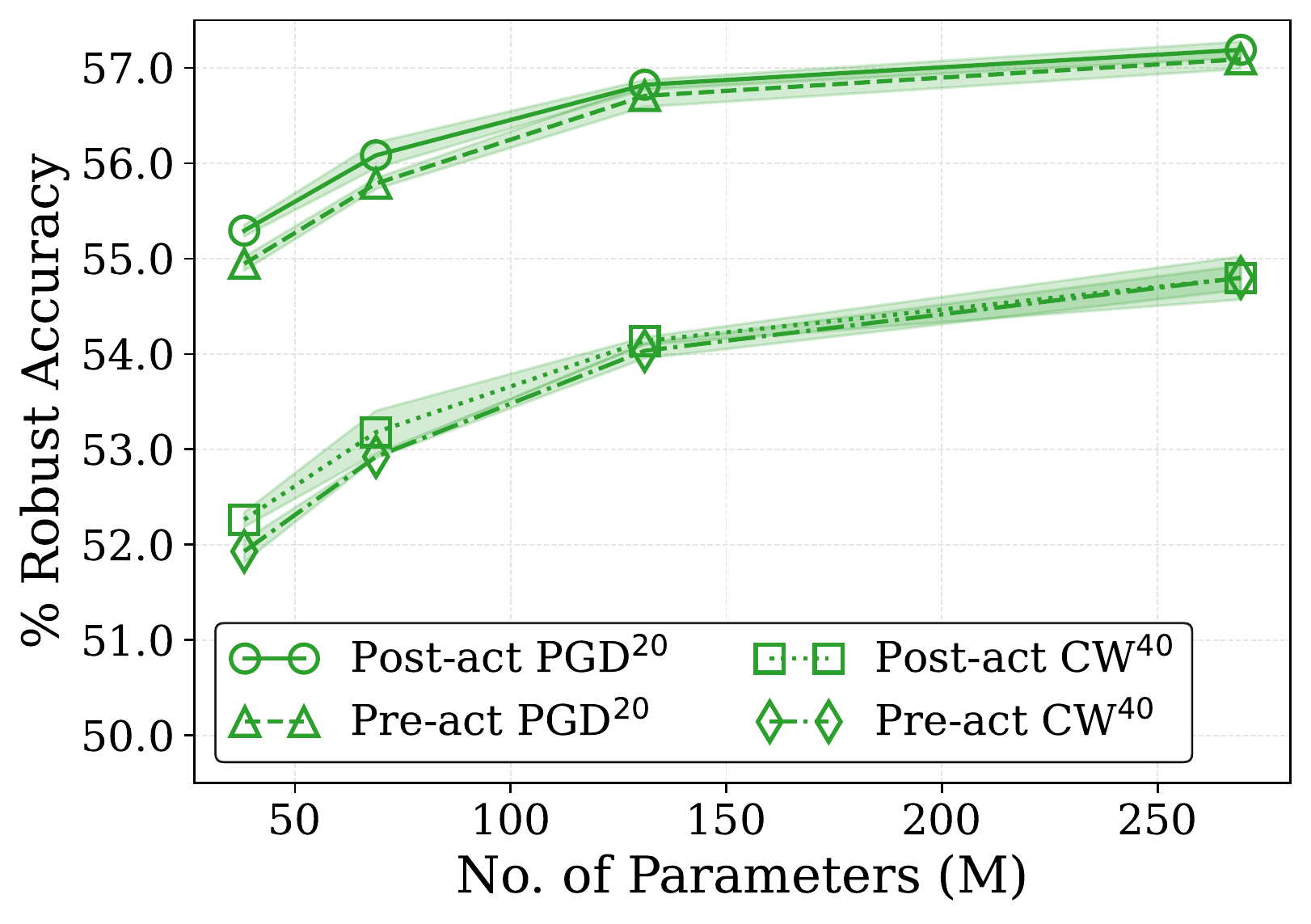}
    \caption{\scriptsize Inverted Bottleneck \label{fig:abl_topology_c10_inverted}}
    \end{subfigure}\hfill
    \begin{subfigure}[b]{0.23\textwidth}
    \centering
    \includegraphics[width=0.95\textwidth]{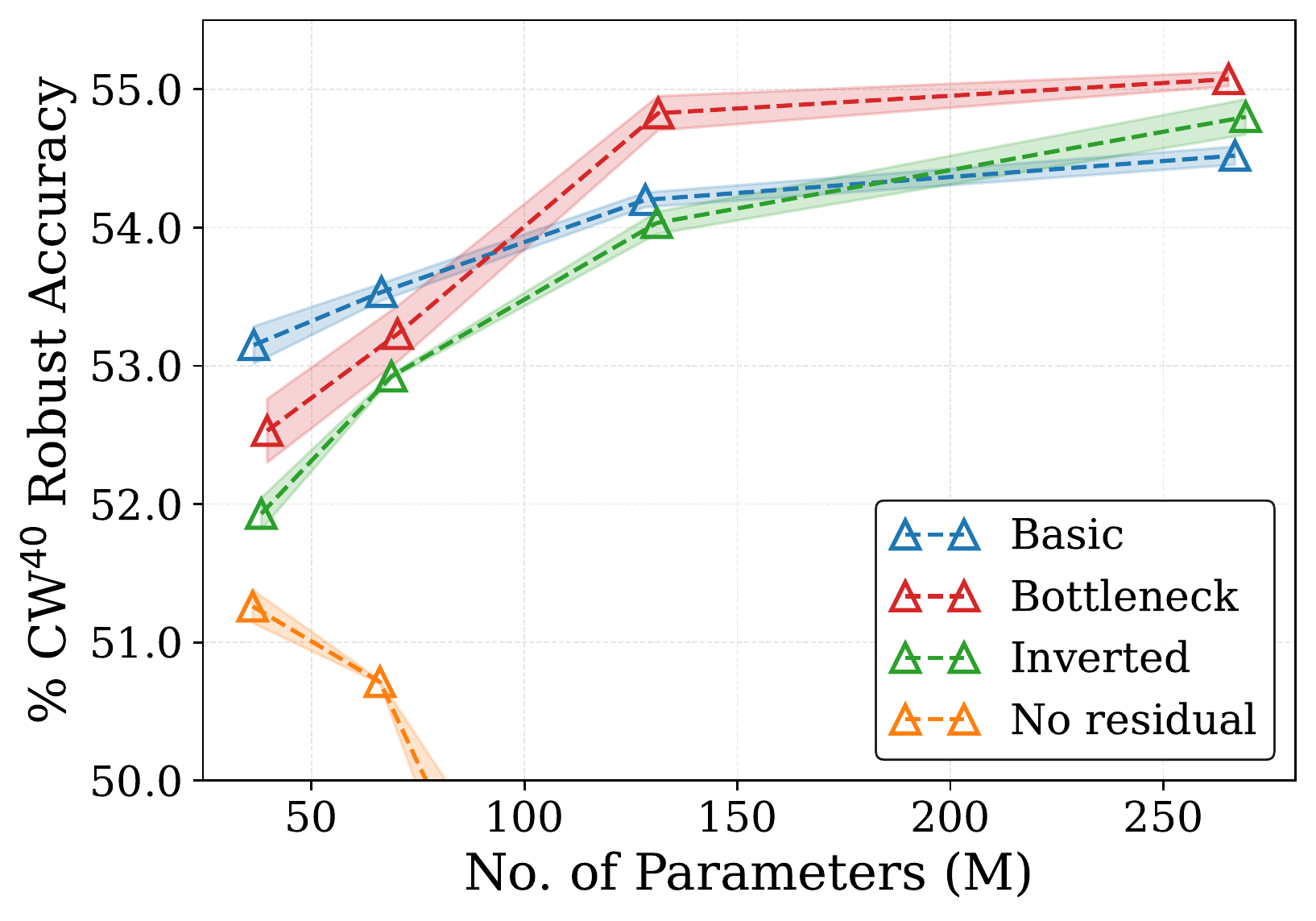}
    \caption{\scriptsize Comparison among (a) – (c) \label{fig:abl_topology_c10_cw40}}
    \end{subfigure}
    \vspace{-8pt}
    \caption{Robust accuracy of networks on C-10 with (a) basic, (b) bottleneck, and (c) inverted bottleneck blocks, with post and pre-activation. (d) Comparison among blocks with pre-activation. ``No residual'' removes the residual connection in the basic block. \label{fig:abl_topology}\vspace{-15pt}}
\end{figure}
% -------------------------------------------------------------------------------------

Figure~\ref{fig:abl_topology} compares the adversarial robustness of the above variants of residual blocks under baseline AT. We observe that (i) the basic block is susceptible to the location of the activation function, with pre-activation leading to a substantial improvement in adversarial robustness (Fig.~\ref{fig:abl_topology_c10_basic}); (ii) performance of the bottleneck and inverted bottleneck blocks are relatively stable w.r.t the position of the activation function, although pre-activation provides a slight but noticeable benefit on large-capacity models with bottleneck blocks and small-capacity models with inverted bottleneck blocks (Figs.~\ref{fig:abl_topology_c10_bottleneck} and \ref{fig:abl_topology_c10_inverted}). Thus, we argue that \emph{pre-activation is preferred over post-activation for adversarial robustness}. Figure~\ref{fig:abl_topology_c10_cw40} compares the three residual blocks with pre-activation under baseline AT. We observe that the basic block is more effective in low model-capacity regions, while the bottleneck block is more effective in high model-capacity regions. Finally, since the inverted bottleneck does not outperform the other two blocks under any model capacity, we do not consider it any further. Additional results are available in Appendix \S\ref{sec:app_block_topology}.

% -------------------------------------------------------------------------------------
\begin{figure}[t]
    \begin{subfigure}[b]{0.23\textwidth}
    \captionsetup{justification=centering}
    \includegraphics[width=.95\textwidth]{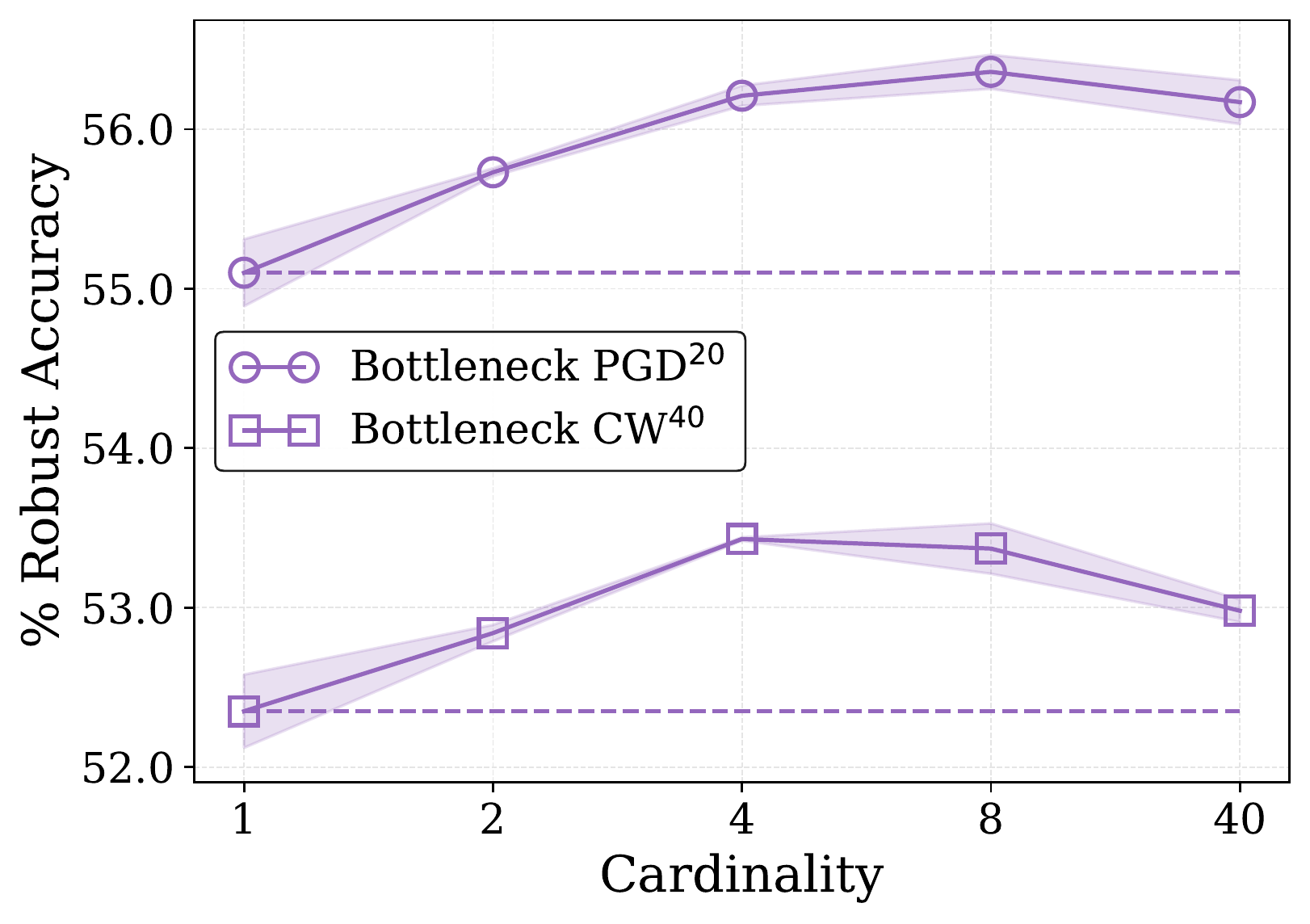}
    \caption{\label{fig:abl_aggre_d28_w10_c10}}
    \end{subfigure}\hfill
    \begin{subfigure}[b]{0.23\textwidth}
    \captionsetup{justification=centering}
    \includegraphics[width=.95\textwidth]{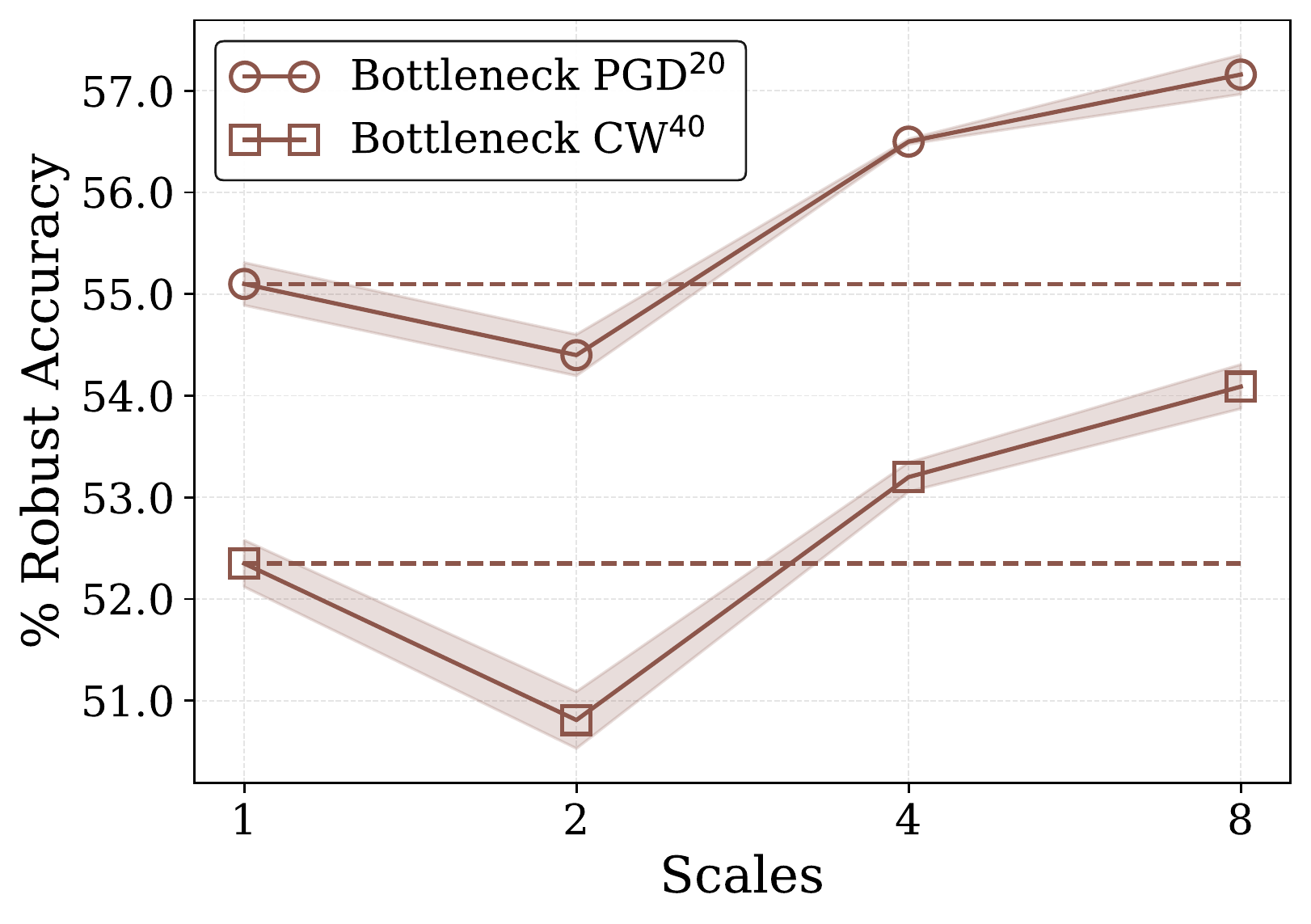}
    \caption{\label{fig:abl_hier_d28_w10_c10}}
    \end{subfigure} \\
    \begin{subfigure}[b]{0.23\textwidth}
    \captionsetup{justification=centering}
    \includegraphics[width=.95\textwidth]{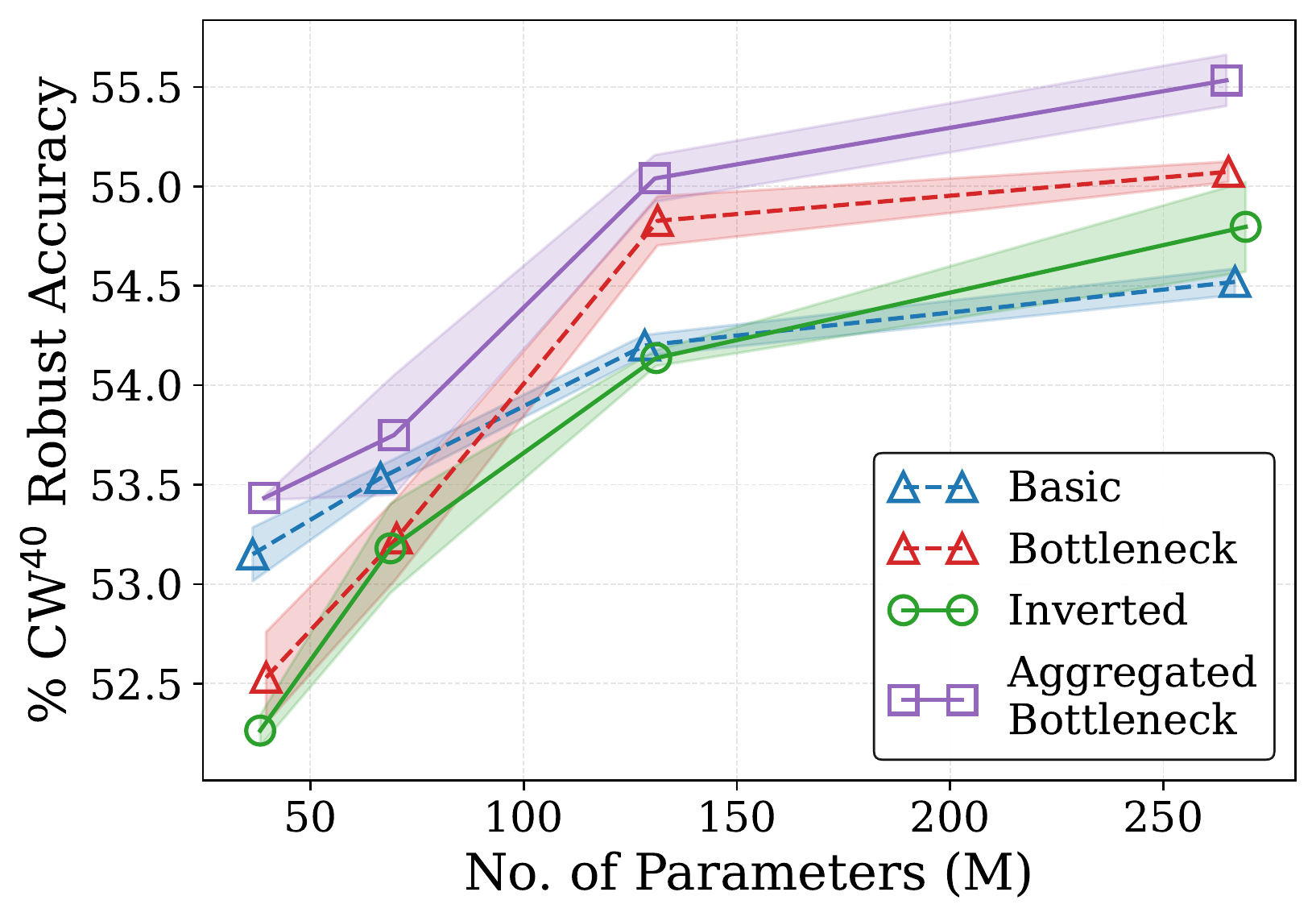}
    \caption{\label{fig:abl_topology_aggre_c10_cw40}}
    \end{subfigure}\hfill
    \begin{subfigure}[b]{0.23\textwidth}
    \captionsetup{justification=centering}
    \includegraphics[width=.95\textwidth]{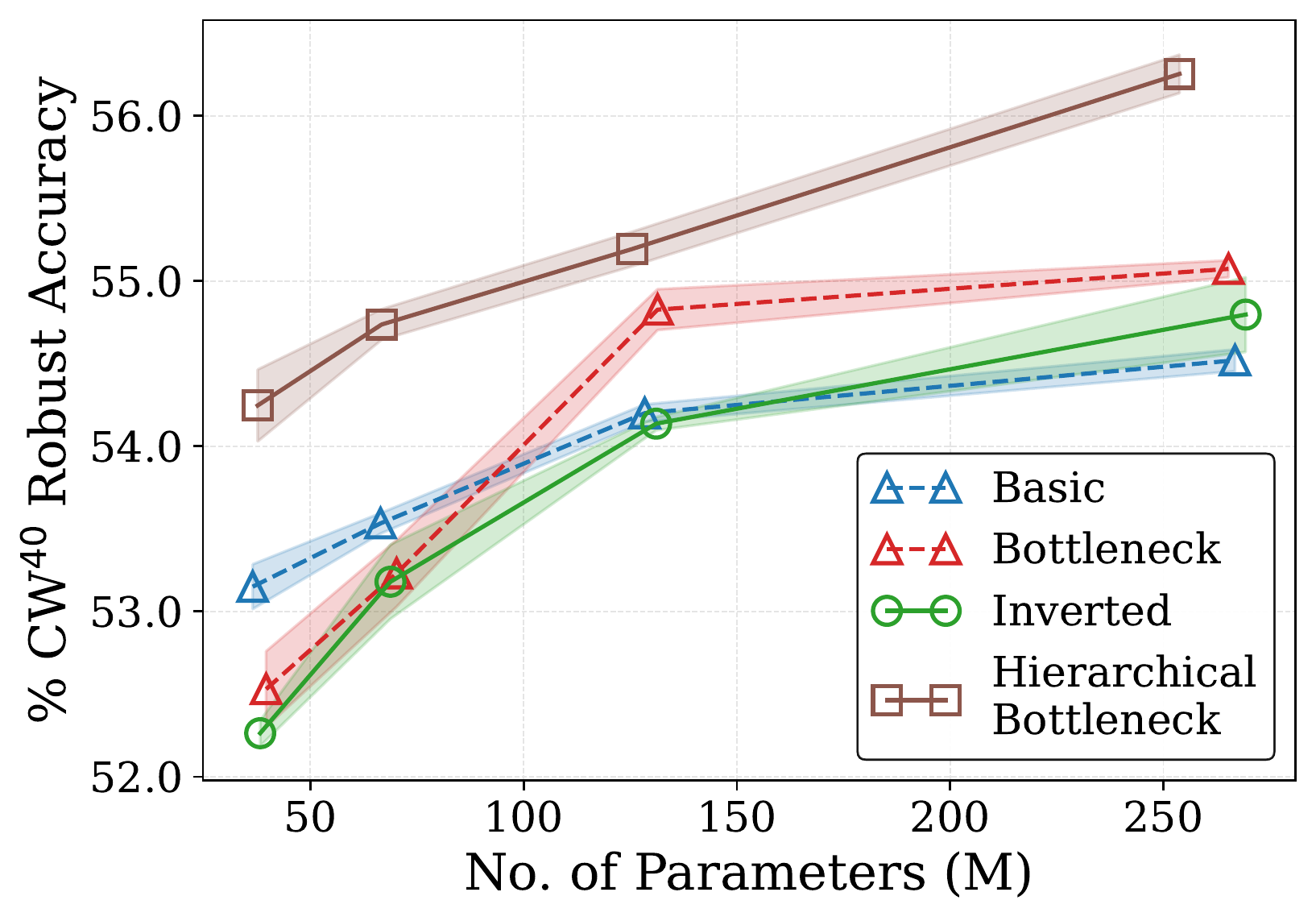}
    \caption{\label{fig:abl_topology_hier_c10_cw40}}
    \end{subfigure}
    \vspace{-10pt}
    \caption{(a, b) show effects of cardinality and scales for a low-capacity model ({\small$D_i=4, W_i=10$}). (c, d) Comparing aggregated ({\small cardinality $=4$}) and hierarchical ({\small scales $=8$}) bottleneck to other blocks. All results are on CIFAR-10. \label{fig:abl_aggre_hier_c10}\vspace{-15pt}}
\end{figure}
% -------------------------------------------------------------------------------------

% -------------------------------------------------------------------------------------
\begin{table*}[ht]
\centering
\caption{Break-down of the contribution of each identified topological enhancement. Both basic and bottleneck blocks use pre-activation. The cardinality for aggregated conv is 4, and the scale for hierarchical conv is 8. All results are for a large model with $D_i=11, W_i=16$. \label{tab:topology_summary}\vspace{-5pt}}
\resizebox{.9\textwidth}{!}{%
\begin{tabular}{@{\hspace{2mm}}cc|cc|c|cc|ccc|ccc@{\hspace{2mm}}}
\toprule
\multicolumn{5}{c|}{Topology} & 
\multicolumn{2}{c|}{Complexity} & \multicolumn{3}{c|}{CIFAR-10} & \multicolumn{3}{c}{CIFAR-100} \\ \cmidrule(lr){1-5} \cmidrule(lr){6-7} \cmidrule(lr){8-10} \cmidrule(lr){11-13}
Basic & Bottle & Aggr. & Hier. & SE & $^{\#}$P & $^{\#}$F & Clean & PGD$^{20}$ & CW$^{40}$ & Clean & PGD$^{20}$ & CW$^{40}$  \\ \midrule
$\checkmark$ &  &  &  &  & 267M & 38.8G & $85.51_{\pm0.19}$ & $56.78_{\pm0.13}$ & $54.52_{\pm0.13}$ & $56.93_{\pm0.49}$ & $29.76_{\pm0.14}$ & $27.24_{\pm0.15}$ \\
% bottle
 & $\checkmark$ &  &  &  & 265M & 39.0G & $85.47_{\pm0.21}$ & $57.49_{\pm0.21}$ & $55.07_{\pm0.10}$ & $59.24_{\pm0.36}$ & $32.08_{\pm0.26}$ & $28.61_{\pm0.17}$ \\
 % bottle + aggregate
 & $\checkmark$ & $\checkmark$ &  &  & 265M & 39.4G & $85.47_{\pm0.10}$ & $57.50_{\pm0.28}$ &  $55.53_{\pm0.26}$ & $59.27_{\pm0.34}$ & $31.63_{\pm0.36}$ & $28.80_{\pm0.18}$\\
 % bottle + aggregate + hierarchy 
 & $\checkmark$ & $\checkmark$ & $\checkmark$ &  & 262M & 39.3G & $86.29_{\pm0.07}$ & $59.48_{\pm0.12}$ & $56.94_{\pm0.27}$ & $59.32_{\pm0.13}$ & $33.46_{\pm0.22}$ & $29.65_{\pm0.14}$ \\
 % final block 
 & $\checkmark$ & $\checkmark$ & $\checkmark$ & $\checkmark$ & 270M & 39.3G & \bm{$86.55_{\pm0.10}$} & \bm{$60.48_{\pm0.00}$} & \bm{$57.78_{\pm0.09}$} & \bm{$60.22_{\pm0.57}$} & \bm{$33.88_{\pm0.03}$} & \bm{$29.91_{\pm0.15}$} \\
\bottomrule
\end{tabular}%
}
\vspace{-10pt}
\end{table*}
% -------------------------------------------------------------------------------------

\vspace{2pt}
\noindent\textbf{Aggregated and Hierarchical Convolutions:} Next, we consider two enhanced arrangements of convolution, \emph{aggregated}~\cite{xie2017aggregated}, and \emph{hierarchical}~\cite{res2net}, which have proven to be effective for residual blocks under standard EMR training on standard tasks. Aggregated and hierarchical convolutions split a regular convolution into multiple parallel convolutions and hierarchical convolutions; see Figure~\ref{fig:overview} (g, h) for visualizations. We incorporate both of them within the bottleneck block. For each enhancement, experiments with parameter sweeps were carried out to determine appropriate values for their hyperparameters, i.e., \emph{cardinality} for aggregated (Figure~\ref{fig:abl_aggre_d28_w10_c10}) and \emph{scales} for hierarchical convolutions (Figure~\ref{fig:abl_hier_d28_w10_c10}). Figures~\ref{fig:abl_aggre_hier_c10} (c, d) compare the bottleneck block with aggregated and hierarchical convolutions under baseline AT, respectively. We observe that the \emph{bottleneck block consistently benefits from both enhancements} and outperforms the basic block across a wide spectrum of model-capacity regions. \emph{On the other hand, aggregated convolution adversely affects adversarial robustness when paired with the basic block.} More detailed results can be found in Appendix \S\ref{sec:app_block_connection}.

% -------------------------------------------------------------------------------------
\begin{figure}[t]
    \begin{subfigure}[b]{0.1\textwidth}
    \centering
    \captionsetup{justification=centering}
    \includegraphics[height=.105\textheight]{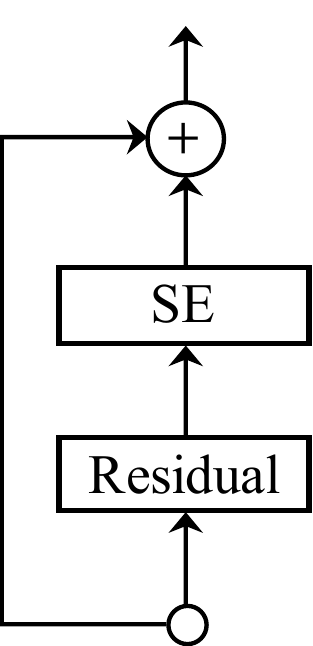}
    \caption{\scriptsize SE \label{fig:se}}
    \end{subfigure} \hfill
    \begin{subfigure}[b]{0.1\textwidth}
    \centering
    \captionsetup{justification=centering}
    \includegraphics[height=.105\textheight]{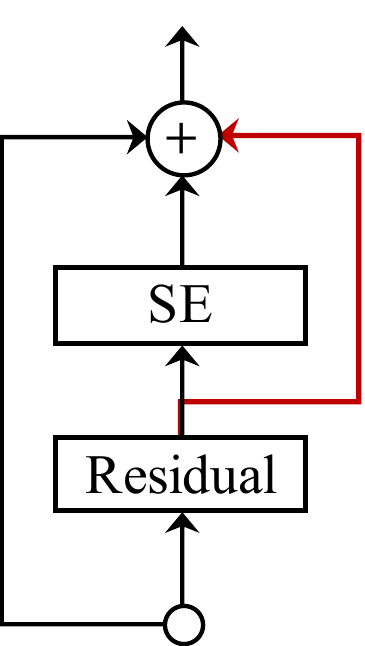}
    \caption{\scriptsize Our SE \label{fig:our_se}}
    \end{subfigure} \hfill
    \begin{subfigure}[b]{0.23\textwidth}
    \centering
    \captionsetup{justification=centering}
    \includegraphics[trim={0, 3mm, 0, 0}, clip, height=.105\textheight]{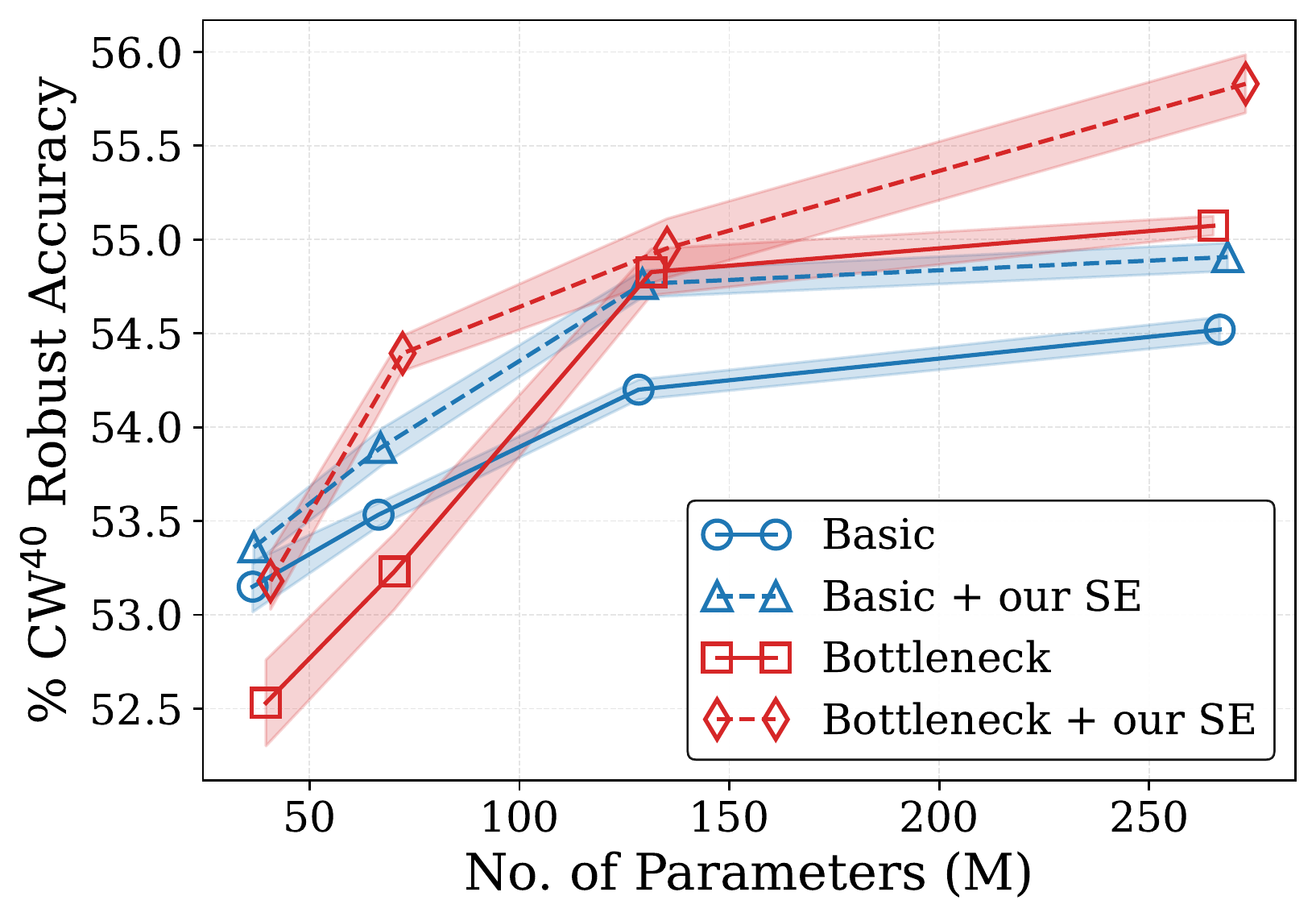}
    \caption{\label{fig:se_comparison}}
    \end{subfigure}
    \\ 
    \centering
    \begin{subfigure}[b]{0.45\textwidth}
    \centering
    \captionsetup{justification=centering}
    \resizebox{.98\textwidth}{!}{%
        \begin{tabular}{@{\hspace{2mm}}lcccc@{\hspace{2mm}}}
        \toprule
        Designs (reduction ratio) & $^{\#}$P (M) & $^{\#}$F (G) & Clean & Robust (CW$^{40}$) \\ \midrule
        w/o SE & 265 & 39.0 & 85.47 & 55.07 \\
        Standard SE ($r=16$) & 296 & 39.1 & 84.56 (\color{red}{-0.91}) & 54.52 (\color{red}{-0.55}) \\
        Conv3$\times$3-SE ($r=16$) & 273 & 39.1 & 85.26 (\color{red}{-0.21}) & 54.77 (\color{red}{-0.40}) \\
        Identity-SE ($r=16$) & 293 & 39.1 & 85.20 (\color{red}{-0.27}) & 54.94 (\color{red}{-0.13}) \\ \midrule
        Our residual SE ($r=16$) & 296 & 39.1 & \textbf{85.75} (\color{Emerald}{+0.28}) & {55.95} ({\color{Emerald}+0.88}) \\
        Our residual SE ($r=64$) & 273 & 39.1 & {85.61} (\color{Emerald}{+0.14}) & \textbf{56.05} ({\color{Emerald}+0.98}) \\ \bottomrule
        \end{tabular}%
    }
    \caption{\scriptsize Ablation study on SE integration designs. \label{tab:abl_se_integration}}
    \end{subfigure}
    \vspace{-10pt}
    \caption{ (a) Standard \emph{SE} block. (b) Our \emph{residual SE} adds an extra skip connection around the SE module. (c) Comparison of residual blocks w/ and w/o our residual SE. (d) Ablation results with relative {\color{Emerald}improvement}/{\color{red}degradation} shown in parentheses. \label{fig:abl_se} \vspace{-15pt}}
\end{figure}
% -------------------------------------------------------------------------------------

\vspace{2pt}
\noindent\textbf{Squeeze and Excitation:} Next, we consider squeeze-and-excitation (SE) \cite{senet}, which emerged as a standard component of modern CNN architectures, such as MobileNetV3 \cite{howard2019searching}, and EfficientNet \cite{tan2019efficientnet}. However, we observe (see Table~\ref{tab:abl_se_integration}) that a straightforward application of SE, and all its variants explored by Hu \etal~\cite{senet}, degrade adversarial robustness. This is unlike the case in standard ERM training, where SE consistently improves performance across most vision tasks when added to residual networks. We hypothesize that this may be due to the SE layer excessively suppressing or amplifying channels. Therefore, we present an alternative variant of SE, dubbed \emph{residual SE}, for adversarial robustness. As shown in Figure~\ref{fig:our_se}, we add another skip connection around the SE module---a simple yet crucial modification. During adversarial training, this extra skip connection provides additional regularization to prevent channels from being excessively suppressed or amplified by SE. Figure~\ref{fig:se_comparison} compares the basic and bottleneck blocks with and without \emph{residual SE} under baseline AT. Results indicate that our \emph{residual SE} consistently improves the adversarial robustness of both blocks across different model-capacity regions. Furthermore, as shown in Table~\ref{tab:abl_se_integration}, we observe that a higher reduction ratio can reduce the computational complexity of the SE module at the cost of a marginal degradation in clean accuracy. Additional results are available in Appendix \S\ref{sec:app_se}.

\vspace{2pt}
\noindent\textbf{Summary:} We break down the contribution of each identified topological enhancement, namely, pre-activation, aggregated and hierarchical convolutions, and residual SE in Table~\ref{tab:topology_summary}. We demonstrate that all these enhancements can be naturally integrated within the bottleneck topology. Empirically, our final topology yields a {\footnotesize$\sim$}3\% improvement under baseline AT over the basic block used in WRNs, the de-facto topology of choice for designing robust architectures.

% -------------------------------------------------------------------------------------
\begin{figure}[t]
    \begin{subfigure}[b]{0.15\textwidth}
    \centering
    \includegraphics[width=\textwidth]{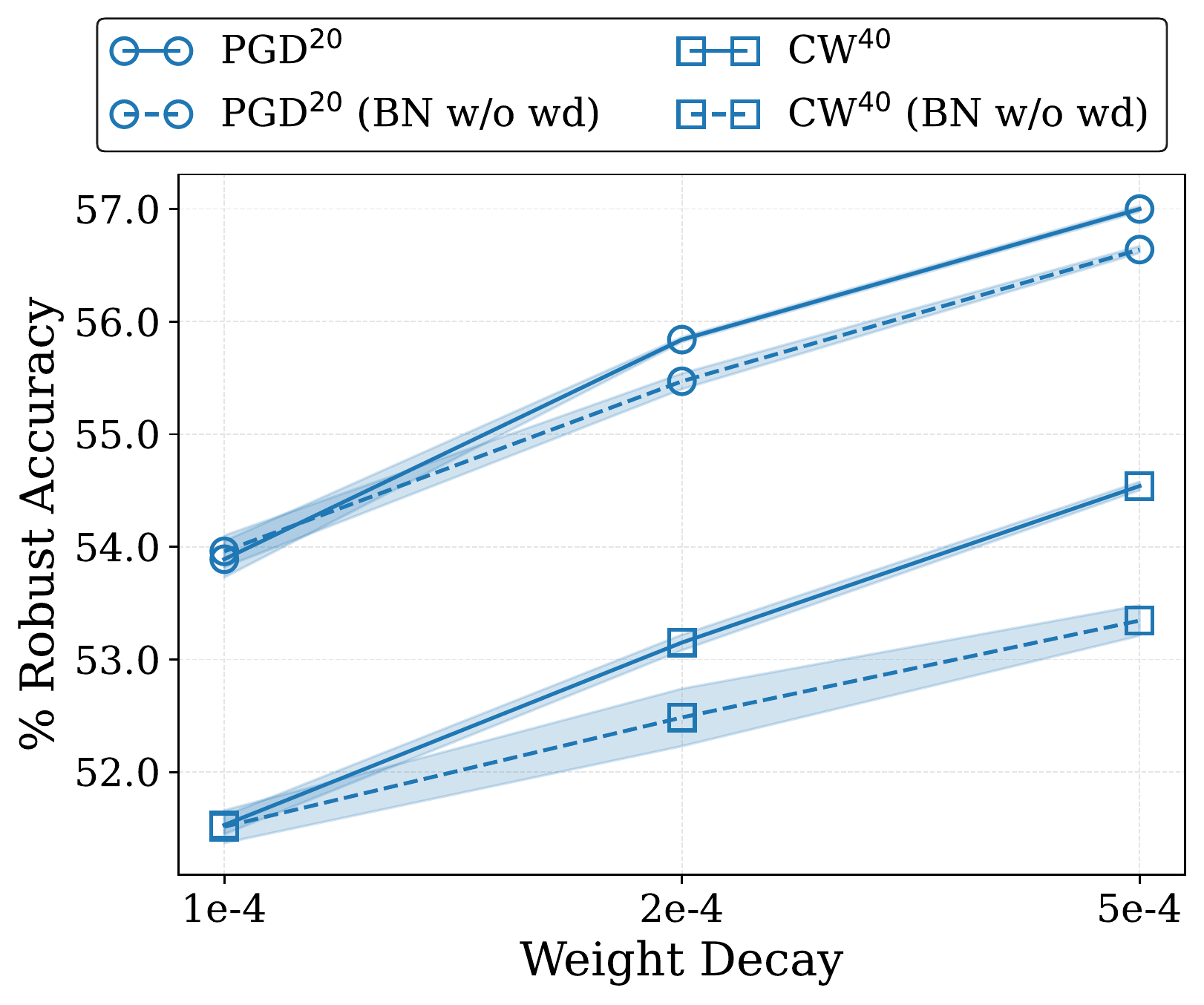}
    \caption{\scriptsize ReLU \label{fig:abl_act_d28_w10_c10_relu}}
    \end{subfigure} \hfill
    \begin{subfigure}[b]{0.15\textwidth}
    \centering
    \includegraphics[width=\textwidth]{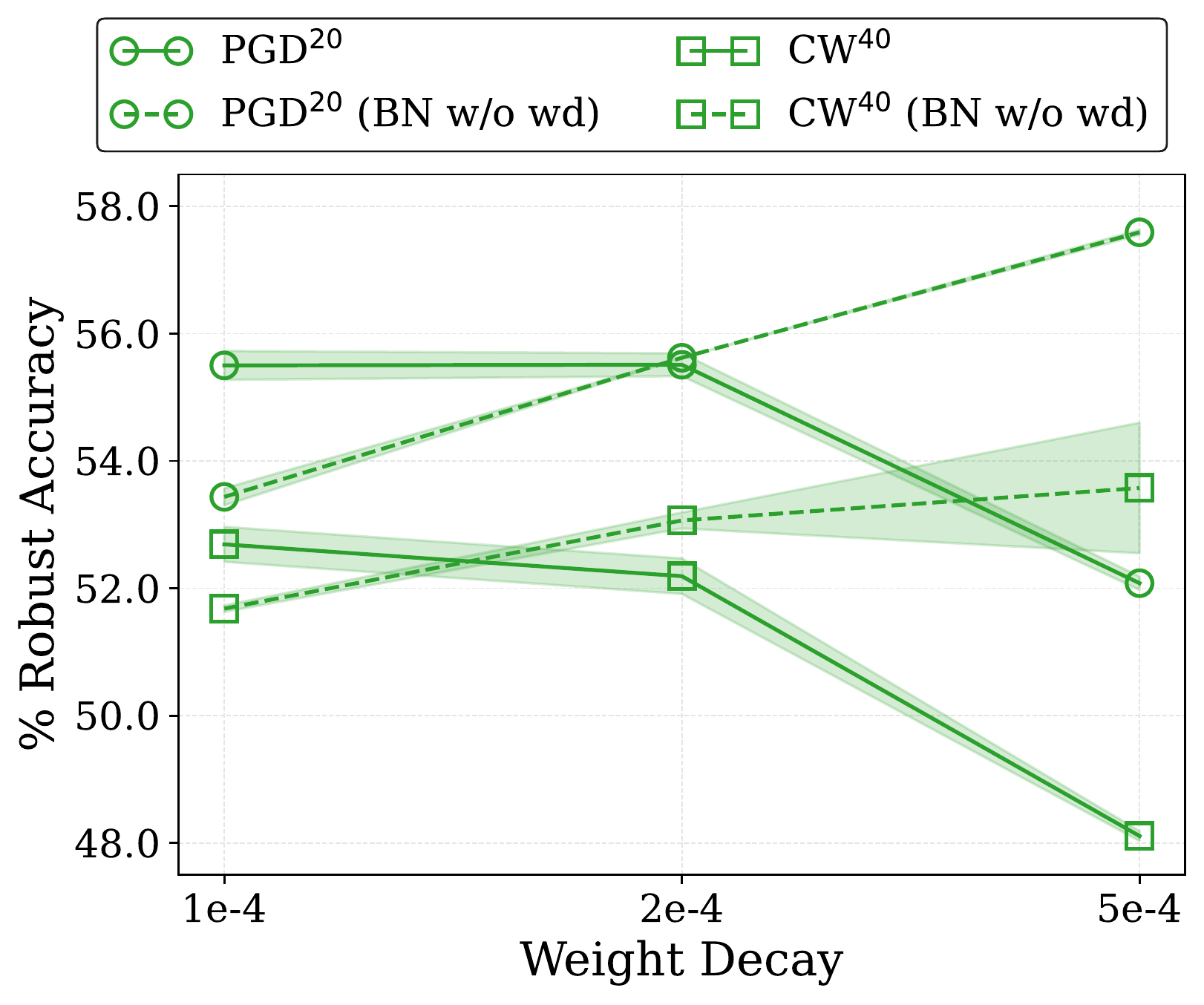}
    \caption{\scriptsize SiLU/Swish \label{fig:abl_act_d28_w10_c10_silu}}
    \end{subfigure} \hfill
    \begin{subfigure}[b]{0.15\textwidth}
    \centering
    \includegraphics[width=\textwidth]{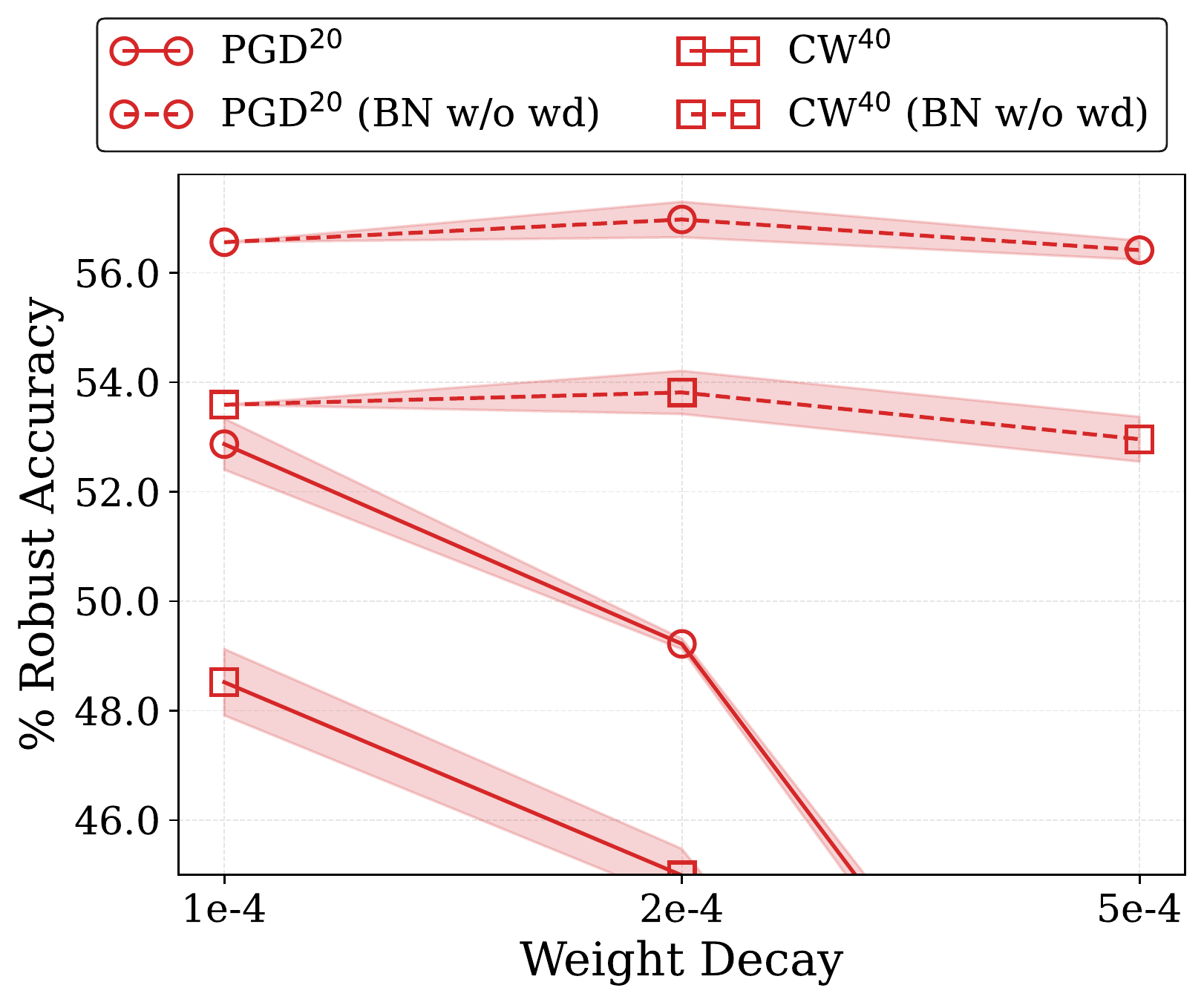}
    \caption{\scriptsize Softplus \label{fig:abl_act_d28_w10_c10_softplus}}
    \end{subfigure} \\
    \begin{subfigure}[b]{0.15\textwidth}
    \centering
    \includegraphics[width=\textwidth]{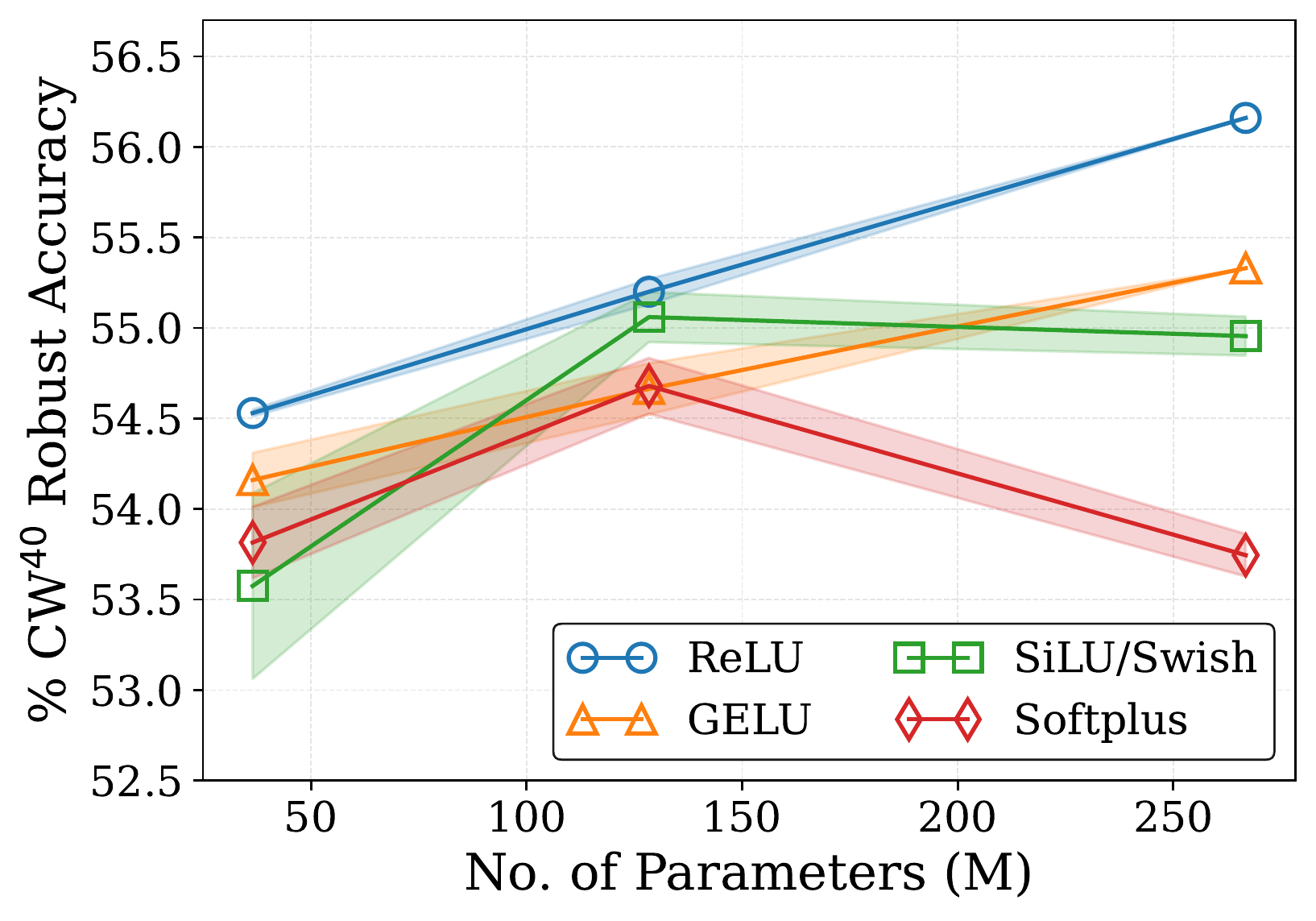}
    \caption{\scriptsize CIFAR-10 \label{fig:abl_act_c10_cw40}}
    \end{subfigure} \hfill
    \begin{subfigure}[b]{0.15\textwidth}
    \centering
    \includegraphics[width=\textwidth]{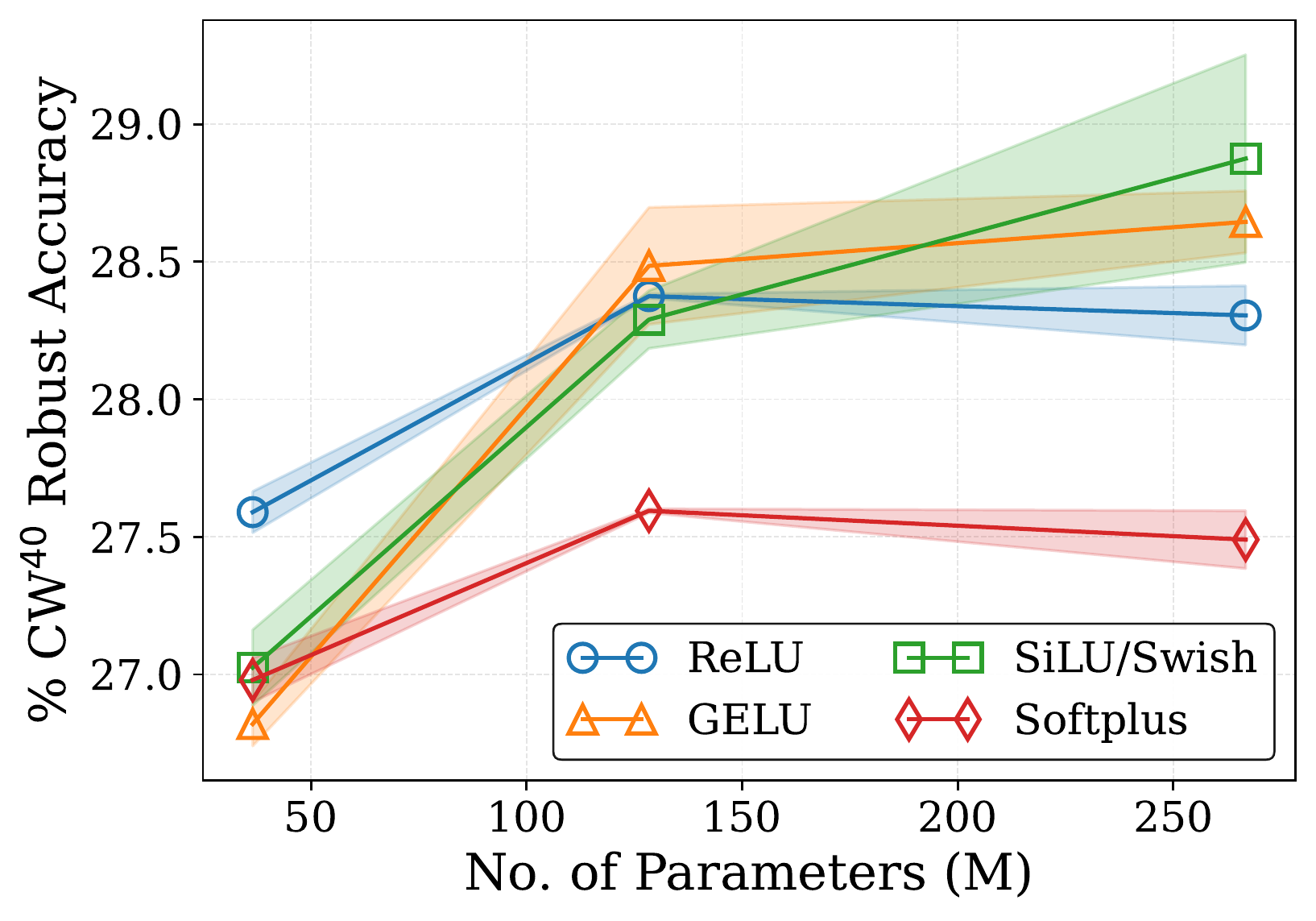}
    \caption{\scriptsize CIFAR-100 \label{fig:abl_act_c100_cw40}}
    \end{subfigure} \hfill
    \begin{subfigure}[b]{0.15\textwidth}
    \centering
    \includegraphics[width=\textwidth]{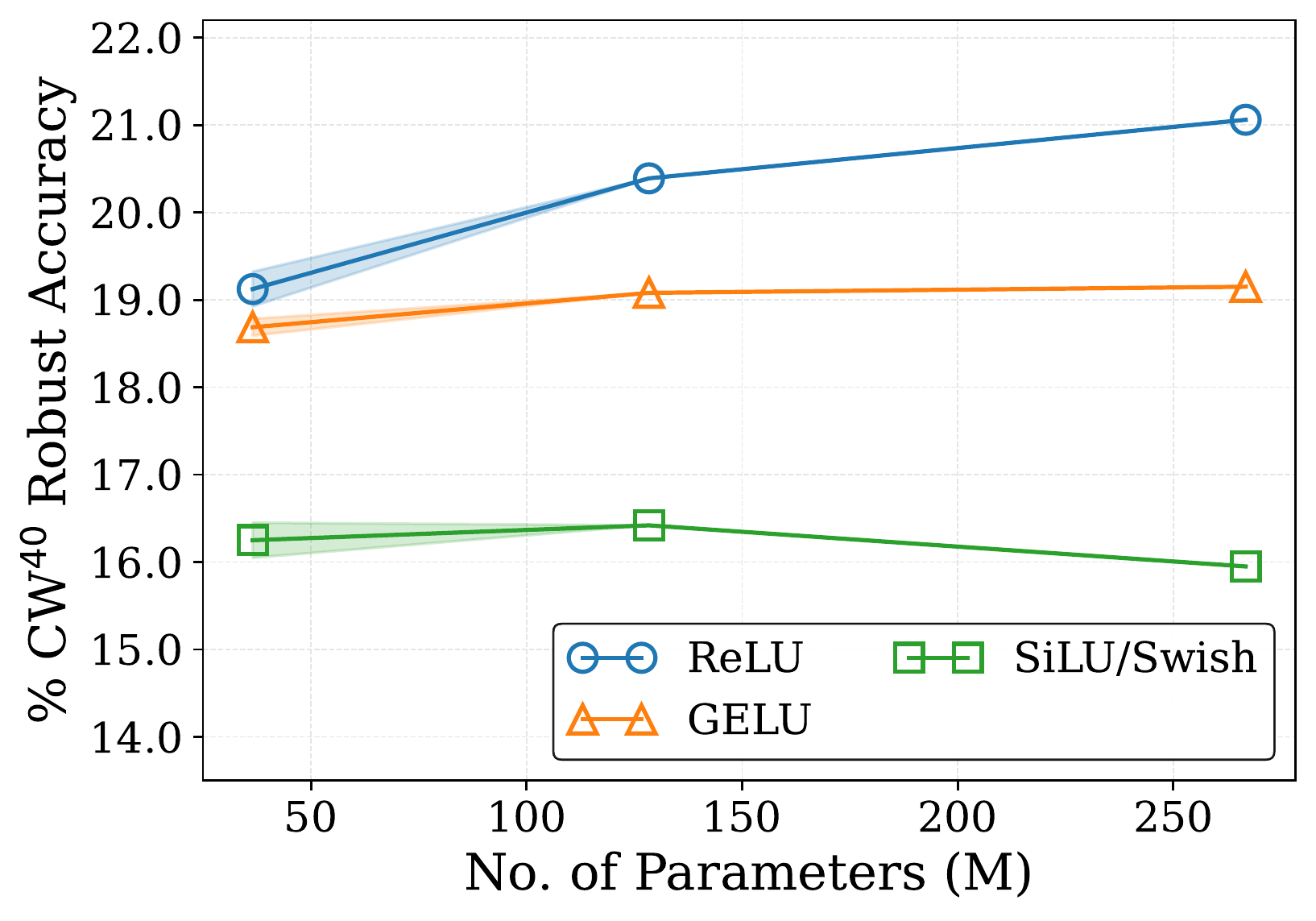}
    \caption{\scriptsize Tiny-ImageNet \label{fig:abl_act_tiny_cw40}}
    \end{subfigure} \\ 
    \begin{subfigure}[b]{0.15\textwidth}
    \centering
    \includegraphics[width=\textwidth]{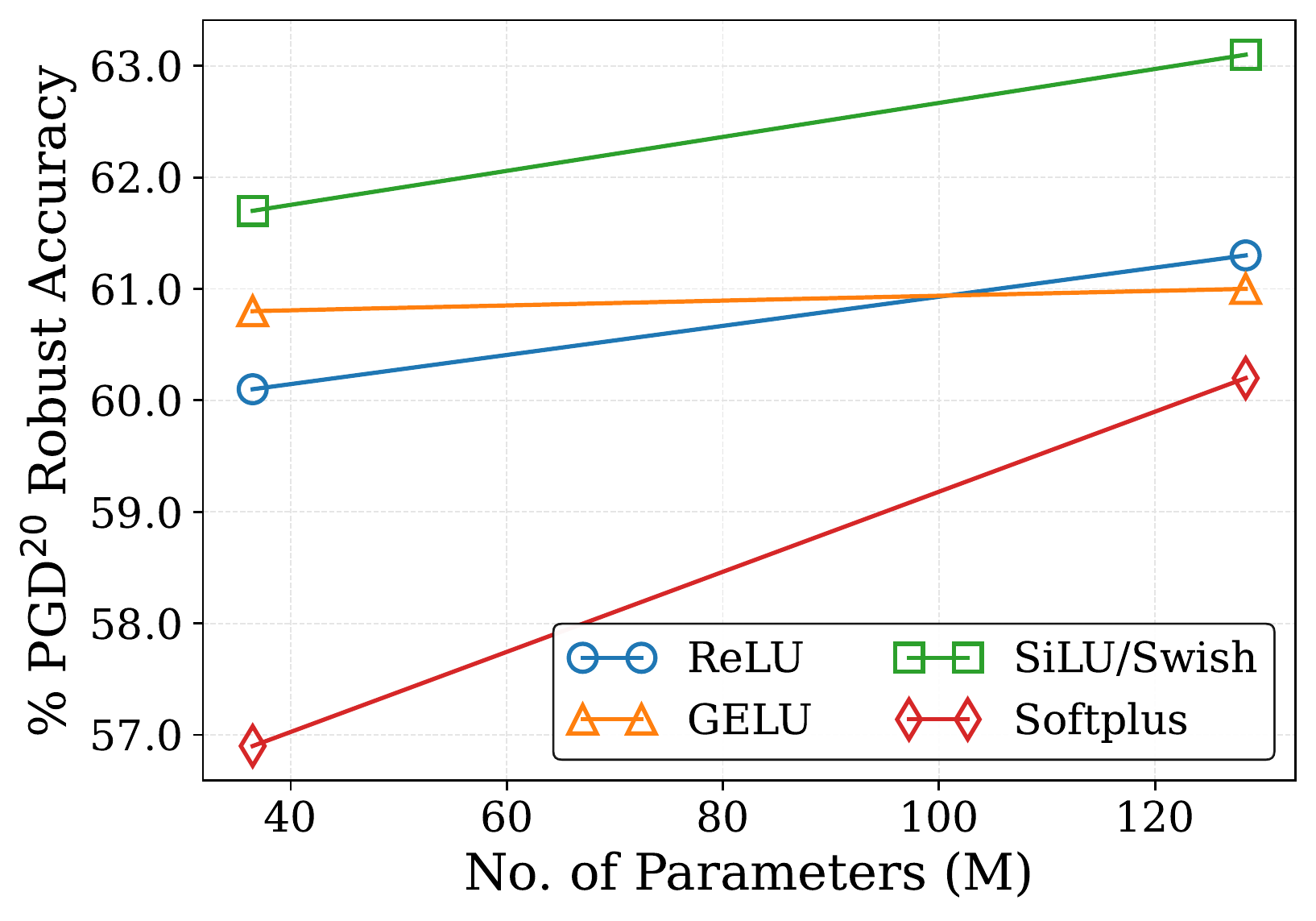}
    \caption{\scriptsize PGD$^{20}$ \label{fig:abl_act_c10_pgd20_adv}}
    \end{subfigure} \hfill
    \begin{subfigure}[b]{0.15\textwidth}
    \centering
    \includegraphics[width=\textwidth]{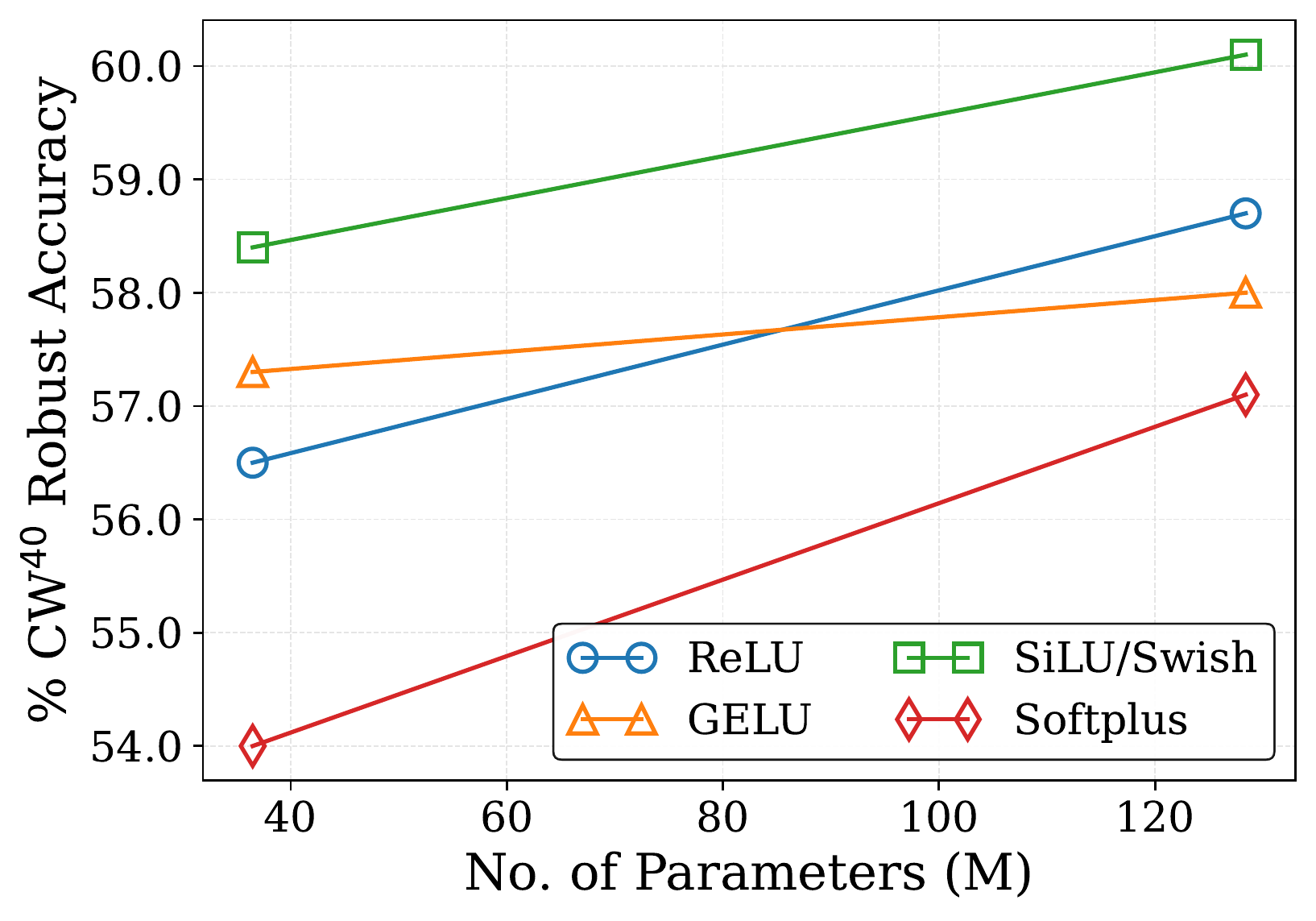}
    \caption{\scriptsize CW$^{40}$ \label{fig:abl_act_c10_cw40_adv}}
    \end{subfigure} \hfill
    \begin{subfigure}[b]{0.15\textwidth}
    \centering
    \includegraphics[width=\textwidth]{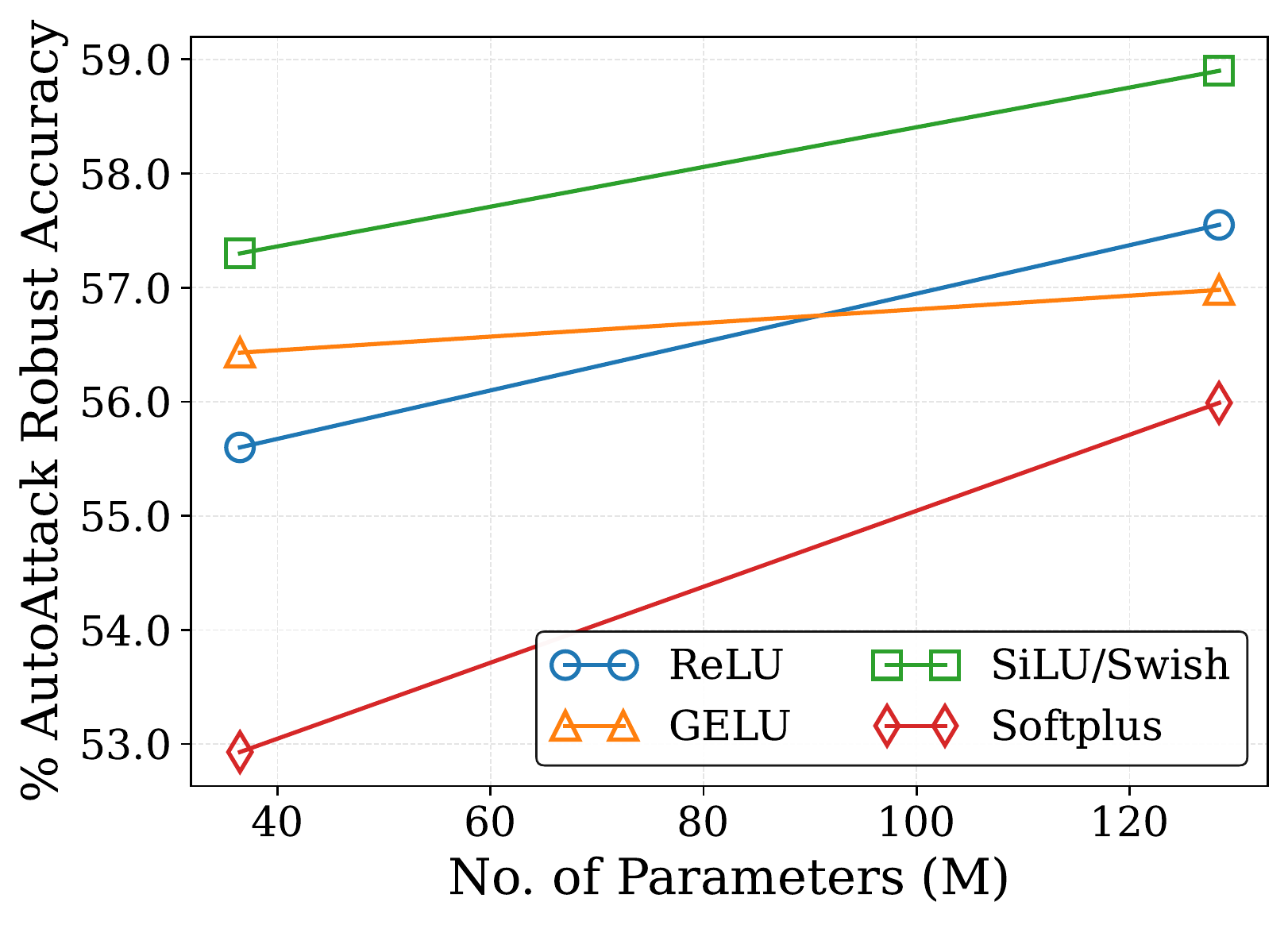}
    \caption{\scriptsize AutoAttack \label{fig:abl_act_c10_aa_adv}}
    \end{subfigure}
    \vspace{-8pt}
    \caption{(a) - (c) Effect of weight decay on robust accuracy of models with different activation functions on CIFAR-10. (d) - (f) Robust accuracy of models with different activation functions across a range of model capacities. (g) - (i) Robust accuracy under advanced AT for different activation functions on CIFAR-10. \label{fig:abl_act} \vspace{-15pt}}
\end{figure}
% -------------------------------------------------------------------------------------

\subsubsection{Activation and Normalization \label{sec:act_norm}}
\noindent\textbf{Activation:} Since the first demonstration by Xie \etal~\cite{xie2020smooth}, several researchers \cite{pang2020bag,singla2021low,gowal2020uncovering} reaffirmed that \emph{smooth activation functions improve adversarial training}, which in turn improves adversarial robustness. However, these observations are primarily based on CIFAR-10 with low-capacity models (e.g., ResNet-18 or WRN-34-10) and for a fixed set of training hyperparameters. We hypothesize that, smooth or not, different activation functions may perform differently depending on training hyperparameters, especially \emph{weight decay}, as observed by Pang \etal{} \cite{pang2020bag}. Therefore, we revisit the adversarial robustness of smooth and non-smooth activation functions under appropriate weight decay settings. We consider ReLU (non-smooth) and three smooth activation functions, SiLU/Swish \cite{xie2020smooth,rebuffi2021data,gowal2021improving}, Softplus~\cite{qin2019adversarial,pang2020bag}, and GELU~\cite{bai2021are}, given their prevalence in the literature.

We first identify a suitable weight decay value for each activation function from $\{1, 2, 5\} \times 10^{-4}$. From results in Figure~\ref{fig:abl_act} (a, b, c), we observe, under baseline AT, that (i) removing BN affine parameters from weight decay is crucial for smooth activation; (ii) different (but in general higher) values of weight decay are preferred by different activation functions. Then we compare performance under their respective optimal weight decay settings across a wide range of model capacities on three datasets. Surprisingly, the results in Figure~\ref{fig:abl_act} (d, e, f) suggest, under baseline AT, that smooth activation functions do not improve performance over ReLU in most cases, which contrasts with the prevailing consensus.

To verify the generality of our observations, we consider advanced adversarial training as described in \S\ref{sec:preliminaries} and repeat the experiment on CIFAR-10. Now we observe from Figure~\ref{fig:abl_act} (g, h, i) that smooth activation functions, particularly SiLU/Swish, start to provide meaningful improvements over non-smooth activation (i.e., ReLU) under advanced adversarial training. 

To summarize, our empirical findings provide further context to understand the AT conditions under which models with smooth activation functions outperform ReLU and vice-versa. First, we showed that the adversarial robustness of models with smooth activation functions is sensitive to AT hyperparameters, where removing BatchNorm affine parameters from weight decay is crucial. Then, we support the prevailing consensus by reaffirming that SiLU/Swish outperforms ReLU under advanced AT. At the same time, we demonstrate that ReLU outperforms smooth activation functions in most cases (i.e., different combinations of datasets and model capacities) under baseline AT.

\vspace{2pt}
\noindent\textbf{Normalization:} We find that standard \emph{BatchNorm outperforms other alternatives} such as GroupNorm \cite{wu2018group}, LayerNorm \cite{ba2016layer}, and InstanceNorm \cite{ulyanov2016instance}. We refer the readers to Appendix \S\ref{sec:app_normalization} for details and results.

\subsection{Impact of Network-level Design \label{sec:scale}}
Architectural design at the network level involves controlling the width and depth. Huang et al. \cite{huang2021exploring} presented an initial study showing the importance of network-level architectural design for adversarial robustness, from which we draw inspiration. However, despite demonstrating the utility of scaling only width or depth for adversarial robustness, Huang et al.'s attempt to identify a compound scaling rule to simultaneously scale depth and width was unsuccessful. Nevertheless, we hypothesize the existence of a compound scaling that is more effective than independent scaling by depth or width.

We re-visit network-level scaling from a two-objective perspective of maximizing adversarial robustness and network efficiency. Specifically, we seek to design an algorithm to identify an effective scaling rule for a given complexity measure, e.g., FLOPs, number of parameters, etc. First, we study the impact of these two scaling factors independently, followed by the interplay between them. Then, building upon the insights derived from these studies, we present a compound scaling rule, dubbed \emph{\ourscale{}}, to efficiently and effectively scale depth and width simultaneously for improving adversarial robustness. While \emph{\ourscale{}} is agnostic to any complexity measure, as an illustration, we consider minimizing FLOPs to improve network efficiency. A preview of \ourscale{} is provided below.

% -------------------------------------------------------------------------------------
\begin{figure}
\begin{tcolorbox}[title=Summary of Compound Scaling (\emph{\ourscale{}})]
    \noindent\emph{-- Ratio between Depth and Width:} {\footnotesize$\sum D_i:\sum W_i = 7:3$} such that {\footnotesize$^{\#}$}FLOPs{\footnotesize$\big(\sum D_i, \sum W_i\big) \approx$} target (\S\ref{sec:compund_scale}).
    
    \vspace{2pt}
    \noindent\emph{-- Distribution of Depth/Width among stages:} {\footnotesize$D_1:D_2:D_3=2:2:1$},  {\footnotesize$W_1:W_2:W_3=2:2.5:1$} (\S\ref{sec:independent_scale}).
    \vspace{-1.5em}
    \begin{center}
    \resizebox{.85\textwidth}{!}{%
    \begin{tabular}{@{\hspace{2mm}}lc|c|cc|cc|cc@{\hspace{2mm}}}
    \toprule
    \multicolumn{2}{l|}{\multirow{2}{*}{\begin{tabular}[c]{@{}l@{}}Desired\\ $^{\#}$FLOPs\end{tabular}}} & \multirow{2}{*}{{\begin{tabular}[c]{@{}c@{}}Referred\\as\end{tabular}}} & \multicolumn{2}{c|}{Stage 1} & \multicolumn{2}{c|}{Stage 2} & \multicolumn{2}{c}{Stage 3} \\
     &  &  & $D_1$ & $W_1$ & $D_2$ & $W_2$ & $D_3$ & $W_3$ \\ \midrule
    \parbox[t]{2mm}{\multirow{4}{*}{\rotatebox[origin=c]{90}{{\footnotesize\emph{\ourscale{}}}}}} & 5G & A1 & 14 & 5 & 14 & 7 & 7 & 3 \\
    & 10G & A2 & 17 & 7 & 17 & 9 & 8 & 4 \\
    & 20G & A3 & 22 & 8 & 22 & 11 & 11 & 5 \\
    & 40G & A4 & 27 & 10 & 28 & 14 & 13 & 6 \\ \bottomrule
    \end{tabular}%
    }\\
    {\includegraphics[width=0.85\textwidth]{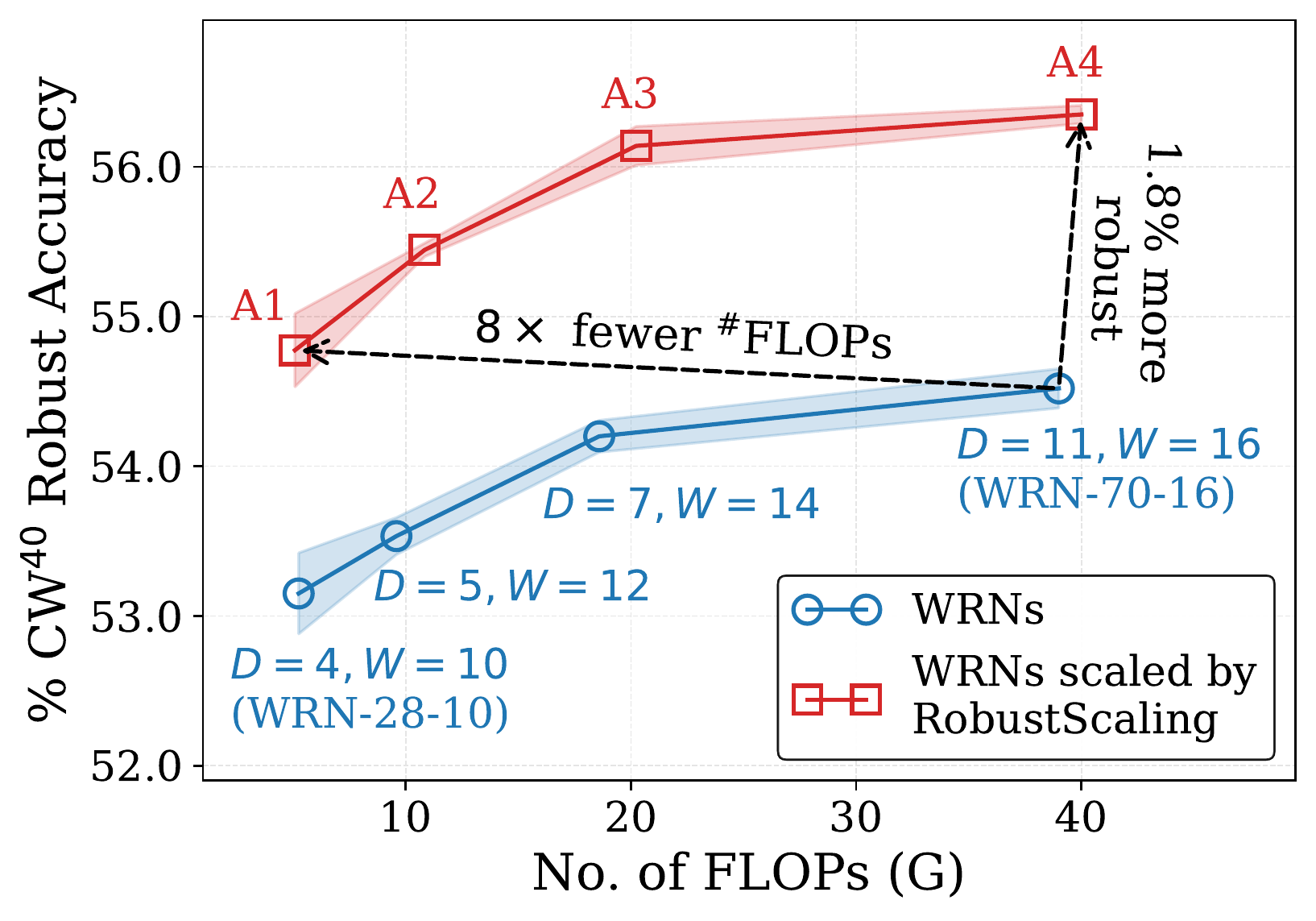}}
    \end{center}
    \vspace{-1.em}
    \noindent\textbf{\emph{-- Wide or Deep:}} For a targeted {\footnotesize$^{\#}$}FLOPs, \emph{deep (but narrow) networks are adversarially more robust than wide (but shallow) networks}.
\end{tcolorbox}
\end{figure}
% -------------------------------------------------------------------------------------

% -------------------------------------------------------------------------------------
\begin{figure*}[t]
    \begin{subfigure}[b]{0.245\textwidth}
    \centering
    \includegraphics[trim={0 3mm 0 0}, clip, width=.85\textwidth]{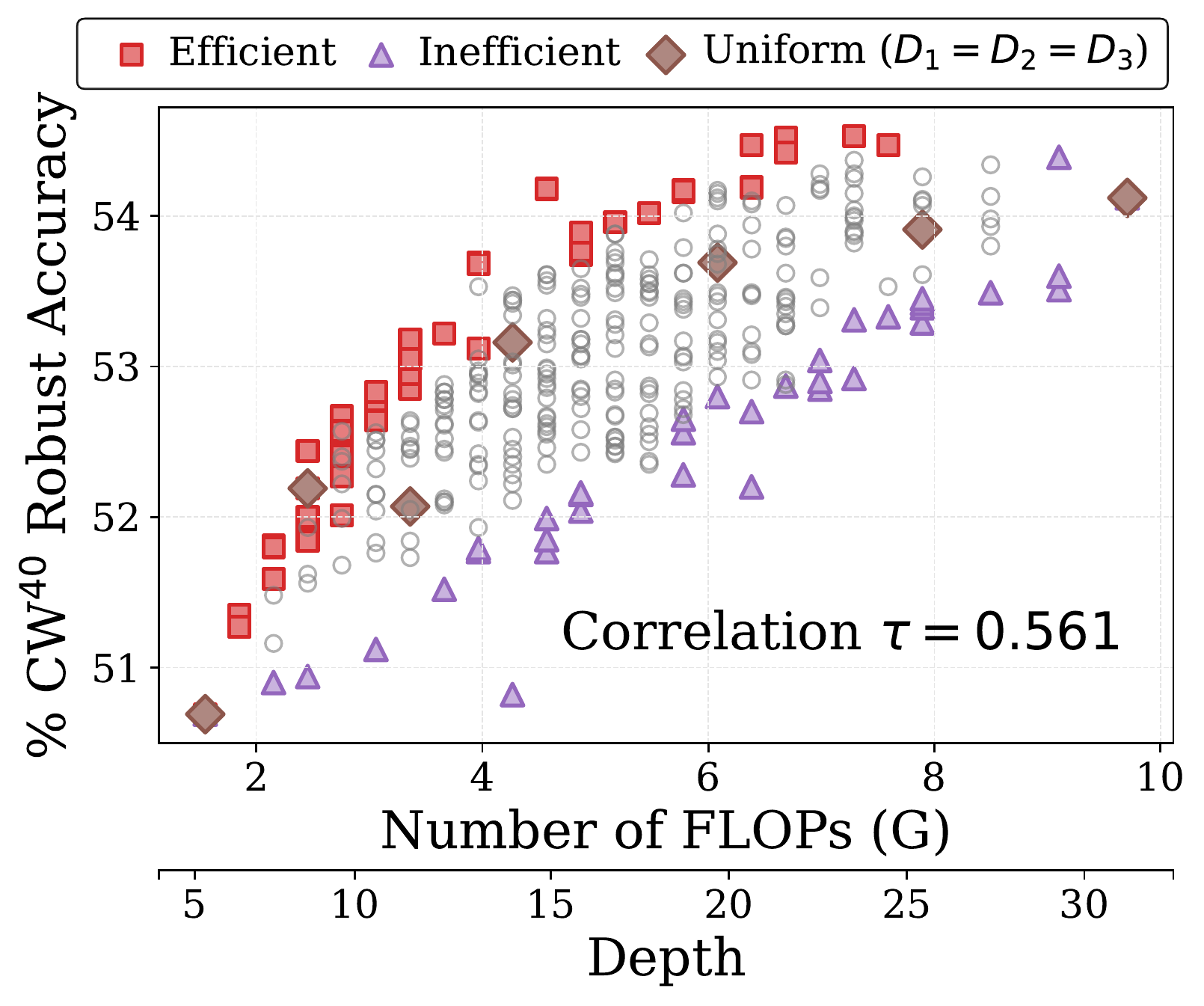}
    \caption{\scriptsize Depth vs. Robustness \label{fig:abl_c10_cw40_depth_flops_depth}}
    \end{subfigure} \hfill
    \begin{subfigure}[b]{0.245\textwidth}
    \centering
    \includegraphics[trim={0 0 0 0}, clip,width=\textwidth]{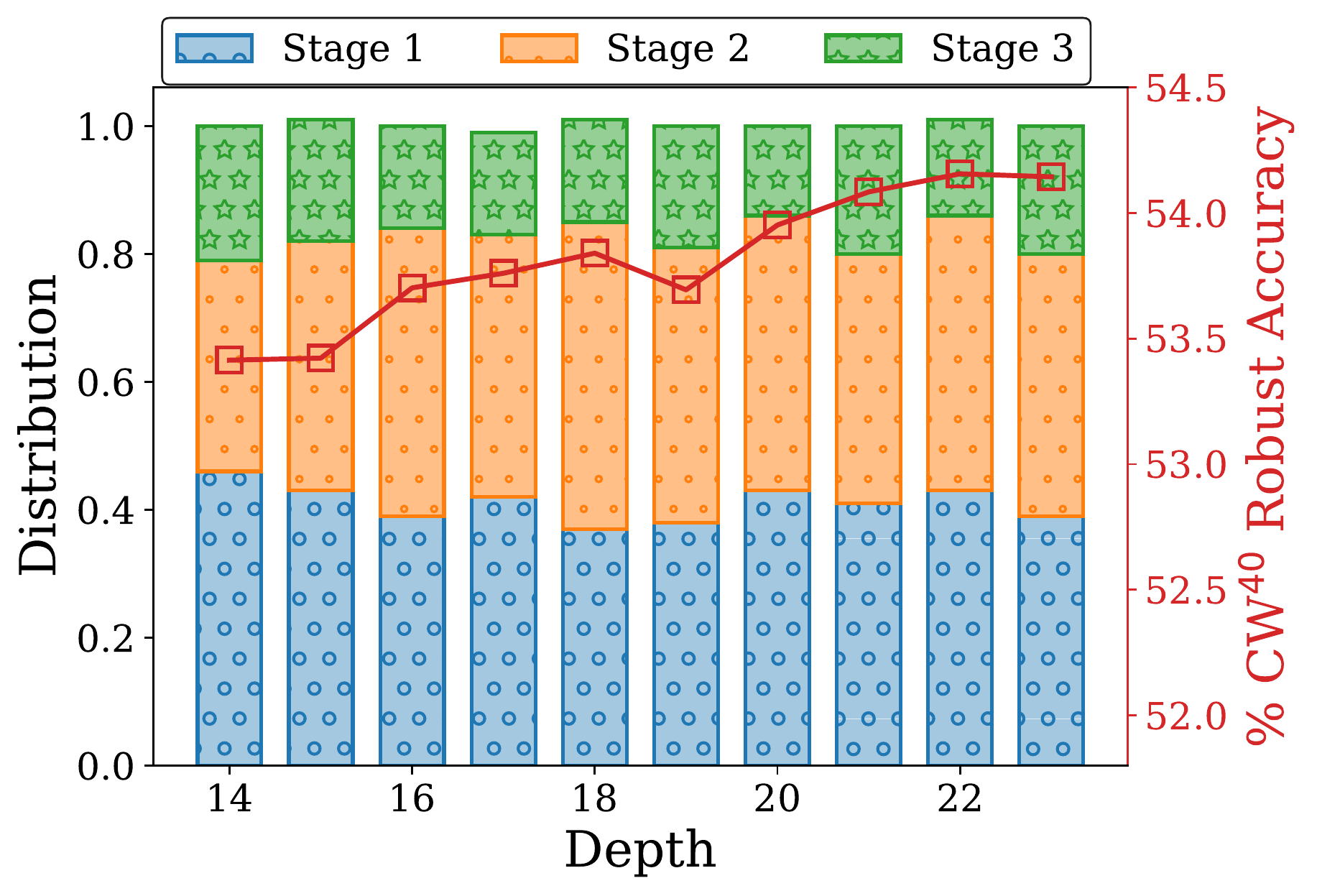}
    \caption{\scriptsize Top-Ranked Networks \label{fig:abl_c10_cw40_depth_distribution_top5}}
    \end{subfigure} \hfill
    \begin{subfigure}[b]{0.245\textwidth}
    \centering
    \includegraphics[trim={0 0 0 0}, clip,width=\textwidth]{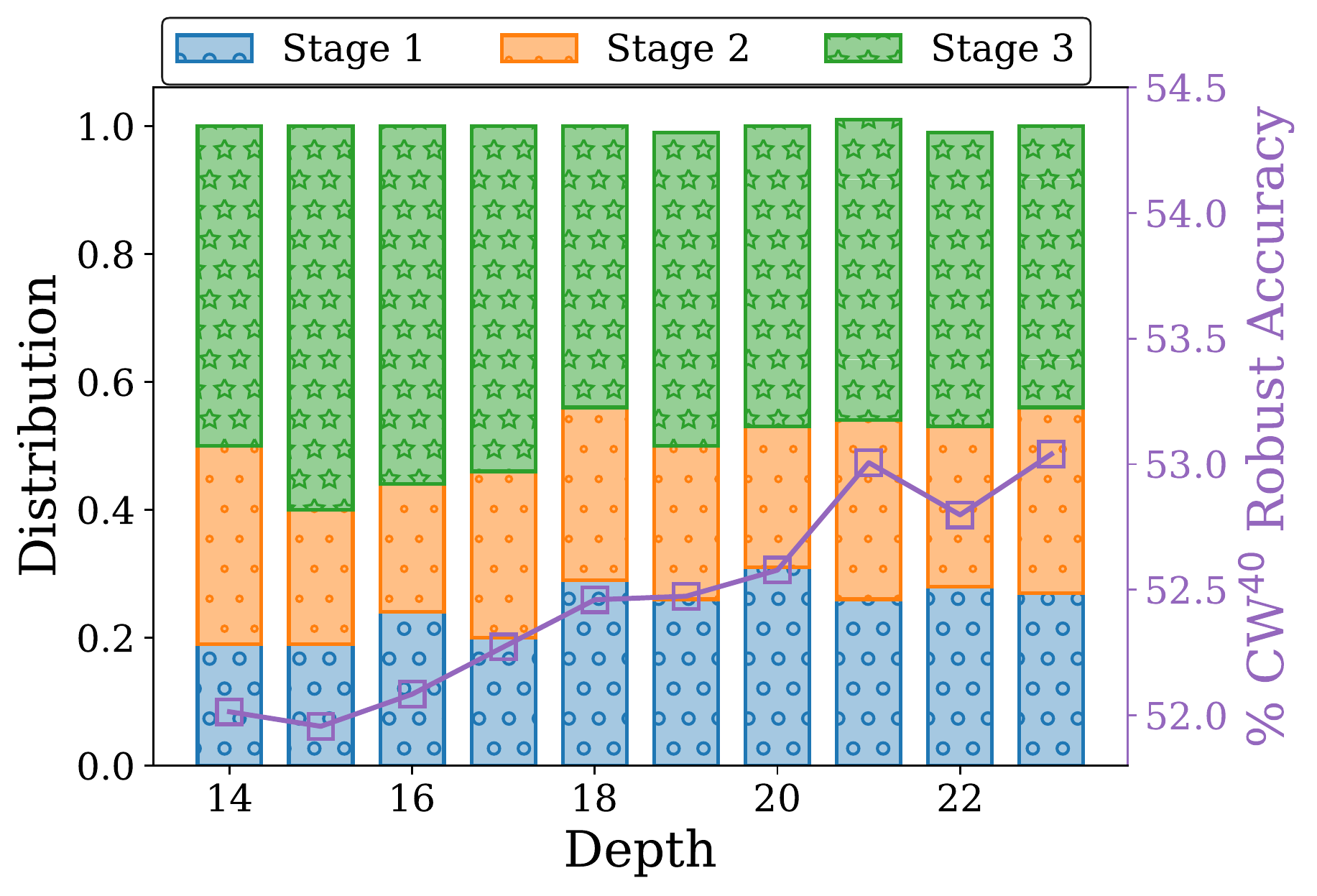}
    \caption{\scriptsize Last-Ranked Networks \label{fig:abl_c10_cw40_depth_distribution_last5}}
    \end{subfigure} \hfill
    \begin{subfigure}[b]{0.245\textwidth}
    \centering
    \includegraphics[trim={0 0 0 0}, clip,width=\textwidth]{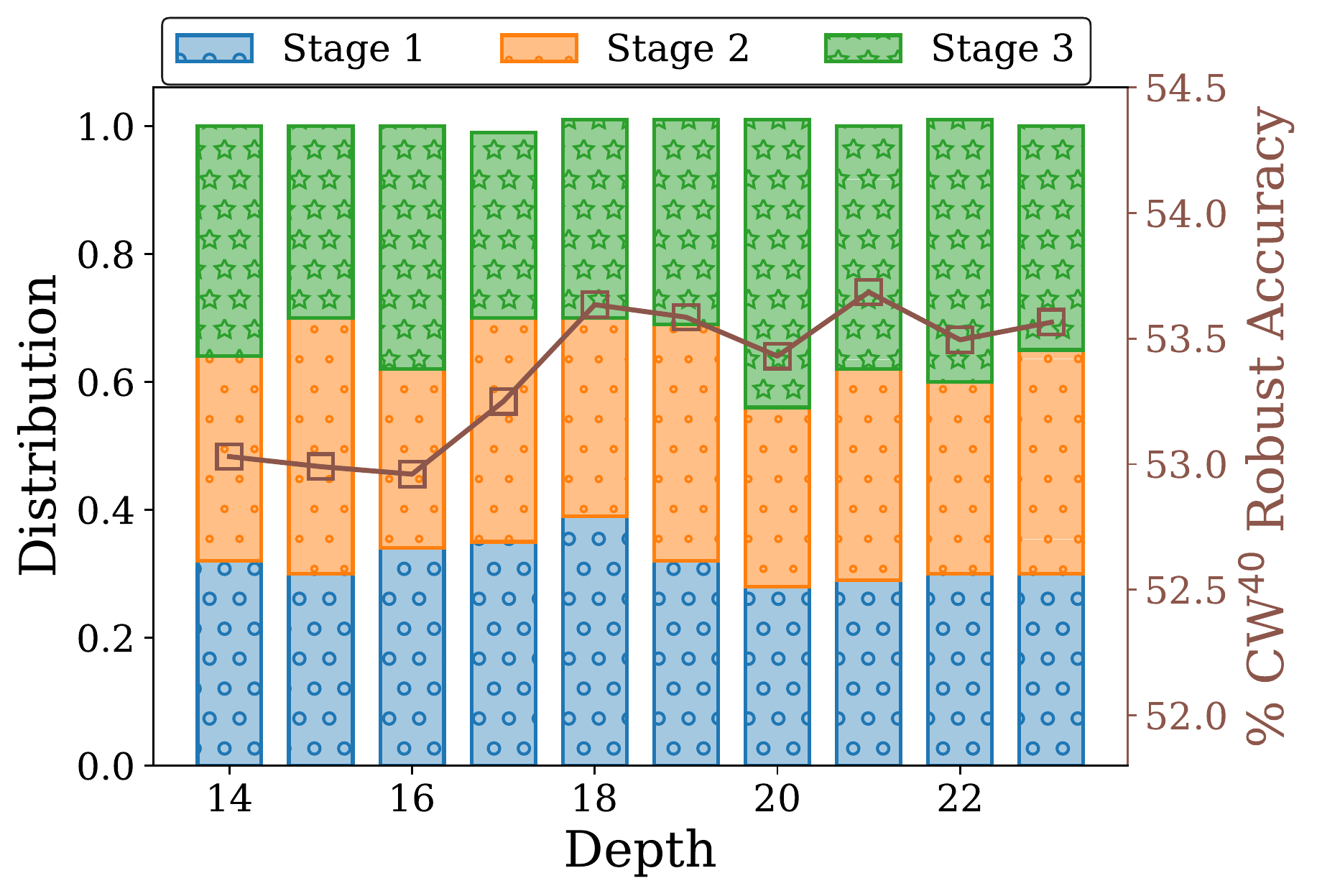}
    \caption{\scriptsize Standard Uniform Scaling \label{fig:abl_c10_cw40_depth_distribution_uniform}}
    \end{subfigure} 
    \begin{subfigure}[b]{0.245\textwidth}
    \centering
    \includegraphics[trim={0 3mm 0 0}, clip, width=.85\textwidth]{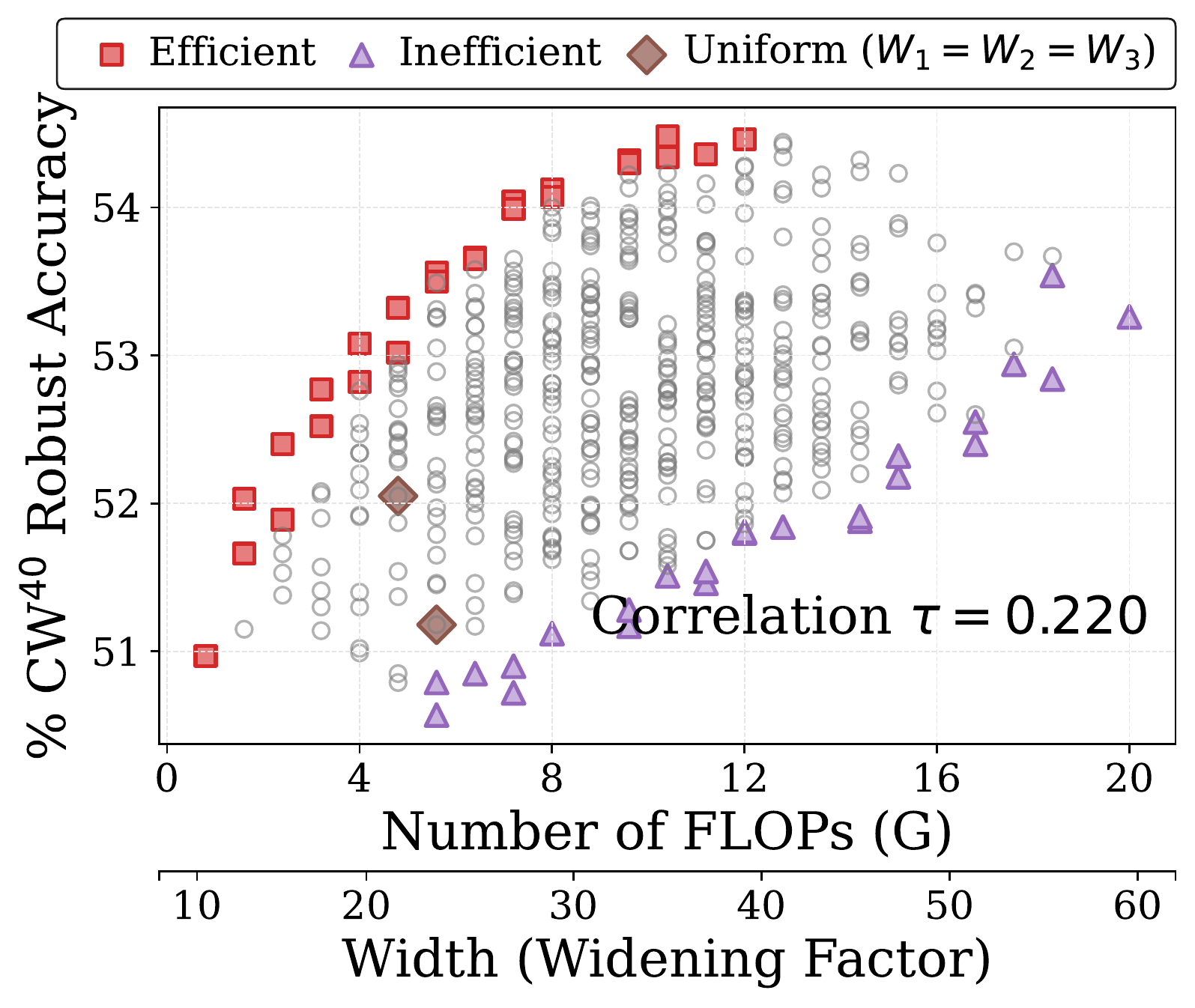}
    \caption{\scriptsize Width vs. Robustness \label{fig:abl_c10_cw40_channel_flops_width}}
    \end{subfigure} \hfill
    \begin{subfigure}[b]{0.245\textwidth}
    \centering
    \includegraphics[trim={0 0 0 0}, clip,width=\textwidth]{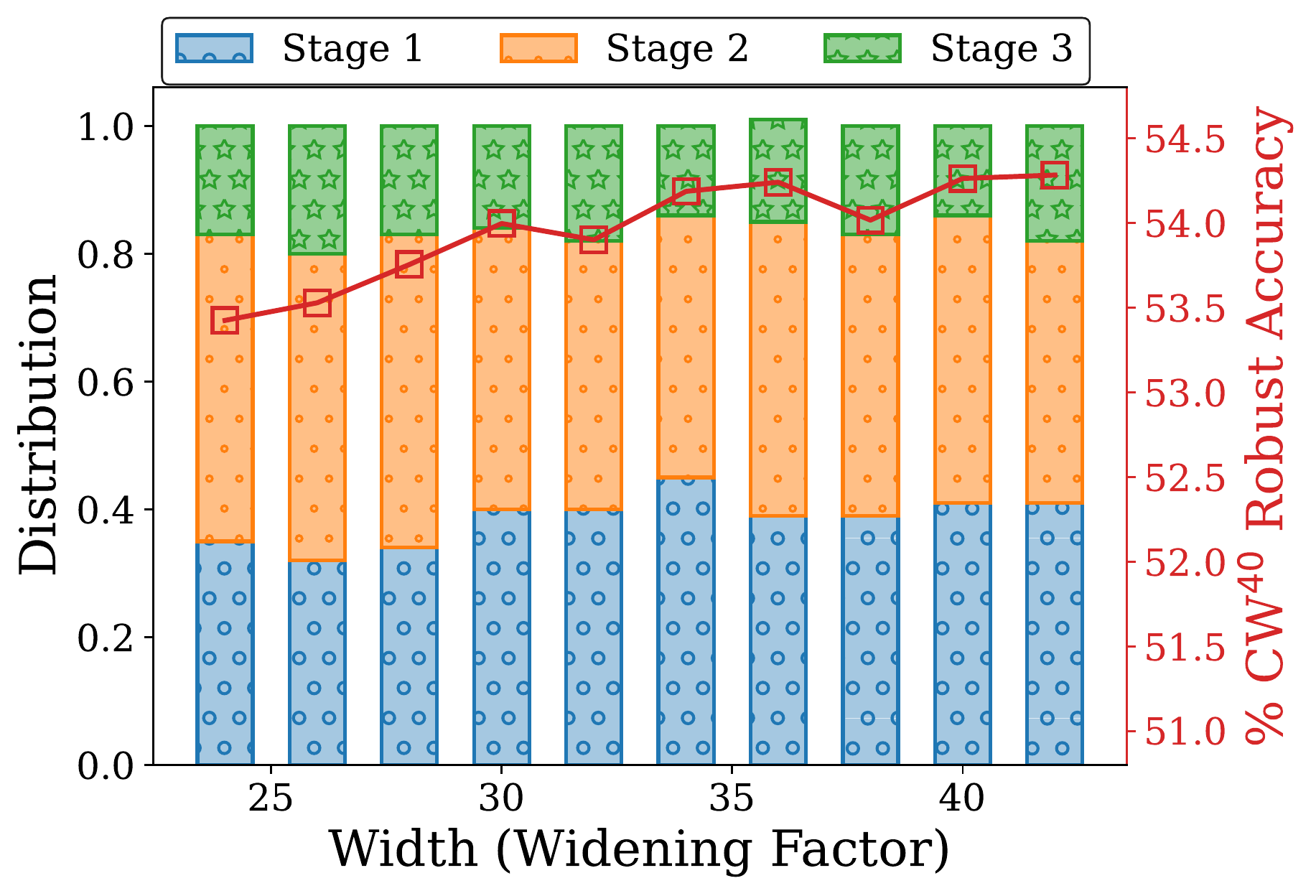}
    \caption{\scriptsize Top-Ranked Networks \label{fig:abl_c10_cw40_widen_distribution_top5}}
    \end{subfigure} \hfill
    \begin{subfigure}[b]{0.245\textwidth}
    \centering
    \includegraphics[trim={0 0 0 0}, clip,width=\textwidth]{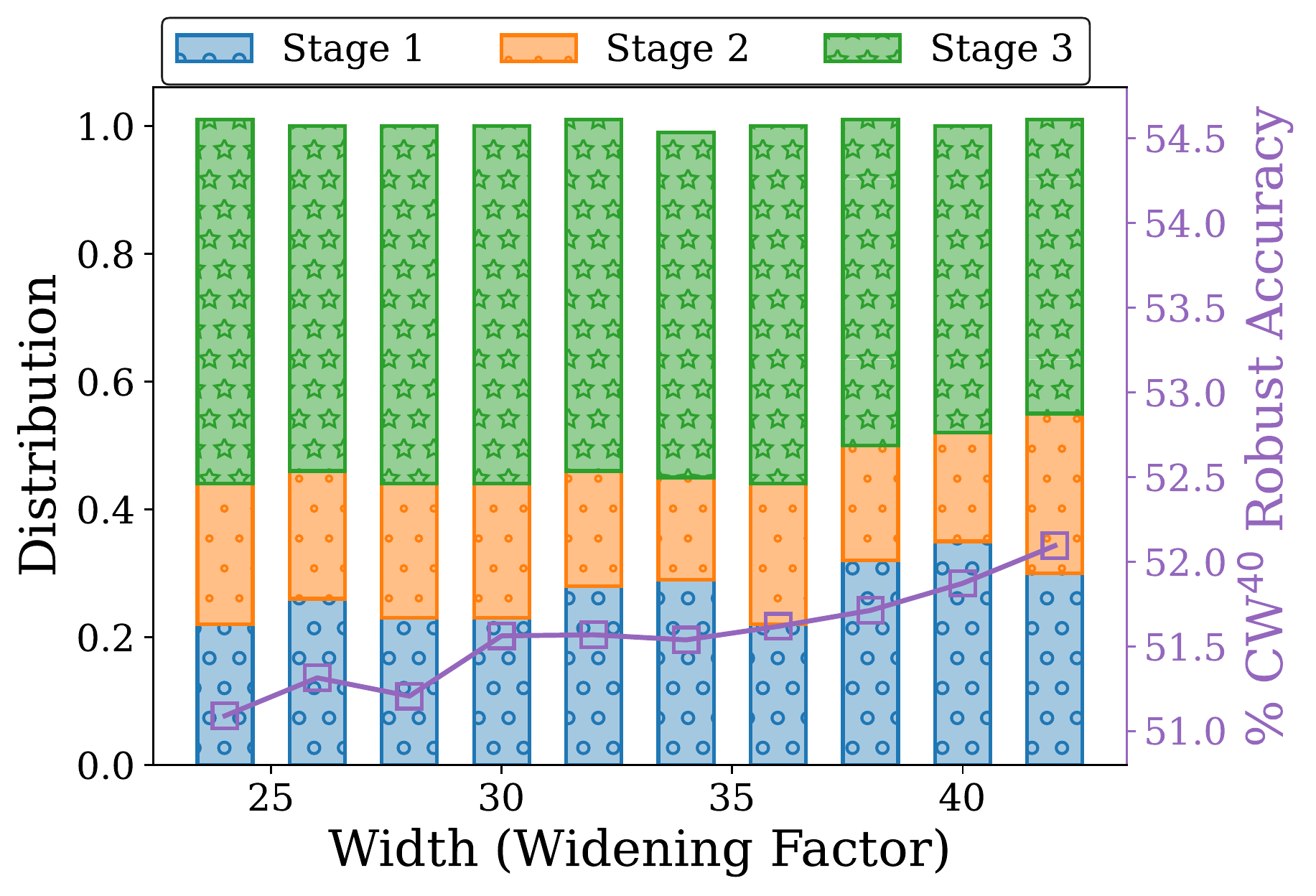}
    \caption{\scriptsize Last-Ranked Networks \label{fig:abl_c10_cw40_widen_distribution_last5}}
    \end{subfigure} \hfill
    \begin{subfigure}[b]{0.245\textwidth}
    \centering
    \includegraphics[trim={0 0 0 0}, clip,width=\textwidth]{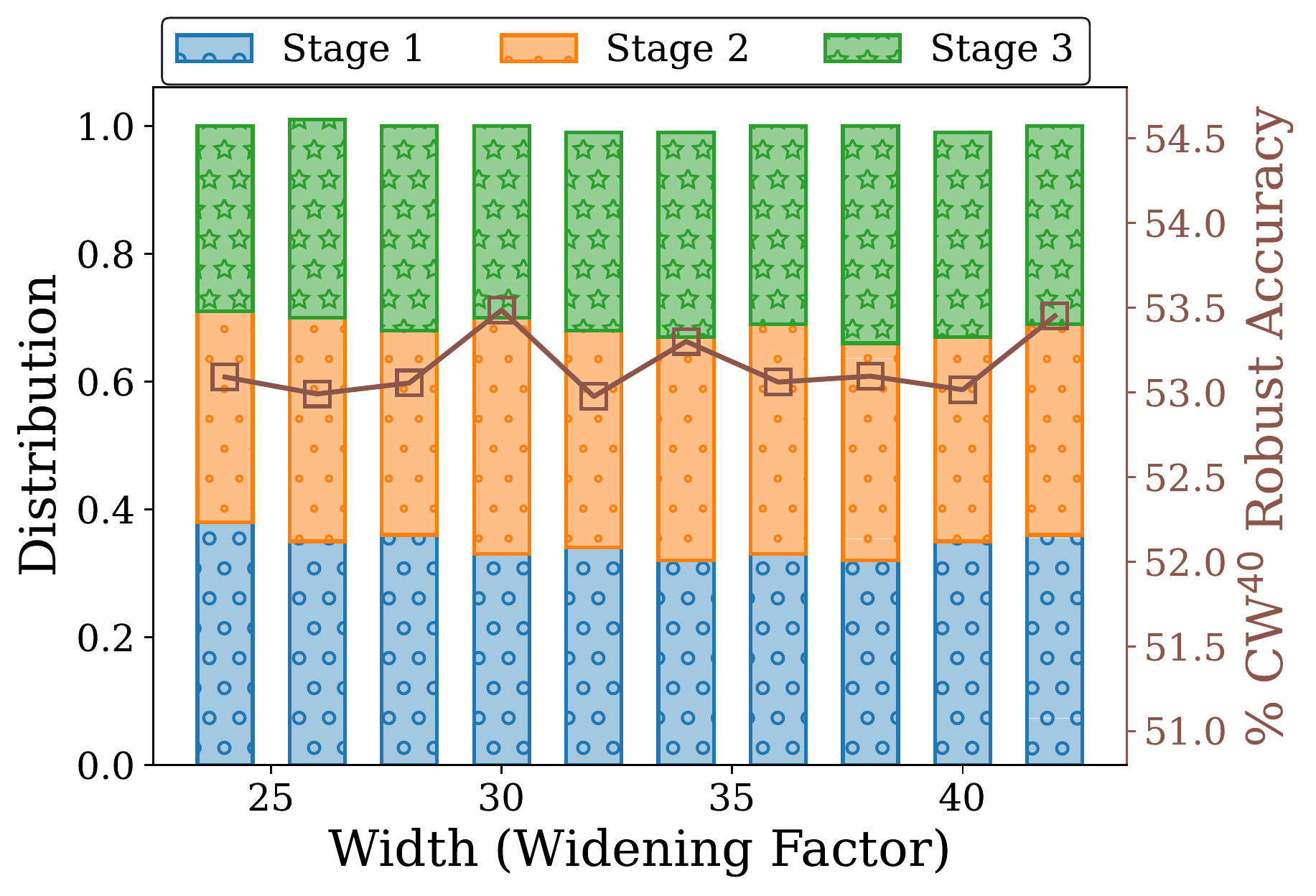}
    \caption{\scriptsize Standard Uniform Scaling \label{fig:abl_c10_cw40_widen_distribution_uniform}}
    \end{subfigure} 
    \caption{Adversarial robustness of networks with (a) 343 depth and (e) 512 width settings on CIFAR-10. \emph{Pareto-efficient} models (robust and compact) are in \textcolor{red}{red squares}, \emph{inefficient} models (sensitive and complex) are in \textcolor{violet}{violet triangles}, and networks with standard \emph{uniform} scaling ({\footnotesize$D_1=D_2=D_3$} and {\footnotesize$W_1=W_2=W_3$}) are in \textcolor{brown}{brown diamonds}. Rank correlation (Kendall $\tau$) between depth/width and robust accuracy is annotated. Distribution among the three stages for models with the efficient (b, f), standard uniform (c, g), and inefficient (d, h) distribution of depth and width are visualized, where the secondary y-axis with color corresponds to robust accuracy.\label{fig:abl_c10_cw40_depth_channel}\vspace{-10pt}}
\end{figure*}

\subsubsection{Independent Scaling by Depth or Width \label{sec:independent_scale}}

We independently study the relationship between adversarial robustness and network depth (i.e., number of blocks) or width in terms of widening factors (i.e., number of channels). We allow the depth of each stage ({\footnotesize$D_{i\in\{1,2,3\}}$}) to vary among $\{2, 3, 4, 5, 7, 9, 11\}$, and the width widening factor ({\footnotesize$W_{i\in\{1,2,3\}}$}) to vary among $\{4, 6, 8, 10, 12, 14, 16, 20\}$, while fixing the other architecture components to the baseline settings described in \S\ref{sec:preliminaries}. As a result, the number of layers in the resulting networks ranges from 16 to 70 in the case of depth variations. We adversarially train all possible networks (i.e., $7^3=343$ for depth and $8^3=512$ for width) using the baseline AT and present the results in Figs.~\ref{fig:abl_c10_cw40_depth_flops_depth} and~\ref{fig:abl_c10_cw40_channel_flops_width}, respectively. We highlight the networks, which we refer to as ``efficient", ``inefficient," and ``standard uniform depth/width" settings with different colored markers from a trade-off perspective of maximizing adversarial robustness and minimizing network complexity (FLOPs). Empirically, we observe that (i) there are no substantial correlations between network depth/width and adversarial robustness, implying that \emph{adding more blocks or channels does not automatically lead to better adversarial robustness}; and (ii) at any given computational budget, there is a significant variation in adversarial robustness, suggesting that \emph{the distribution of depth/width between the different stages needs to be carefully selected for improving adversarial robustness}.

Next, we perform a more detailed analysis of the depth/width distribution and robust accuracy of networks. At each level of total network depth/width, we rank the networks by their adversarial robustness and visualize the distribution of the number of blocks/widening factors among the three stages. We present the results in Fig.~\ref{fig:abl_c10_cw40_depth_channel} and make the following observations, (i) networks that distribute more blocks evenly between the first two stages and decrease the number of blocks in the third stage are {ranked at the top} (Fig.~\ref{fig:abl_c10_cw40_depth_distribution_top5}), (ii) networks that distribute more blocks in the third stage and reduce the number of blocks in the first two stages are ranked last (Fig.~\ref{fig:abl_c10_cw40_depth_distribution_last5}), (iii) top-ranked networks tend to use small widening factors in stage 3 and allocate larger widening factors to the first two stages, particularly the second stage (Fig.~\ref{fig:abl_c10_cw40_widen_distribution_top5}), and (iv) last-ranked networks use larger widening factors in the last stage by reducing the widening factors of the second stage (Fig.~\ref{fig:abl_c10_cw40_widen_distribution_last5}). 

After that, for both depth and width, by averaging the number of blocks/widening factor distribution in the top-ranked models across all levels of depth/width, we identify that distributing the depth, i.e., the number of layers, as {\footnotesize$D_1:D_2:D_3 = 2:2:1$} and width, in terms of widening factors, as {\footnotesize$W_1:W_2:W_3 = 2:2.5:1$} across the stages leads to robust and efficient models. Finally, for completeness, we also show the depth (Fig.~\ref{fig:abl_c10_cw40_depth_distribution_uniform}) / widening factor (Fig.~\ref{fig:abl_c10_cw40_widen_distribution_uniform}) distribution and robust accuracy for networks with the standard uniform depth/width settings.

\subsubsection{Compound Scaling by Depth and Width \label{sec:compund_scale} }
Leveraging the interplay among scaling factors (e.g., depth, width, etc.) is particularly effective in scaling networks under standard ERM training \cite{tan2019efficientnet}. However, under adversarial training, a prior attempt to identify such a compound scaling rule was not successful \cite{huang2021exploring}. To this end, we revisit the question, \emph{does an effective compound policy for scaling networks under adversarial training exist?} Specifically, we seek to identify a compound scaling policy to simultaneously adjust network depth and width by studying their interplay. Building upon the independent depth/width scaling rules specified in \S\ref{sec:independent_scale}, for a fixed computational complexity, compound scaling can be realized as a competition between network depth and width for resources.

% -------------------------------------------------------------------------------------
\begin{figure}[t]
    \begin{subfigure}[b]{0.235\textwidth}
    \centering
    \includegraphics[width=0.95\textwidth]{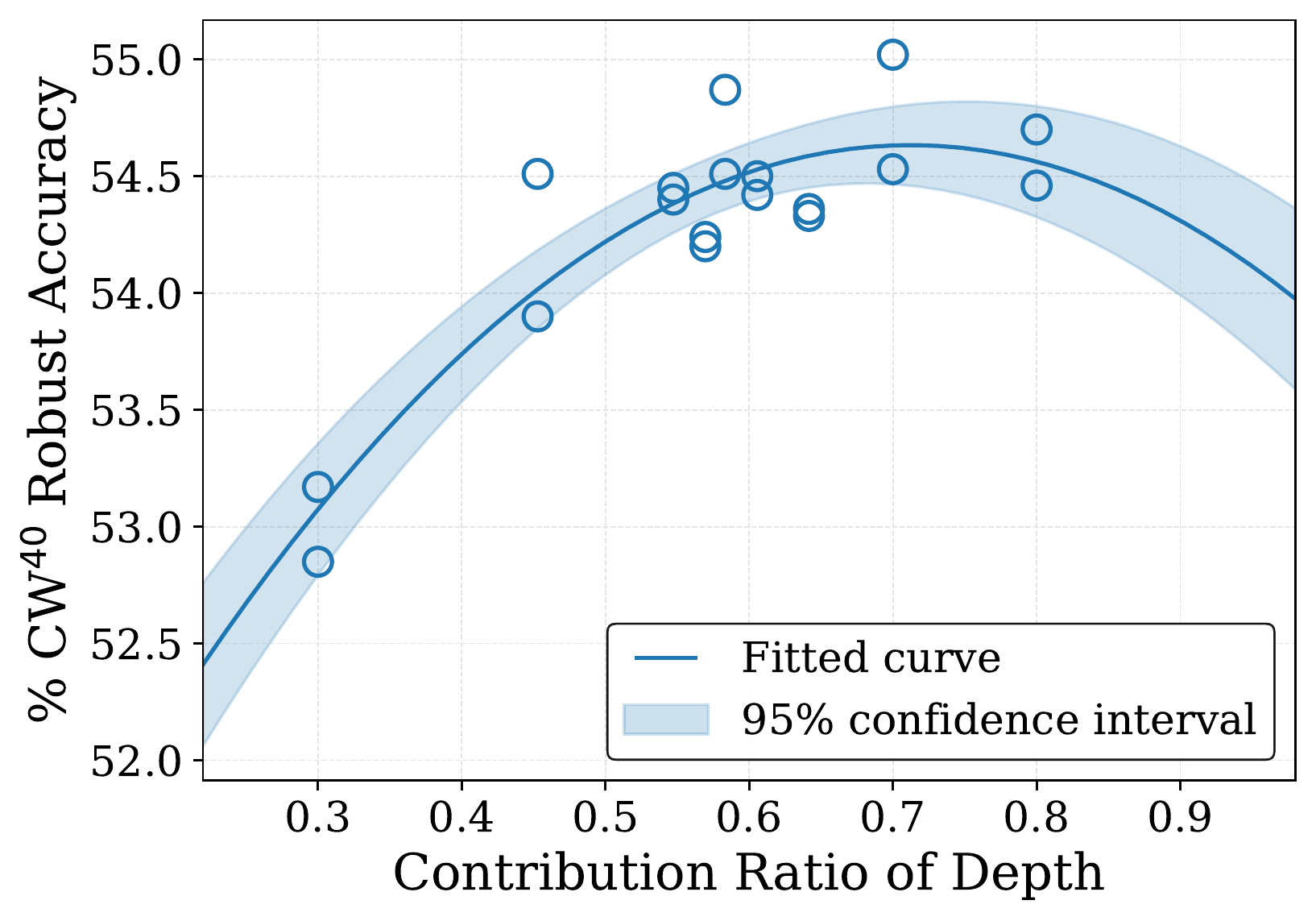}
    \caption{\scriptsize 5G FLOPs \label{fig:abl_compound_scale_depth_ratio_5g}}
    \end{subfigure} \hfill
    \begin{subfigure}[b]{0.235\textwidth}
    \centering
    \includegraphics[width=0.95\textwidth]{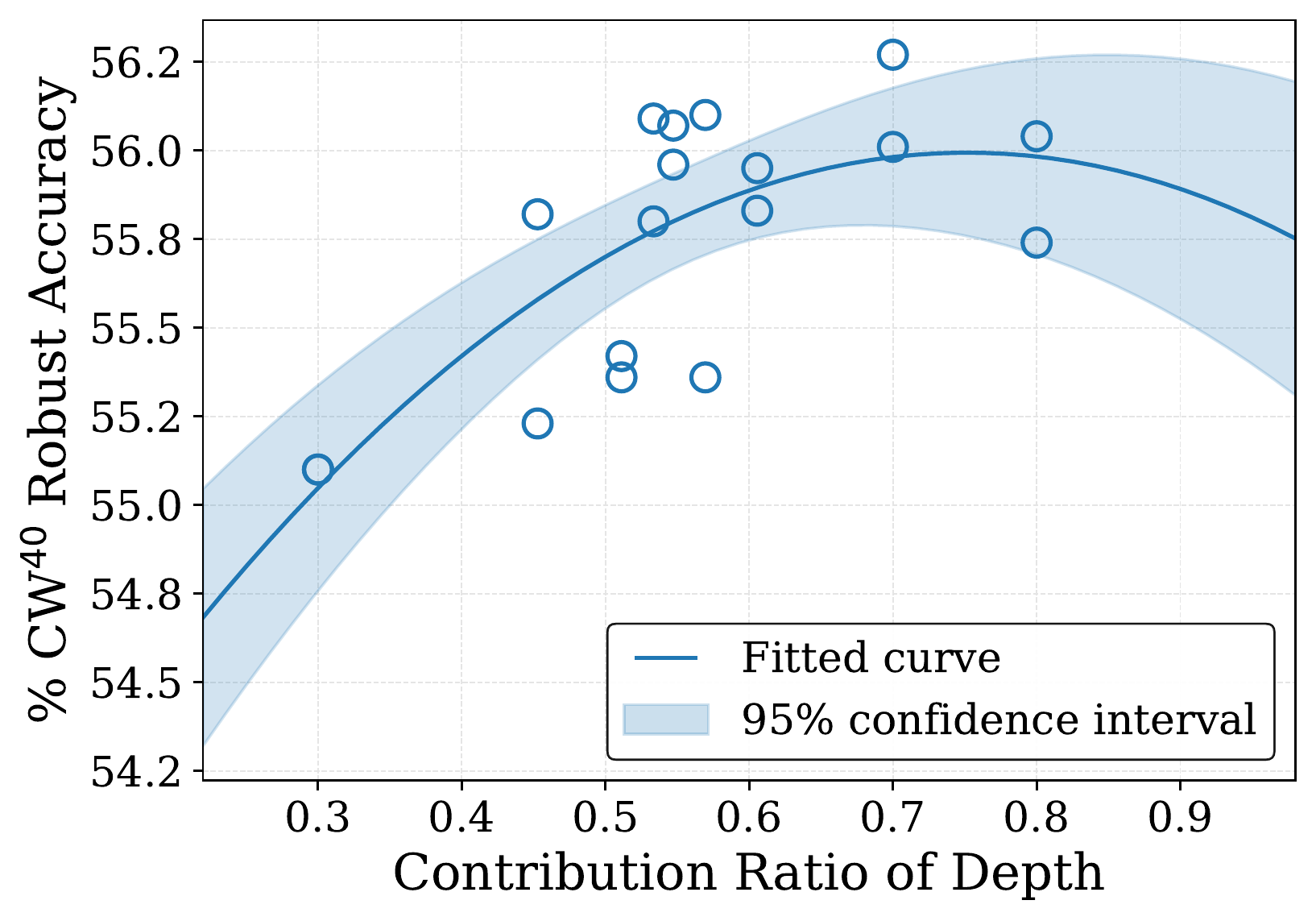}
    \caption{\scriptsize 20G FLOPs \label{fig:abl_compound_scale_depth_ratio_20g}}
    \end{subfigure}
    \caption{(a, b) Adversarial robustness vs. contribution ratio of depth ({\footnotesize $r_D$}) at different FLOP levels, where {\footnotesize$r_D=\sum D_i/(\sum D_i+\sum W_i)$}. A larger {\footnotesize $r_D$} indicates a deeper (more blocks) but narrower (fewer channels) network.\label{fig:abl_compound_scale}\vspace{0pt}}
\end{figure}
\begin{figure}[ht]
    \captionsetup[subfigure]{justification=centering}
    \begin{subfigure}[b]{0.235\textwidth}
    \centering
    \includegraphics[width=0.98\textwidth]{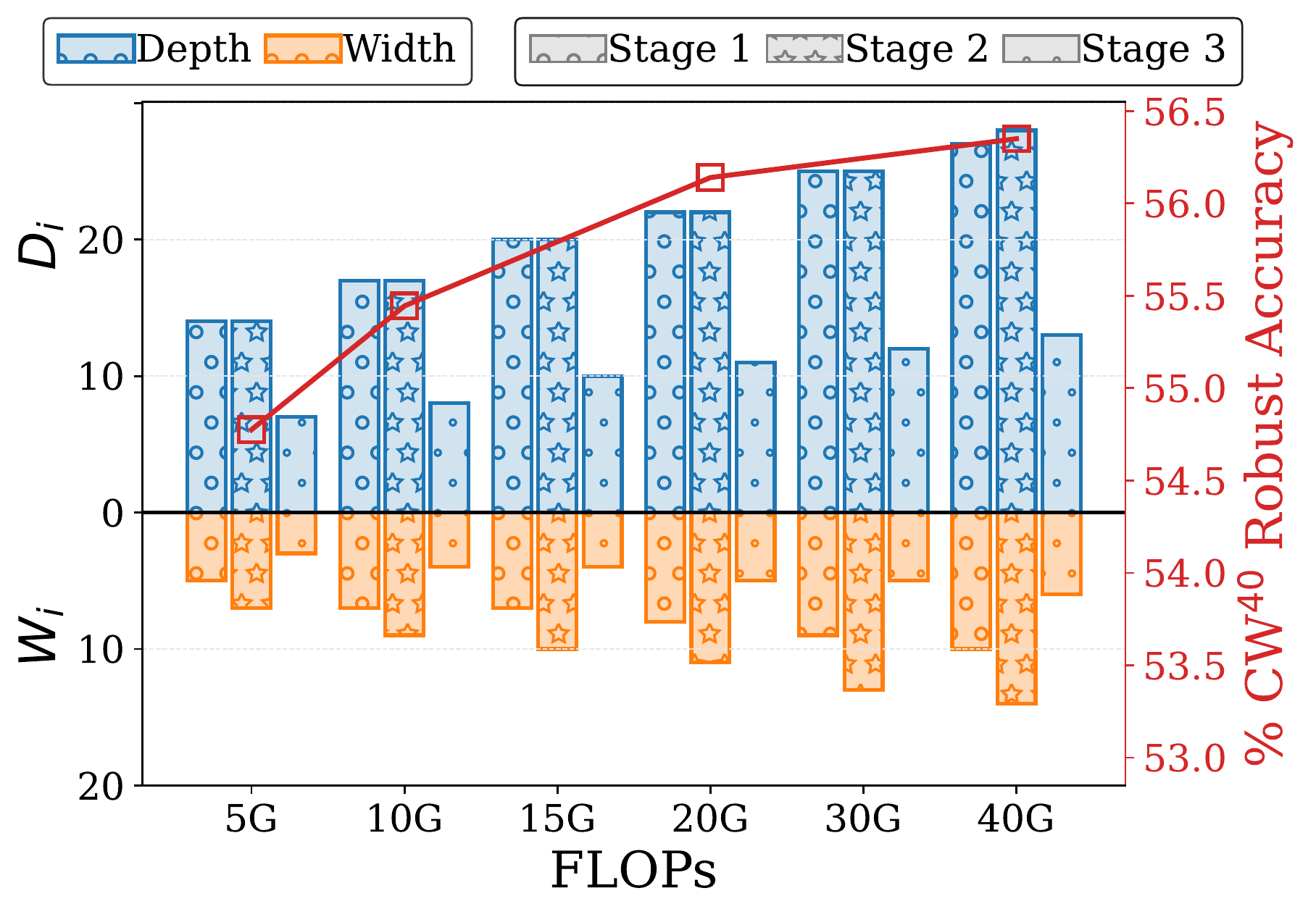}
    \vspace{-5pt}
    \caption{\scriptsize Compound scaling\\($\sum D_i:\sum W_i = 7:3$)\label{fig:abl_c10_cw40_compound_scaling}}
    \end{subfigure} \hfill
    \begin{subfigure}[b]{0.235\textwidth}
    \centering
    \includegraphics[width=0.98\textwidth]{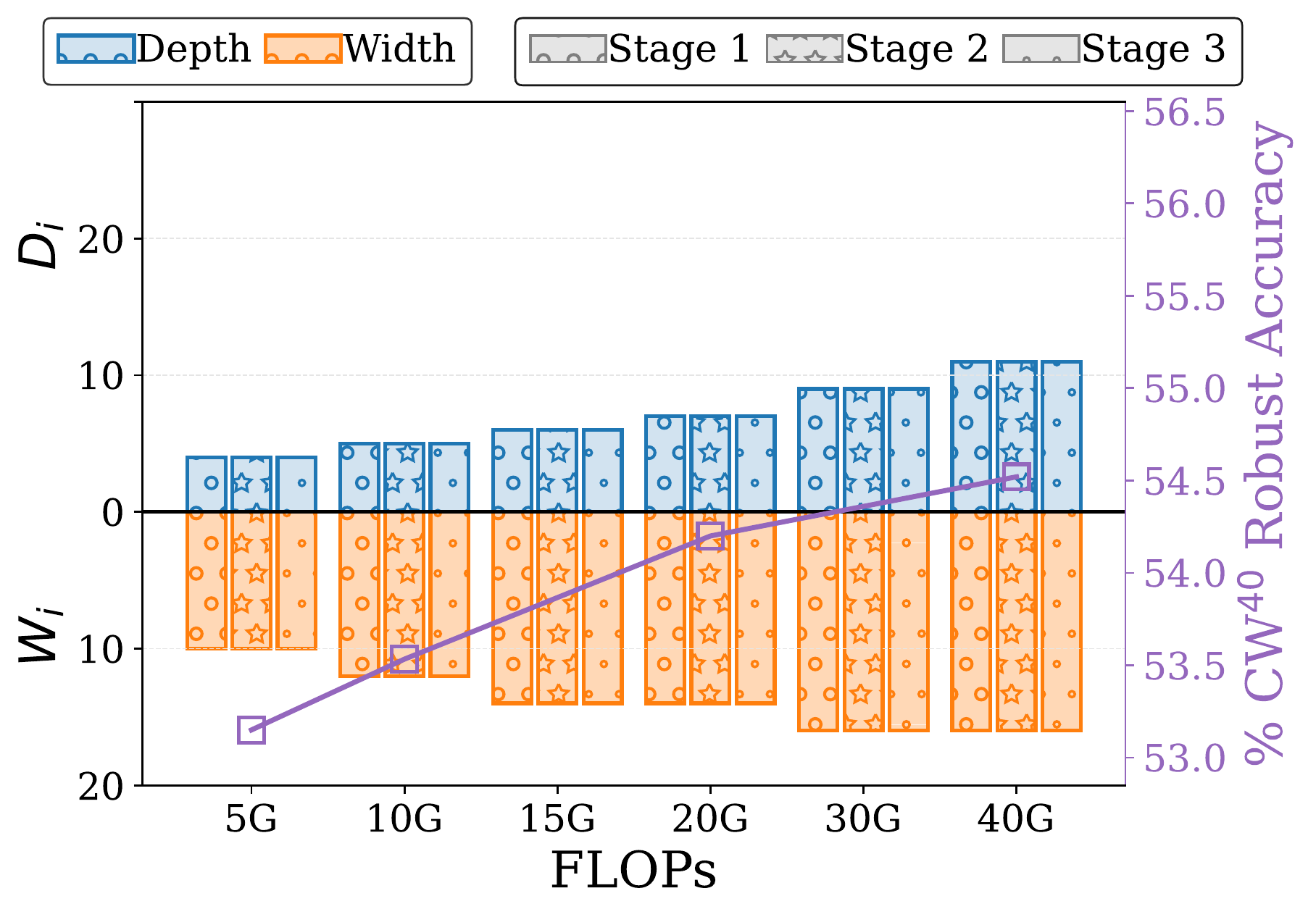}
    \vspace{-5pt}
    \caption{\scriptsize Standard scaling\\(i.e., WRN-28-10, WRN-70-16) \label{fig:abl_c10_cw40_standard_scaling}}
    \end{subfigure} \\
    \begin{subfigure}[b]{0.235\textwidth}
    \centering
    \includegraphics[width=0.98\textwidth]{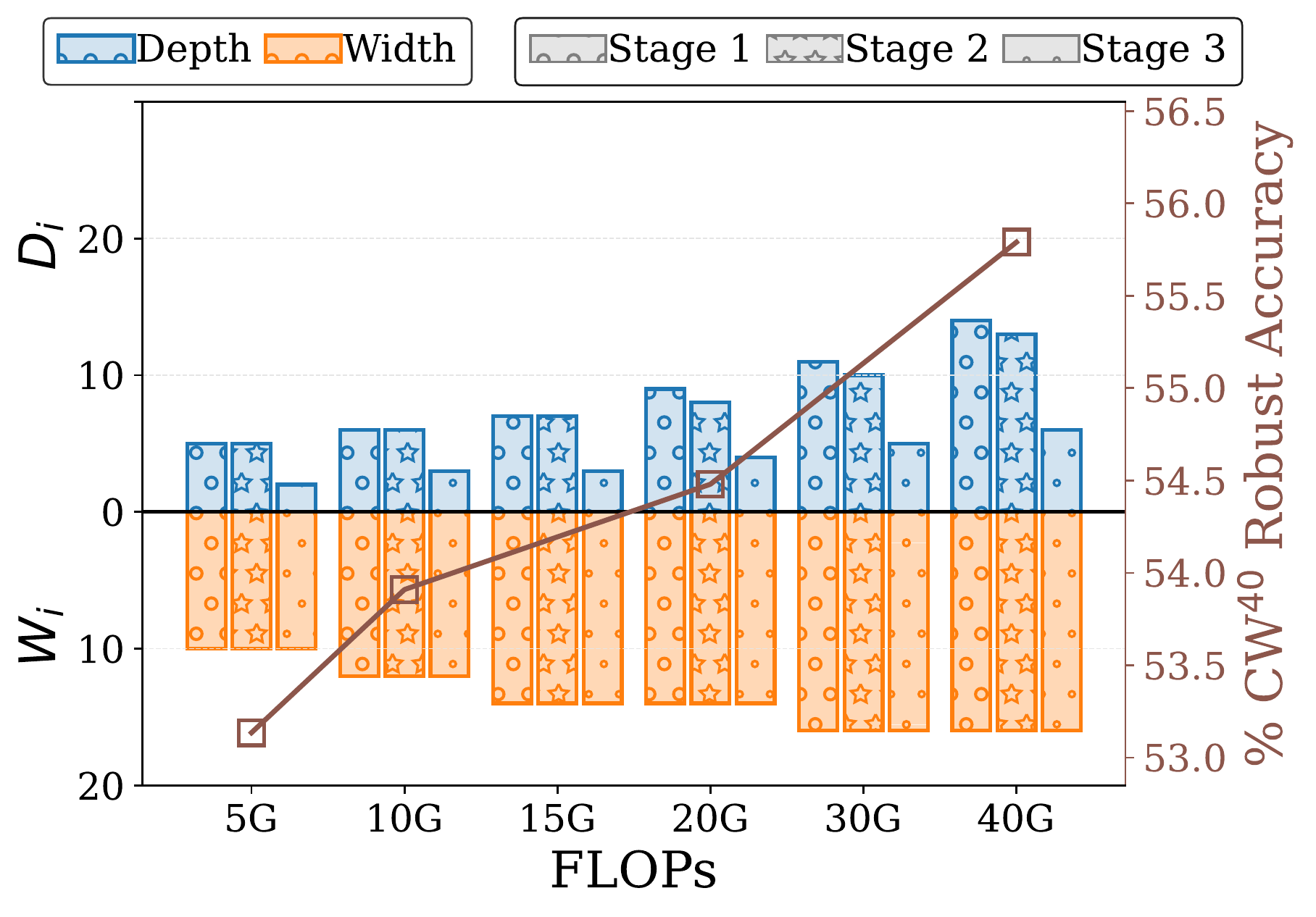}
    \vspace{-5pt}
    \caption{\scriptsize Scaling by depth\\($D_1:D_2:D_3 = 2:2:1$)\label{fig:abl_c10_cw40_depth_scaling}}
    \end{subfigure} \hfill
    \begin{subfigure}[b]{0.235\textwidth}
    \centering
    \includegraphics[width=0.98\textwidth]{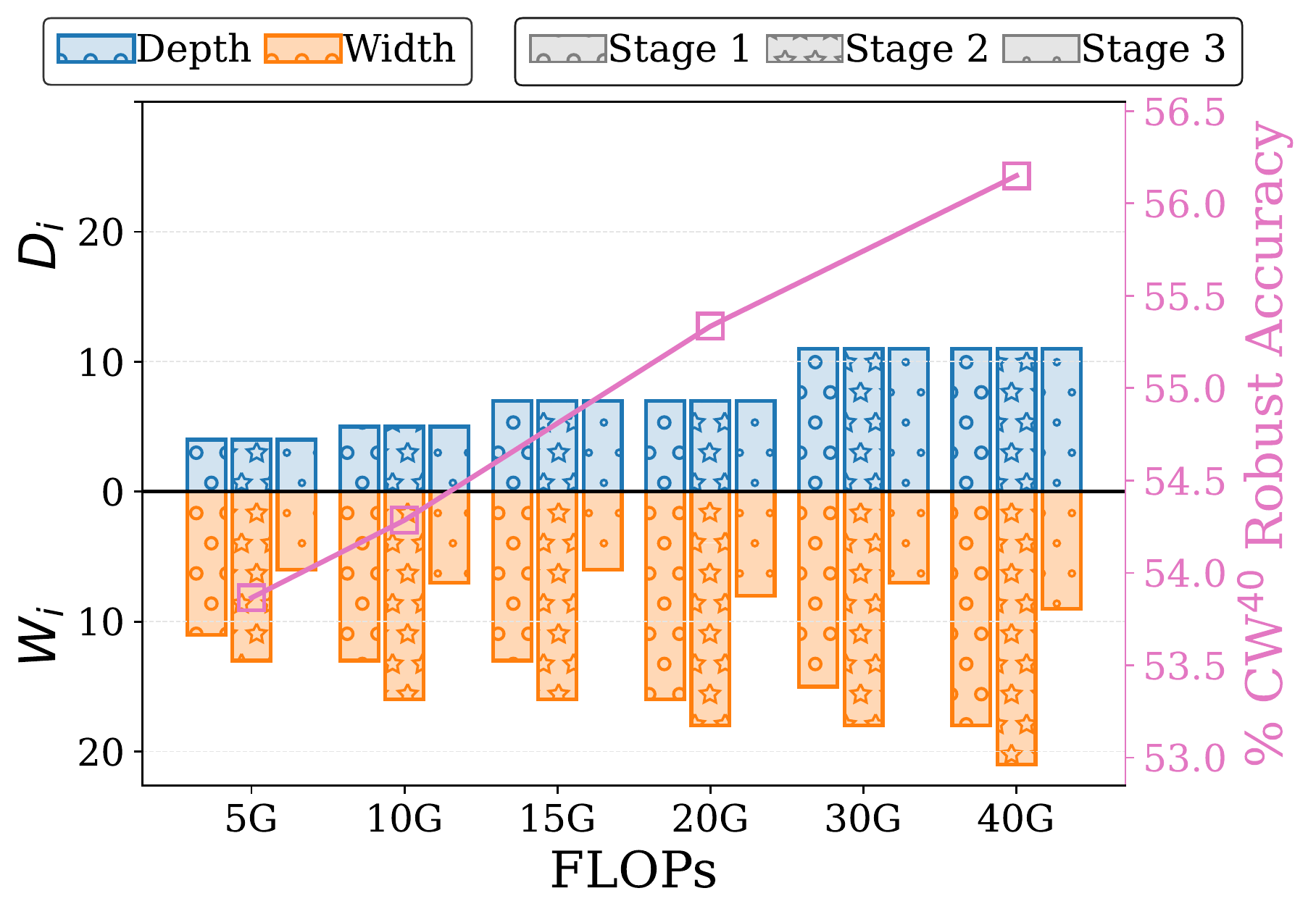}
    \vspace{-5pt}
    \caption{\scriptsize Scaling by width\\($W_1:W_2:W_3 = 2:2.5:1$)\label{fig:abl_c10_cw40_width_scaling}}
    \end{subfigure}
    \caption{Visualization of depth and width distribution among the three stages for (a) our compound scaling, (b) standard scaling, and (c, d) our independent scaling by depth/width. The secondary y-axis shows robust accuracy under baseline adversarial training.\label{fig:abl_compound_scale_visualization} \vspace{-5pt}}
\end{figure}

% -------------------------------------------------------------------------------------
We formulate our goal as searching for an appropriate ratio between total network depth and total network width (in terms of widening factors), i.e., [{\footnotesize$ \sum D_i:\sum W_i$}], to efficiently allocate computational resources while improving adversarial robustness. Given a target network complexity (e.g., FLOPs budget), we systematically tune the contribution ratio of depth (i.e., {\footnotesize$r_D=\sum D_i/(\sum D_i+\sum W_i)$}) between $[0.3, 0.95)$ and compare the relative changes in robustness under baseline AT. From the results shown in Figure~\ref{fig:abl_compound_scale}, we observe that adversarial robustness improves monotonically as $r_D$ increases and peaks at approximately {\footnotesize $r_D=0.7$}. However, as the {\footnotesize $r_D$} continues to increase beyond $0.7$, adversarial robustness starts to deteriorate rapidly. Accordingly, our compound scaling rule, dubbed \emph{\ourscale{}}, is obtained by solving: 
\begin{equation*} \label{eq:compound_scale}
\small
\begin{split}
r_D &= \frac{D_1 + D_2 + D_3}{D_1 + D_2 + D_3 + W_1 + W_2 + W_3} \\
    &= \frac{2D_3 + 2D_3 + D_3}{2D_3 + 2D_3 + D_3 + 2W_3 + 2.5W_3+W_3} = 0.7
\end{split}
\end{equation*}
\noindent such that the {\small$Complexity{\big(\sum D_i, \sum W_i\big) \approx}$} $\mbox{ the target}$.
A pictorial illustration of the compound settings under different FLOP budgets is provided in Figure~\ref{fig:abl_c10_cw40_compound_scaling}, along with the standard settings (i.e., WRN-28-10, WRN-70-16, etc.) in Figure~\ref{fig:abl_c10_cw40_standard_scaling}, the settings obtained by independently scaling depth and width in Figures~\ref{fig:abl_c10_cw40_depth_scaling} and \ref{fig:abl_c10_cw40_width_scaling}, respectively. 
We observe that \emph{deep but narrow networks are preferred over wide but shallow networks for adversarial robustness} at a given FLOPs budget.

% -------------------------------------------------------------------------------------
% -------------------------------------------------------------------------------------

\begin{figure}[t]
    \definecolor{royalblue}{RGB}{65,105,225}
    \definecolor{better}{RGB}{0, 165, 156}
    \definecolor{tab_blue}{RGB}{31, 119, 180}
    \definecolor{tab_orange}{RGB}{255, 127, 14}
    \definecolor{tab_green}{RGB}{44, 160, 44}
    \definecolor{tab_red}{RGB}{214, 39, 40}
    \definecolor{tab_brown}{RGB}{140, 86, 75}
    \begin{minipage}{0.48\textwidth}
    \begin{subfigure}[b]{0.495\textwidth}
    \centering
    \includegraphics[width=0.95\textwidth]{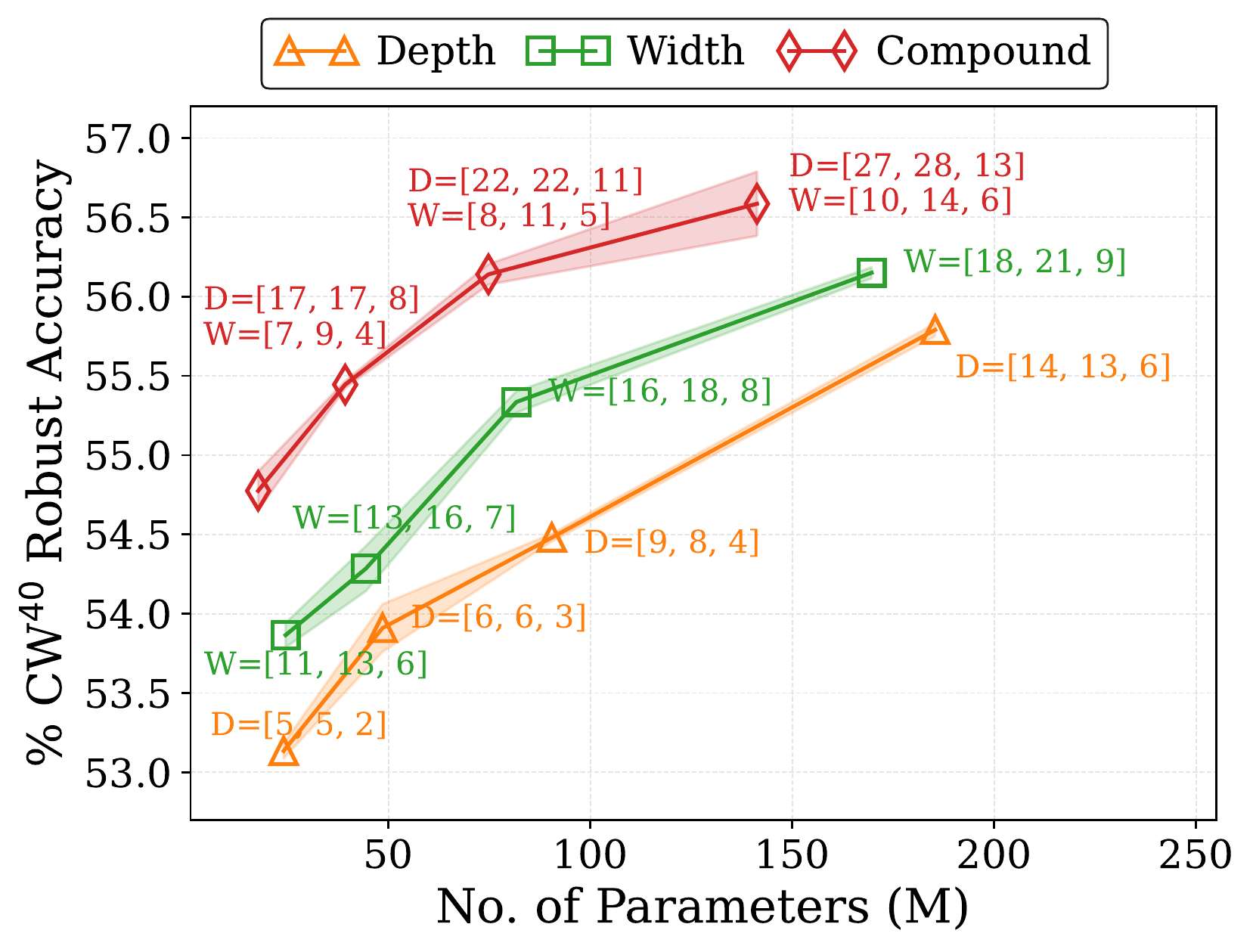}
    \caption{\scriptsize CW$^{40}$ vs. $^{\#}$Params  \label{fig:abl_c10_cw40_params_scale}}
    \end{subfigure}\hfill
    \begin{subfigure}[b]{0.495\textwidth}
    \centering
    \includegraphics[width=0.95\textwidth]{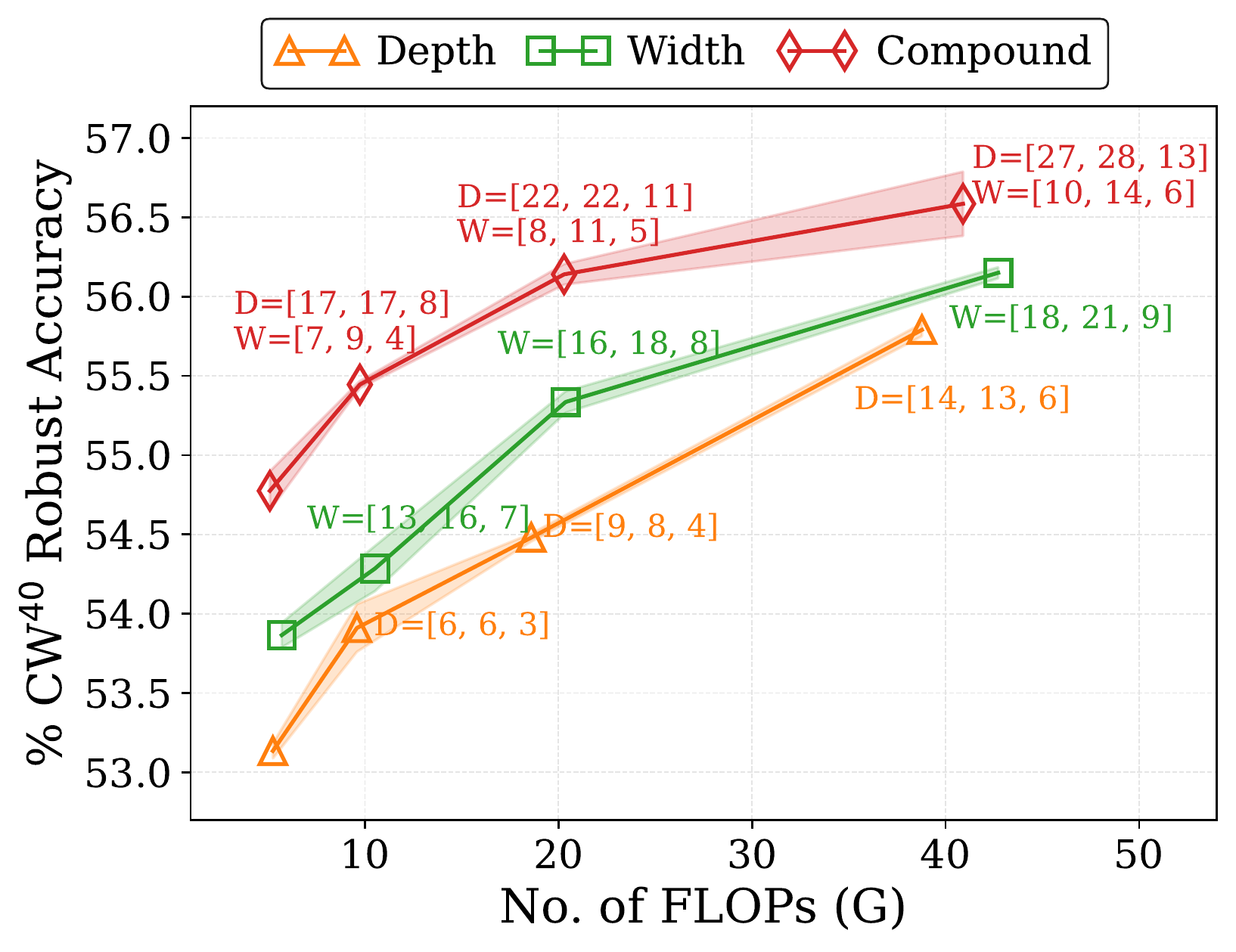}
    \caption{\scriptsize CW$^{40}$ vs. $^{\#}$FLOPs \label{fig:abl_c10_cw40_flops_scale}}
    \end{subfigure}\\
    \begin{subfigure}[b]{0.495\textwidth}
    \centering
    \includegraphics[width=0.95\textwidth]{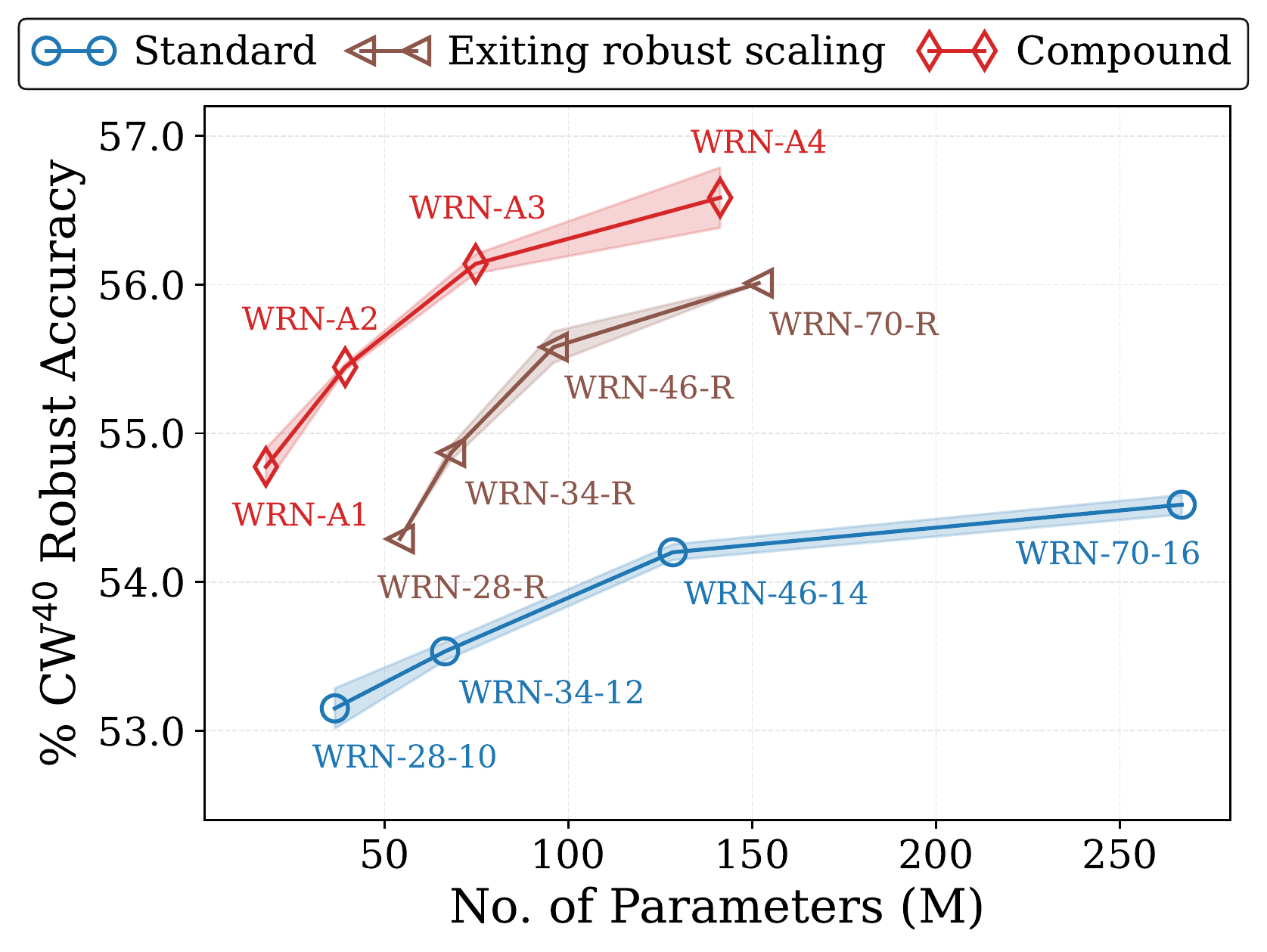}
    \caption{\scriptsize CW$^{40}$ vs. $^{\#}$Params  \label{fig:abl_c10_cw40_params_scale_compare}}
    \end{subfigure}\hfill
    \begin{subfigure}[b]{0.495\textwidth}
    \centering
    \includegraphics[width=0.95\textwidth]{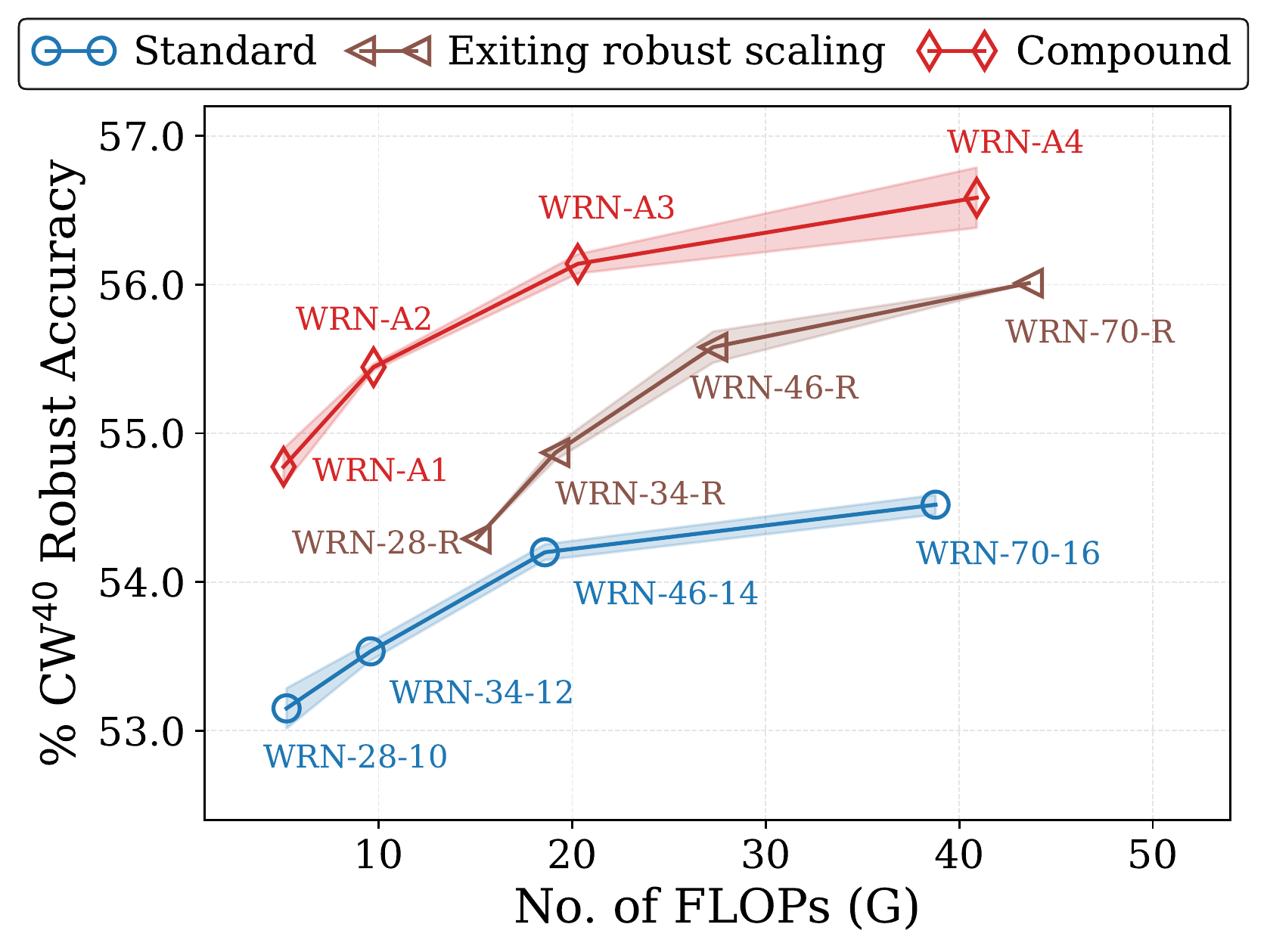}
    \caption{\scriptsize CW$^{40}$ vs. $^{\#}$FLOPs \label{fig:abl_c10_cw40_flops_scale_compare}}
    \end{subfigure}
    \end{minipage}\hfill
    \begin{minipage}{0.48\textwidth}
    \vspace{-8pt}
    \caption{(a, b) Comparison among standard scaling (\textcolor{tab_blue}{\textbf{blue}} curve), existing robust scaling \cite{huang2021exploring} (\textcolor{tab_brown}{\textbf{brown}} curve), the identified independent depth/width scaling (\textcolor{tab_orange}{\textbf{orange}}/\textcolor{tab_green}{\textbf{green}} curve) from \S\ref{sec:independent_scale}, and the identified compound scaling \ourscale{} (\textcolor{tab_red}{\textbf{red}} curve) on CIFAR-10. {\footnotesize[$D_1$, $D_2$, $D_3$]} and {\footnotesize[$W_1$, $W_2$, $W_3$]} denote stage-wise depth and width (in terms of widening factors) settings, respectively. For independent depth scaling, we use the width settings from the standard scaling and vice-versa for independent width scaling. All scaling strategies are applied to WRNs (i.e., basic residual block).\label{fig:scaling}}
    \end{minipage}
    \vspace{-15pt}
\end{figure}

Empirically, we compare our compound scaling to independent scaling by depth/width, the standard scaling (i.e., WRN-28-10, WRN-70-16, etc.), and the existing robust scaling from Huang \etal~\cite{huang2021exploring} under baseline adversarial training in Figure~\ref{fig:scaling}. Note that Huang \etal~\cite{huang2021exploring} only report one network, WRN-34-R. But we apply their (width) scaling rule to other WRN networks at different depths and obtain a set of WRN-R networks.
We observe that, in general, \ourscale{} achieves the best trade-off between robustness and network complexity, yielding networks that offer substantial improvements in robust accuracy over existing scaling methods while being an order of magnitude more efficient. 
In particular, WRN-A1 (i.e., the least complex network from \ourscale{}) is $3.8\times$ more compact ({\footnotesize$^{\#}$}Params) and efficient ({\footnotesize$^{\#}$}FLOPs) than WRN-34-R~\cite{huang2021exploring} while being similar in adversarial robustness. \iffalse{\color{red} The final network obtained by Huang \etal{} through their compound scaling rule used $2\times$ more FLOPs than the standard WRN counterpart. \vishnu{I do not see it in the figure.}}\fi
%
In addition, WRN-A1 is more adversarially robust than WRN-70-16 (i.e., the most complex network from the standard scaling) while being $14\times$ more compact and $8\times$ more efficient. 
Our findings suggest that effective compound policies do exist for scaling networks under adversarial training, and our \ourscale{} is one such realization.

% -------------------------------------------------------------------------------------
% -------------------------------------------------------------------------------------

% -------------------------------------------------------------------------------------
\begin{table*}[t]
\centering
\caption{Comparison of white-box adversarial robustness under baseline AT with TRADES \cite{zhang2019theoretically}. The best results are in bold, and relative improvements over 2$^{\rm{nd}}$ best result in each section are in \textcolor{red}{red}. Results are averaged over three runs with different seeds. \label{tab:robust_acc_vs_attack} \vspace{-5pt}} 
%\label{fig:main_results_nas} 
\resizebox{.98\textwidth}{!}{%
\begin{tabular}{@{\hspace{2mm}}l|cc|cccc|cccc@{\hspace{2mm}}}
    \toprule
    \multirow{2}{*}{Model} & \multirow{2}{*}{\begin{tabular}[c]{@{}c@{}}$^{\#}$P\\(M)\end{tabular}} & \multirow{2}{*}{\begin{tabular}[c]{@{}c@{}}$^{\#}$F\\(G)\end{tabular}} & \multicolumn{4}{c|}{CIFAR-10} & \multicolumn{4}{c}{CIFAR-100} \\ \cmidrule(lr){4-7} \cmidrule(lr){8-11} 
     &  &  & Clean & PGD$^{20}$ & CW$^{40}$ & AutoAttack & Clean & PGD$^{20}$ & CW$^{40}$ & AutoAttack \\ \midrule
    WRN-28-10 & 36.5 & 5.20 & $84.62_{\pm{0.06}}$ & $55.90_{\pm{0.21}}$ & $53.15_{\pm{0.33}}$ & $51.66_{\pm{0.29}}$ & $56.30_{\pm{0.28}}$ & $29.91_{\pm{0.40}}$ & $26.22_{\pm{0.23}}$ & $25.26_{\pm{0.06}}$  \\
    RobNet-large-v2 & 33.3 & 5.10 & $84.57_{\pm{0.16}}$ & $52.79_{\pm{0.08}}$ & $48.94_{\pm{0.13}}$ & $47.48_{\pm{0.04}}$ & $55.27_{\pm{0.02}}$ & $29.23_{\pm{0.15}}$ & $24.63_{\pm{0.11}}$ & $23.69_{\pm{0.19}}$ \\
    AdvRush ($7@96$) & 32.6 & 4.97 & $84.95_{\pm{0.12}}$ & $56.99_{\pm{0.08}}$ & $53.27_{\pm{0.03}}$ & $52.90_{\pm{0.11}}$ & $56.40_{\pm{0.09}}$ & $30.40_{\pm{0.21}}$ & $26.16_{\pm{0.03}}$ & $25.27_{\pm{0.02}}$ \\
    RACL ($7@104$) & 32.5 & \textbf{4.93} & $83.91_{\pm{0.32}}$ & $55.98_{\pm{0.15}}$ & $53.22_{\pm{0.08}}$ & $51.37_{\pm{0.11}}$ & $56.09_{\pm{0.08}}$ & $30.38_{\pm{0.03}}$ & $26.65_{\pm{0.02}}$ & $25.65_{\pm{0.10}}$ \\
    
    \ournetwork{}-A1 (ours) & \textbf{19.2} & 5.11 & \bm{$85.46$} \footnotesize(\textcolor{red}{$\uparrow\bm{0.5}$}) & \bm{$58.74$} \footnotesize(\textcolor{red}{$\uparrow\bm{1.8}$}) & \bm{$55.72$} \footnotesize(\textbf{\textcolor{red}{$\uparrow\bm{2.6}$}}) & \bm{$54.42$} \footnotesize(\textcolor{red}{$\uparrow\bm{1.5}$}) & \bm{$59.34$} \footnotesize(\textcolor{red}{$\uparrow\bm{2.9}$}) & \bm{$32.70$} \footnotesize(\textcolor{red}{$\uparrow\bm{2.3}$}) & \bm{$27.76$} \footnotesize(\textcolor{red}{$\uparrow\bm{1.1}$}) & \bm{$26.75$} \footnotesize(\textcolor{red}{$\uparrow\bm{1.1}$}) \\ \midrule
    WRN-34-12 & 66.5 & \textbf{9.60} & $84.93_{\pm{0.24}}$ & $56.01_{\pm{0.28}}$ & $53.53_{\pm{0.15}}$ & $51.97_{\pm{0.09}}$  & $56.08_{\pm{0.41}}$ & $29.87_{\pm{0.23}}$ & $26.51_{\pm{0.11}}$ & $25.47_{\pm{0.10}}$ \\
    WRN-34-R & 68.1 & 19.1 & $85.80_{\pm{0.08}}$ & $57.35_{\pm{0.09}}$ & $54.77_{\pm{0.10}}$ & $53.23_{\pm{0.07}}$ & $58.78_{\pm{0.11}}$ & $31.17_{\pm{0.08}}$ & $27.33_{\pm{0.11}}$ & $26.31_{\pm{0.03}}$ \\
    % \rowcolor{Gray!50}
    \ournetwork{}-A2 (ours) & \textbf{39.0} & 10.8 & \bm{$85.80$} \footnotesize(\textcolor{red}{$\uparrow\bm{0.0}$}) & \bm{$59.72$} \footnotesize(\textcolor{red}{$\uparrow\bm{2.4}$}) & \bm{$56.74$} \footnotesize(\textcolor{red}{$\uparrow\bm{2.0}$}) & \bm{$55.49$} \footnotesize(\textcolor{red}{$\uparrow\bm{2.3}$}) & \bm{$59.38$} \footnotesize(\textcolor{red}{$\uparrow\bm{0.6}$}) & \bm{$33.0$} \footnotesize(\textcolor{red}{$\uparrow\bm{1.8}$}) & \bm{$28.71$} \footnotesize(\textcolor{red}{$\uparrow\bm{1.4}$}) & \bm{$27.68$} \footnotesize(\textcolor{red}{$\uparrow\bm{1.4}$}) \\ \midrule
    WRN-46-14 & 128 & \textbf{18.6} & $85.22_{\pm{0.15}}$ & $56.37_{\pm{0.18}}$ & $54.19_{\pm{0.11}}$ & $52.63_{\pm{0.18}}$ & $56.78_{\pm{0.47}}$ & $30.03_{\pm{0.07}}$ & $27.27_{\pm{0.05}}$ & $26.28_{\pm{0.03}}$ \\
    % \rowcolor{Gray!50} # 
    \ournetwork{}-A3 (ours) & \textbf{75.9} & 19.9 & \bm{$86.79$} \footnotesize(\textcolor{red}{$\uparrow\bm{1.6}$}) & \bm{$60.10$} \footnotesize(\textcolor{red}{$\uparrow\bm{3.7}$}) & \bm{$57.29$} \footnotesize(\textcolor{red}{$\uparrow\bm{3.1}$}) & \bm{$55.84$} \footnotesize(\textcolor{red}{$\uparrow\bm{3.2}$}) & \bm{$60.16$} \footnotesize(\textcolor{red}{$\uparrow\bm{3.4}$}) & \bm{$33.59$} \footnotesize(\textcolor{red}{$\uparrow\bm{3.6}$}) & \bm{$29.58$} \footnotesize(\textcolor{red}{$\uparrow\bm{2.3}$}) & \bm{$28.48$} \footnotesize(\textcolor{red}{$\uparrow\bm{2.2}$}) \\ \midrule
    WRN-70-16 & 267 & \textbf{38.8} & $	85.51_{\pm{0.24}}$ & $56.78_{\pm{0.16}}$ & $54.52_{\pm{0.16}}$ & $52.80_{\pm{0.14}}$ & $56.93_{\pm{0.61}}$ & $29.76_{\pm{0.17}}$ & $27.20_{\pm{0.16}}$ & $26.12_{\pm{0.24}}$ \\
    % \rowcolor{Gray!50}
    \ournetwork{}-A4 (ours) & \textbf{147} & 39.4 & \bm{$87.10$} \footnotesize(\textcolor{red}{$\uparrow\bm{1.6}$}) & \bm{$60.26$} \footnotesize(\textcolor{red}{$\uparrow\bm{3.5}$}) & \bm{$57.9$} \footnotesize(\textcolor{red}{$\uparrow\bm{3.4}$}) & \bm{$56.29$} \footnotesize(\textcolor{red}{$\uparrow\bm{3.5}$}) & \bm{$61.66$} \footnotesize(\textcolor{red}{$\uparrow\bm{4.7}$}) & \bm{$34.25$} \footnotesize(\textcolor{red}{$\uparrow\bm{4.5}$}) & \bm{$30.04$} \footnotesize(\textcolor{red}{$\uparrow\bm{2.8}$}) & \bm{$29.00$} \footnotesize(\textcolor{red}{$\uparrow\bm{2.9}$}) \\
    \bottomrule
    \end{tabular}%
    }
    \vspace{-10pt}
\end{table*}

\section{Adversarially Robust Residual Networks\label{sec:robustresnet}}
We use \ourscale{} to scale our \ourblock{} to present a portfolio of adversarially robust residual networks, dubbed \emph{\ournetwork{}s}, spanning a broad spectrum of model FLOP budgets (i.e., $5$G - $40$G FLOPs). For reference, we name them as \ournetwork{}-A1 to -A4, where the FLOPs budget is doubled for every subsequent network. See Table~\ref{tab:network_spec} for detailed specifications. We then compare \ournetwork{}s to a set of representative robust architectures proposed in the literature. These include, RobNet \cite{NAS_Meets_Robustness}, RACL \cite{dong2020adversarially}, AdvRush \cite{mok2021advrush}, and WRN-34-R \cite{huang2021exploring}. Specifically, we align the network complexity of AdvRush and RACL models by adjusting the number of repetitions of the normal cell $N$ and the input {\footnotesize$^{\#}$}channels of the first normal cell $C$, denoted as ($N@C$).

\vspace{2pt}
\noindent\textbf{Comparison under Baseline Adversarial Training:} Table~\ref{tab:robust_acc_vs_attack} presents the results under baseline adversarial training with TRADES \cite{zhang2019theoretically}. In general, \ournetwork{}s consistently outperform existing robust networks across multiple datasets, attacks, and model-capacity regions. In particular, \ournetwork{}-A1 achieves \emph{$1.5\%$ higher AutoAttack robust accuracy} with \emph{$1.7{\times}$ fewer parameters} than AdvRush \cite{mok2021advrush}, a robust block designed by differentiable neural architecture search; \ournetwork{}-A2 achieves \emph{$2.3\%$ higher AutoAttack robust accuracy} with \emph{$1.8{\times}$ fewer parameters and FLOPs} than WRN-34-R \cite{huang2021exploring}. Additional comparisons under baseline adversarial training methods with different loss functions (i.e., SAT \cite{madry2018towards}, and MART \cite{Wang2020Improving}) are presented in Table~\ref{tab:robust_acc_vs_attack_other}. We observe that the improvements afforded by \ournetwork{}s generalize well to other loss formulations under baseline adversarial training routines.

% -------------------------------------------------------------------------------------
\begin{table}[t]
\centering
\caption{Additional comparison of white-box adversarial robustness under baseline adversarial training with SAT \cite{madry2018towards}, and MART \cite{Wang2020Improving}. The best results are in bold, and relative improvements are in \textcolor{red}{red}. Results are averaged over three runs with different seeds. \label{tab:robust_acc_vs_attack_other}\vspace{-5pt}} 
%\label{fig:main_results_nas} 
\resizebox{.485\textwidth}{!}{%
\begin{tabular}{@{\hspace{2mm}}l|cc|cc|cc@{\hspace{2mm}}}
    \toprule
    \multirow{2}{*}{Model} & \multirow{2}{*}{\begin{tabular}[c]{@{}c@{}}$^{\#}$P\\(M)\end{tabular}} & \multirow{2}{*}{\begin{tabular}[c]{@{}c@{}}$^{\#}$F\\(G)\end{tabular}} & \multicolumn{2}{c|}{SAT \cite{madry2018towards}} & \multicolumn{2}{c}{MART \cite{Wang2020Improving}} \\ \cmidrule(lr){4-5} \cmidrule(lr){6-7} 
     &  &  & PGD$^{20}$ & CW$^{40}$ & PGD$^{20}$ & CW$^{40}$  \\ \midrule
    WRN-28-10 & 36.5 & 5.20 & $52.44_{\pm{0.36}}$ & $50.97_{\pm{0.09}}$ & $57.69_{\pm{0.11}}$ & $52.88_{\pm{0.28}}$ \\
    \ournetwork{}-A1 & \textbf{19.2} & \textbf{5.11} & \bm{$57.62$} \footnotesize(\textcolor{red}{$\uparrow\bm{5.2}$}) & \bm{$56.06$} \footnotesize(\textcolor{red}{$\uparrow\bm{5.1}$}) & \bm{$59.34$} \footnotesize(\textbf{\textcolor{red}{$\uparrow\bm{1.7}$}}) & \bm{$54.42$} \footnotesize(\textcolor{red}{$\uparrow\bm{1.5}$}) \\ \midrule
    WRN-34-12 & 66.5 & \textbf{9.60} & $52.85_{\pm{0.40}}$ & $51.36_{\pm{0.33}}$ & $57.40_{\pm{0.13}}$ & $53.11_{\pm{0.00}}$  \\
    % 
    % WRN-34-R & 68.1 & 19.1 & $85.80$ & $57.35$ & $54.77$ & $53.23$ \\
    % \rowcolor{Gray!50}
    \ournetwork{}-A2 & \textbf{39.0} & 10.8 & \bm{$58.39$} \footnotesize(\textcolor{red}{$\uparrow\bm{5.5}$}) & \bm{$56.99$} \footnotesize(\textcolor{red}{$\uparrow\bm{5.6}$}) & \bm{$60.33$} \footnotesize(\textcolor{red}{$\uparrow\bm{2.9}$}) & \bm{$55.51$} \footnotesize(\textcolor{red}{$\uparrow\bm{2.4}$}) \\ \midrule
    WRN-46-14 & 128 & \textbf{18.6} & $53.67_{\pm{0.03}}$ & $52.95_{\pm{0.04}}$ & $58.43_{\pm{0.15}}$ & $54.32_{\pm{0.17}}$\\
    % \rowcolor{Gray!50} # 
    \ournetwork{}-A3  & \textbf{75.9} & 19.9 & \bm{$58.81$} \footnotesize(\textcolor{red}{$\uparrow\bm{5.1}$}) & \bm{$57.60$} \footnotesize(\textcolor{red}{$\uparrow\bm{4.7}$}) & \bm{$60.95$} \footnotesize(\textcolor{red}{$\uparrow\bm{2.5}$}) & \bm{$56.52$} \footnotesize(\textcolor{red}{$\uparrow\bm{2.2}$}) \\ \midrule
    WRN-70-16 & 267 & \textbf{38.8} & $54.12_{\pm{0.08}}$ & $50.52_{\pm{0.18}}$ & $58.15_{\pm{0.28}}$ & $54.37_{\pm{0.07}}$\\
    % \rowcolor{Gray!50}
    \ournetwork{}-A4 & \textbf{147} & 39.4 & \bm{$59.01$} \footnotesize(\textcolor{red}{$\uparrow\bm{4.9}$}) & \bm{$57.85$} \footnotesize(\textcolor{red}{$\uparrow\bm{7.3}$}) & \bm{$61.88$} \footnotesize(\textcolor{red}{$\uparrow\bm{3.7}$}) & \bm{$57.55$} \footnotesize(\textcolor{red}{$\uparrow\bm{3.2}$}) \\
    \bottomrule
    \end{tabular}%
    }
    \vspace{-10pt}
\end{table}

\vspace{2pt}
\noindent\textbf{Comparison under Advanced Adversarial Training:} 
Table~\ref{tab:robust_acc_vs_attack_500k} presents results under advanced AT with an additional 500K unlabeled external images extracted from the 80M Tiny Images dataset \cite{carmon2019unlabeled}. \ournetwork{}-A1 achieves $63.70\%$ AutoAttack robust accuracy with $19.2$M parameters, ranking 7$^{\rm{th}}$ (as of 20th December, 2022)  on RobustBench CIFAR-10 leaderboard \cite{croce2020robustbench}. Note that higher-ranked methods (i.e., top-6 on RobustBench CIFAR-10 leaderboard) use networks WRN-70-16 or WRN-106-16, which have at least \emph{$250$M more parameters} than \ournetwork{}-A1. Furthermore, \ournetwork{}-A1 is \emph{$\sim1.2\%$ more robust} against AutoAttack with \emph{$3.5\times$ fewer parameters} and \emph{$3.7\times$ fewer FLOPs} than WRN-34-R \cite{huang2021exploring}.

% -------------------------------------------------------------------------------------
\begin{table}[ht]
\centering
\caption{Comparison of white-box adversarial robustness under advanced adversarial training with extra 500k external data \cite{carmon2019unlabeled}. \label{tab:robust_acc_vs_attack_500k}\vspace{-5pt}} 
\resizebox{.485\textwidth}{!}{%
\begin{tabular}{@{\hspace{2mm}}lccccc@{\hspace{2mm}}}
    \toprule
    Method & Architecture & $^{\#}$P (M) & $^{\#}$F (G) & AutoAttack \\
    \midrule
    RST~\cite{carmon2019unlabeled} & WRN-28-10 & $36.5$ & $5.20$ & $59.53$ \\ 
    AWP~\cite{wu2020adversarial} & WRN-28-10 & $36.5$ & $5.20$ & $60.04$ \\
    HAT~\cite{rade2021reducing} & WRN-28-10 & $36.5$ & $5.20$ & $62.50$ \\ 
    Gowal \etal{}~\cite{gowal2020uncovering} & WRN-28-10 & $36.5$ & $5.20$ & $62.80$ \\ \midrule
    Huang \etal{}~\cite{huang2021exploring} & WRN-34-R & $68.1$ & $19.1$ & $62.54$ \\ 
    Ours & \ournetwork{}-A1 & $\bm{19.2}$ \footnotesize(\textcolor{red}{$\downarrow\bm{3.5\times}$}) & $\bm{5.11}$ \footnotesize(\textcolor{red}{$\downarrow\bm{3.7\times}$}) & $\bm{63.70}$ \footnotesize(\textcolor{red}{$\uparrow\bm{1.2}$}) \\
    \bottomrule
    \end{tabular}%
    }
\end{table}
% -------------------------------------------------------------------------------------

Table~\ref{tab:robust_acc_vs_attack_wo_500k} presents results under advanced adversarial training without additional data, either external (e.g., the 500K data \cite{carmon2019unlabeled}) or generated by generative models (e.g., DDPM \cite{gowal2021improving}). In particular, \ournetwork{}-A1 achieves \emph{$1.9\%$ higher AutoAttack robust accuracy} with \emph{$1.9{\times}$ fewer parameters} than state-of-the-art\footnote{We consider state-of-the-art without external or generated data. Note that the ``Extra data'' column of the RobustBench CIFAR-10 leaderboard only accounts for external data; please also see the description under the ``Method'' column for approaches that do leverage \emph{generated} data.} method built upon WRN-28-10~\cite{rebuffi2021data}. 
Furthermore, \ournetwork{}-A2 is \emph{$6.8\times$ more compact (parameters}) and \emph{$3.7\times$ more efficient (FLOPs)} while matching the state-of-the-art AutoAttack robust accuracy.
And \ournetwork{}-A4 achieves $61.10\%$ AutoAttack robust accuracy with $39.4$M parameters on CIFAR-10 against AutoAttack with $\ell_{\infty}$ perturbations of size $\epsilon=8/255$---\emph{an improvement of $1.0\%$ robust accuracy with $120$ million fewer parameters} compared to the state-of-the-art without external or generated data \cite{rebuffi2021data}. 

% -------------------------------------------------------------------------------------
\begin{table}[ht]
\centering
\caption{Comparison of white-box adversarial robustness under advanced adversarial training. Our method builds upon Rebuffi \etal{}~\cite{rebuffi2021data}, which applies CutMix \cite{yun2019cutmix} data augmentation. \label{tab:robust_acc_vs_attack_wo_500k}\vspace{-5pt}} 
%\label{fig:main_results_nas} 
\resizebox{.485\textwidth}{!}{%
\begin{tabular}{@{\hspace{2mm}}lccccc@{\hspace{2mm}}}
    \toprule
    Method & Architecture & $^{\#}$P (M) & $^{\#}$F (G) & AutoAttack \\
    \midrule
    TRADES~\cite{zhang2019theoretically} & WRN-34-10 & $46.2$ & $6.66$ & $53.08$ \\ 
    Rebuffi \etal{}~\cite{rebuffi2021data} & WRN-28-10 & $36.5$ & $5.20$ & $57.50$ \\ 
    Ours & \ournetwork{}-A1 & $\bm{19.2}$ \footnotesize(\textcolor{red}{$\downarrow\bm{1.9\times}$}) & $\bm{5.11}$ \footnotesize(\textcolor{black}{$\sim$}) & $\bm{59.39}$ \footnotesize(\textcolor{red}{$\uparrow\bm{1.9}$}) \\
    \midrule
    % Rice \etal{}~\cite{rice2020overfitting} & WRN-34-20 & 185 & \textbf{26.6} & 53.42 \\ 
    % Gowal \textit{et al.}~\cite{gowal2020uncovering} & WRN-34-20 & 185 & 56.86 \\ 
    Gowal \etal{}~\cite{gowal2020uncovering} & WRN-70-16 & $267$ & $\bm{38.8}$ & $57.20$ \\ 
    Rebuffi \etal{}~\cite{rebuffi2021data} & WRN-70-16 & $267$ & $\bm{38.8}$ & $60.07$ \\
    Ours & \ournetwork{}-A2 & $\bm{39.0}$ \footnotesize(\textcolor{red}{$\downarrow\bm{6.8\times}$}) & $\bm{10.6}$ \footnotesize(\textcolor{red}{$\downarrow\bm{3.7\times}$}) & ${60.00}$ \footnotesize(\textcolor{black}{$\sim$}) \\
    Ours & \ournetwork{}-A4 & ${147}$ \footnotesize(\textcolor{red}{$\downarrow{1.8\times}$}) & $39.4$ \footnotesize(\textcolor{black}{$\sim$}) & $\bm{61.10}$ \footnotesize(\textcolor{red}{$\uparrow\bm{1.0}$}) \\
    \bottomrule
    \end{tabular}%
    }
    \vspace{-10pt}
\end{table}
% -------------------------------------------------------------------------------------
\section{Discussion}
This paper identified specific architectural design elements that impact adversarial robustness. The reliability of our observations has been ensured by systematically verifying them on multiple datasets, across multiple adversarial attacks, and over multiple repetitions. We affirm that the proposed \ourblock{}, \ourscale{}, and \ournetwork{} have immediate practical relevance in designing adversarially robust networks. Nonetheless, our observations and contributions have been made through empirical experiments instead of theoretical analysis. However, as is often the case in deep learning (e.g., batch normalization~\cite{ioffe2015batch}, lottery ticket hypothesis~\cite{frankle2018the}, etc.), theoretical analysis usually follows empirical observations. Furthermore, most of the theoretical studies in adversarial robustness have focused on loss formulation. We hope this paper inspires theoretical exploration of the adversarial robustness properties of different architectural design elements as well.

\section{Concluding Remarks}
Novel architectural designs played a critical role in the overwhelming success of CNNs in various image analysis tasks. Despite this knowledge, studies on adversarial robustness have primarily been limited to a handful of basic residual networks, thus overlooking the impact of architecture on adversarial robustness. However, as we demonstrate in this paper, architectural design significantly affects adversarial robustness. As an illustration, we considered residual networks. We observed through systematically designed experiments that many advancements of residual blocks for standard ERM training translate well to improve adversarial robustness under adversarial training, albeit with minor modifications in some cases. 

Based on our observations, we designed RobustResNets as an alternative baseline for standard Wide Residual Networks, the de facto architecture of choice for designing adversarially robust networks. RobustResNets afford significant improvements in adversarial robustness while being more compact than state-of-the-art solutions, both without extra data and with 500K extra data. We hope that RobustResNets can serve as a new benchmark architecture for studying adversarial robustness and that our work inspires future exploration into the adversarial robustness of the wide range of architectures that have already proven effective under standard ERM training.

\section*{Acknowledgements}
Shihua Huang and Vishnu Naresh Boddeti are supported in part by the following financial assistance award 60NANB18D210 from U.S. Department of Commerce, National Institute of Standards and Technology. Zhichao Lu is supported by the National Natural Science Foundation of China (62106097) and the China Postdoctoral Science Foundation (2021M691424).

%%%%%%%%% REFERENCES
{\small
\bibliographystyle{ieee_fullname}
\bibliography{egbib}
}

\appendix
\section{Appendix}
\subsection{Extended Description of Related Work \label{sec:app_related_work}}
\noindent\textbf{Relation to existing works based on NAS.} Several recent works sought to find more robust DNN architectures via neural architecture search (NAS) -- Guo~\etal{} applied a one-shot NAS algorithm to design the topology of a cell structure (i.e., operations and connections among them) while leaving the network skeleton (i.e., width and depth) to human designs \cite{NAS_Meets_Robustness}; Mok \etal{} incorporated the smoothness of a DNN model's input loss landscape as an additional regularizer for NAS \cite{mok2021advrush}, among others \cite{ning2020discovering,chen2020anti,liu2021multi}. 

These NAS-based prior arts are limited in the following three aspects: (1) they focus on only one aspect of architecture (i.e., block topology) while leaving other components (e.g., activation, network depth, and width, etc.) to human designs; (2) they treat the design of an adversarially robust architecture as a black-box search problem where minimal architectural insights can be derived; (3) NAS is computationally expensive and adversarial training makes this challenge especially acute. 

In contrast, this work presents (i) a holistic study of different aspects of architecture, including block topology, activation, normalization, and scaling factors (i.e., network depth and width); (ii) through controlled and fine-grained experiments, we deliver precise knowledge on the impacts of these choices; (iii) empirically, we demonstrate that the network assembled on top of our derived knowledge outperforms existing networks designed via NAS by at least \emph{2.5\% robust accuracy against AutoAttack} (see Table~\ref{tab:robust_acc_vs_attack}).

% -------------------------------------------------------------------------------------
\begin{figure*}[!ht]
    \begin{subfigure}[b]{0.195\textwidth}
    \captionsetup{justification=centering}
    \centering
    \includegraphics[height=.13\textheight]{figure/se_variants/se.pdf}
    \caption{\scriptsize Standard SE\label{fig:app_se} \vspace{-8pt}}
    \end{subfigure}\hfill
    \begin{subfigure}[b]{0.195\textwidth}
    \captionsetup{justification=centering}
    \centering
    \includegraphics[height=.13\textheight]{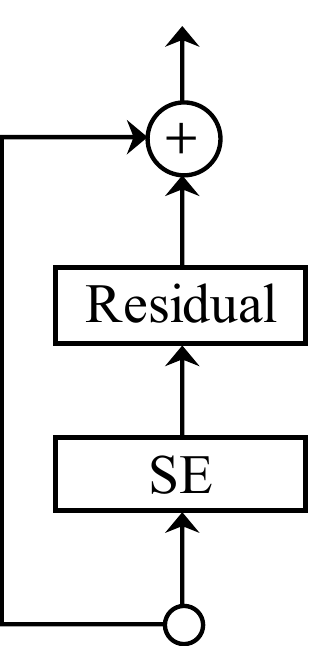}
    \caption{\scriptsize Pre-SE \label{fig:pre-se} \vspace{-8pt}}
    \end{subfigure}\hfill
    \begin{subfigure}[b]{0.195\textwidth}
    \captionsetup{justification=centering}
    \centering
    \includegraphics[height=.13\textheight]{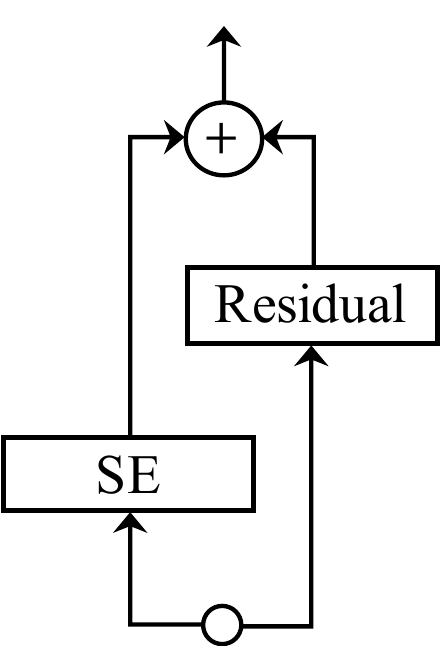}
    \caption{\scriptsize Identity-SE \label{fig:identity-se} \vspace{-8pt}}
    \end{subfigure}\hfill
    \begin{subfigure}[b]{0.195\textwidth}
    \captionsetup{justification=centering}
    \centering
    \includegraphics[height=.13\textheight]{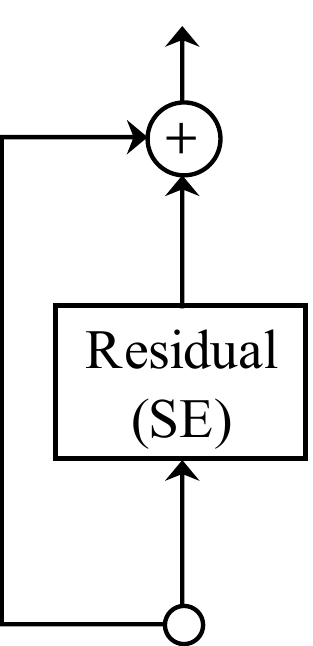}
    \caption{\scriptsize Conv3$\times$3-SE\label{fig:conv-se} \vspace{-8pt}}
    \end{subfigure}\hfill
    \begin{subfigure}[b]{0.195\textwidth}
    \captionsetup{justification=centering}
    \centering
    \includegraphics[height=.13\textheight]{figure/our_se.pdf}
    \caption{\scriptsize Residual SE (ours) \label{fig:our-se} \vspace{-8pt}}
    \end{subfigure} \\
    \centering
    \begin{subfigure}[b]{0.9\textwidth}
    \captionsetup{justification=centering}
    \centering
    \vspace{1.5em}
    \resizebox{\textwidth}{!}{%
    \begin{tabular}{@{\hspace{2mm}}l|l@{\hspace{2mm}}}
    \toprule
    Design & Explanation \\ \midrule
    (a) Standard SE & Place the SE module posterior to the main components of the residual block as proposed in \cite{senet}. \\
    (b) Pre-SE & Place the SE module a priori, i.e., before the main components of the residual block, also tried by \cite{senet}. \\
    (c) Identity-SE & Place the SE module in the skip-connection branch, also tried by \cite{senet}. \\
    (d) Conv3x3-SE & Place the SE module right after the $3\times3$ convolution, as done in MobileNetV3 \cite{howard2019searching}.  \\
    (e) Residual SE (ours) & \begin{tabular}[c]{@{}l@{}}Add an extra skip connection around the SE module to the standard SE integration design, \\ similarly to the FSM module from \cite{huang2021fapn}.\end{tabular} \\
    \bottomrule
    \end{tabular}%
    }
    \caption{\scriptsize \label{tab:se-table} \vspace{-8pt}}
    \end{subfigure}\\
    \begin{subfigure}[b]{0.65\textwidth}
    \centering
    \vspace{1.em}
    \captionsetup{justification=centering}
    \resizebox{.98\textwidth}{!}{%
        \begin{tabular}{@{\hspace{2mm}}lcccccc@{\hspace{2mm}}}
        \toprule
        % Designs (reduction ratio) & $^{\#}$P (M) & $^{\#}$F (G) & Clean & Robust \\ 
        \multirow{2}{*}{\begin{tabular}[c]{@{}l@{}}Design\end{tabular}} & \multirow{2}{*}{\begin{tabular}[c]{@{}c@{}}Reduction\\ratio\end{tabular}}  & \multirow{2}{*}{\begin{tabular}[c]{@{}c@{}}$^{\#}$P (M)\end{tabular}} & \multirow{2}{*}{\begin{tabular}[c]{@{}c@{}}$^{\#}$F (G)\end{tabular}} & \multirow{2}{*}{\begin{tabular}[c]{@{}c@{}}Clean Acc. (\%)\end{tabular}} & \multicolumn{2}{c}{Robust Acc. (\%)} \\
        &  & &  &  & PGD$^{20}$ & CW$^{40}$ \\
        \midrule
        w/o SE & -- & 265 & 39.0 & 85.47 & 57.49 & 55.07 \\
        Standard SE & \multirow{4}{*}{$r=16$} & 296 & 39.1 & 84.56 (\color{red}{-0.91}) & 56.87 (\color{red}{-0.62}) & 54.52 (\color{red}{-0.55}) \\
        Conv3$\times$3-SE &  & 273 & 39.1 & 85.26 (\color{red}{-0.21}) & 57.10 (\color{red}{-0.39}) & 54.77 (\color{red}{-0.40}) \\
        Identity-SE &  & 293 & 39.1 & 85.20 (\color{red}{-0.27}) & 57.04 (\color{red}{-0.45}) & 54.94 (\color{red}{-0.13}) \\
        Pre-SE &  & 293 & 39.1 & \textbf{85.81} (\color{Emerald}{+0.34}) & 57.31 (\color{red}{-0.18}) & 55.32 (\color{Emerald}{+0.25}) \\
        \midrule
        \multirow{3}{*}{Residual SE (ours)} & $r=16$ & 296 & 39.1 & {85.75} (\color{Emerald}{+0.28}) & {57.86} ({\color{Emerald}+0.37}) & {55.95} ({\color{Emerald}+0.88}) \\
         & $r=32$ & 281 & 39.1 & 85.22 (\color{red}{-0.25}) & \textbf{57.98} ({\color{Emerald}+0.49}) & {55.54} ({\color{Emerald}+0.47}) \\
         & $r=64$ & 273 & 39.1 & {85.61} (\color{Emerald}{+0.14}) & 57.77 ({\color{Emerald}+0.28}) & \textbf{56.05} ({\color{Emerald}+0.98}) \\ \bottomrule
        \end{tabular}%
    }
    \caption{\label{tab:se-results} \vspace{-8pt}}
    \end{subfigure}\hfill
    \begin{subfigure}[b]{0.30\textwidth}
    \centering
    \includegraphics[width=\textwidth]{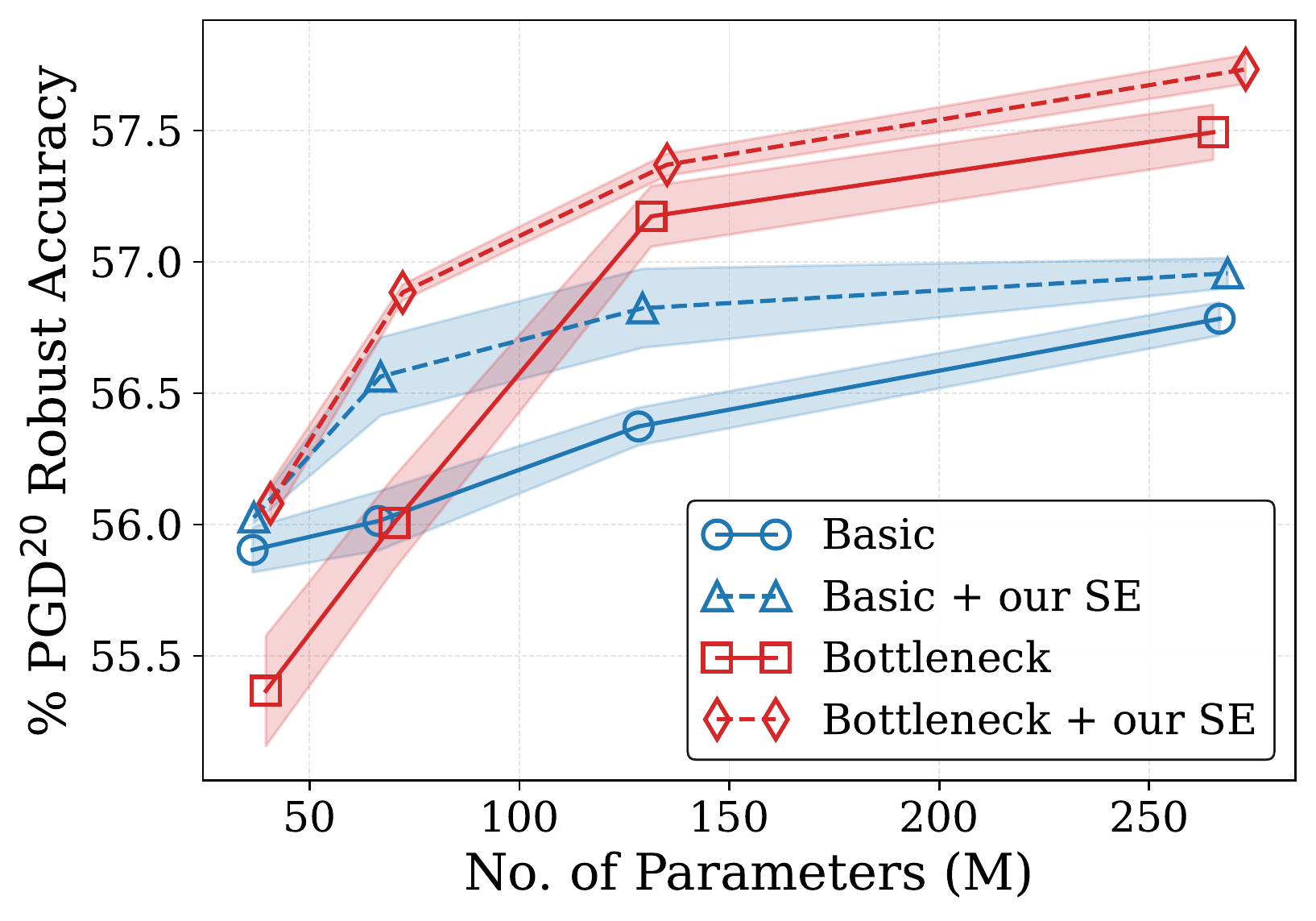}
    \caption{\scriptsize \label{fig:abl_se_c10_pgd20} \vspace{-8pt}}
    \end{subfigure}
    \caption{(a) - (e) An overview of SE integration designs studied in this work. (f) Description and (g) ablation results of the SE integration designs are shown in (a) - (e). (h) Comparing residual blocks with and without the proposed residual SE on CIFAR-10 against PGD$^{20}$ attack. \label{fig:overview_se_variants} \vspace{-1em}}
\end{figure*}
% -------------------------------------------------------------------------------------

\vspace{2pt}
\noindent\textbf{Relation to other existing works.}
There are recent works that aim to gain an understanding of adversarial robustness from an architectural perspective~\cite{xie2020smooth,gowal2020uncovering,cazenavette2021architectural,huang2021exploring,dai2022parameterizing,zhu2022robustness}. Among them, \cite{huang2021exploring} is most closely related to this paper. Accordingly, we provide an elaborated discussion on the relation to \cite{huang2021exploring} below and refer readers to the Related Work section in \S\ref{sec:related_work} for an overview of these methods. 

Huang~\etal{}~\cite{huang2021exploring} also investigated the impact of network width and depth via controlled experiments on the adversarial robustness of adversarially trained DNN models. Despite a similar motivation, our work is primarily different and enhanced in the following aspects:
\begin{enumerate}
    \item Huang~\etal{} only study network scaling factors (i.e., depth and width), while we study both block topology and network scaling. And as we demonstrated in this paper, both are critical architectural components for improving adversarial robustness. Specifically, we show that (i) improvement on block topology alone leads to $\sim3\%$ more robust accuracy; (ii) improvement on network scaling alone leads to $\sim2.5\%$ more robust accuracy; (iii) improvement on both block topology and network scaling leads to $3.5+\%$ more robust accuracy while being $\sim2\times$ more compact in terms of parameters. All results were evaluated against AutoAttack and relative to WRNs, the de-facto model for studying adversarial robustness. 

    \item Huang~\etal{} explored the interplay between network depth and width but observed that the independent scaling rules they identified for depth and width did not work well together and ultimately failed to design a compound rule to scale depth and width simultaneously\footnote{For more details, please refer to Section 4.3 in \cite{huang2021exploring}.}. In contrast, building upon our independent scaling rules, we identify an effective compound rule to simultaneously scale depth and width by properly distributing a given computational budget (e.g., FLOPs) over the number of layers and their width multipliers\footnote{See Section~\ref{sec:compund_scale} for more details.}. Empirically, we demonstrate that the compound scaling rule further improves independent scaling of depth and width by \emph{$\sim2\%$ and $\sim1\%$ more robust accuracy against $CW^{40}$ attack} for a small-capacity model, respectively (see Figure~\ref{fig:abl_compound_scale_visualization}). 

    \item The scaling rule identified by Huang \etal{} was evaluated at one model capacity only (i.e., $\sim68$M $^{\#}$Params), while, in this work, we demonstrate the efficacy of our scaling rules (i.e., both independent and compound scaling rules) across a broad spectrum of model-capacities, from $5$M to $270$M $^{\#}$Params. 

    \item Performance-wise, on top of using almost $2\times$ fewer $^{\#}$Params and $^{\#}$FLOPs, our model (i.e., \ournetwork{}-A2) consistently exhibits $1.4\%$ - $2.4\%$ higher robust accuracy over the model (i.e., WRN-34-R) scaled by Huang \etal{} across multiple datasets, attacks, and training settings. 
\end{enumerate}

\subsection{Extended Description of SE \label{sec:app_se}}
In this section, we first provide pictorial illustrations and descriptions of the five variations of SE that we tried in Figure~\ref{fig:se} - \ref{fig:our-se} and Table~\ref{tab:se-table}, respectively. Then, we provide additional results comparing our proposed residual SE among the five variations of SE in Table~\ref{tab:se-results}. Our residual SE is a simple yet effective variant of the standard SE that improves adversarial robustness while all other variants fail. Finally, we present the effect of incorporating our residual SE to both basic and bottleneck residual blocks in Figure~\ref{fig:abl_se_c10_pgd20}.

\subsection{Additional Results of Block Topology \label{sec:app_block_topology}}

\begin{figure}[ht]
    \begin{subfigure}[b]{0.48\textwidth}
    \centering
    \includegraphics[width=\textwidth]{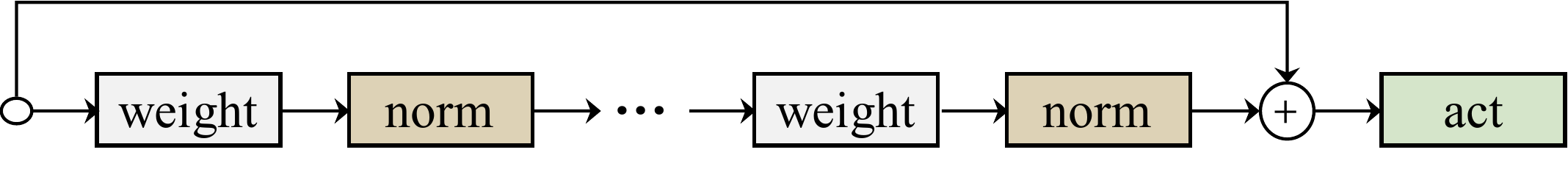}\\ \vspace{2em}
    \includegraphics[width=\textwidth]{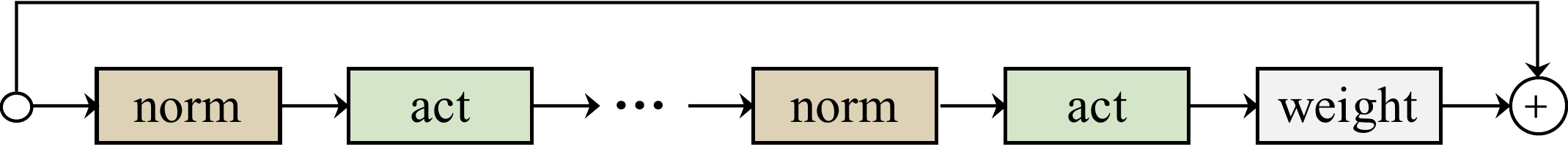}\\ \vspace{1em}
    \caption{Post-activation (top) and Pre-activation (bottom) \label{fig:post_pre_activation}}
    \end{subfigure} \\
    \begin{subfigure}[b]{0.235\textwidth}
    \centering
    \includegraphics[width=\textwidth]{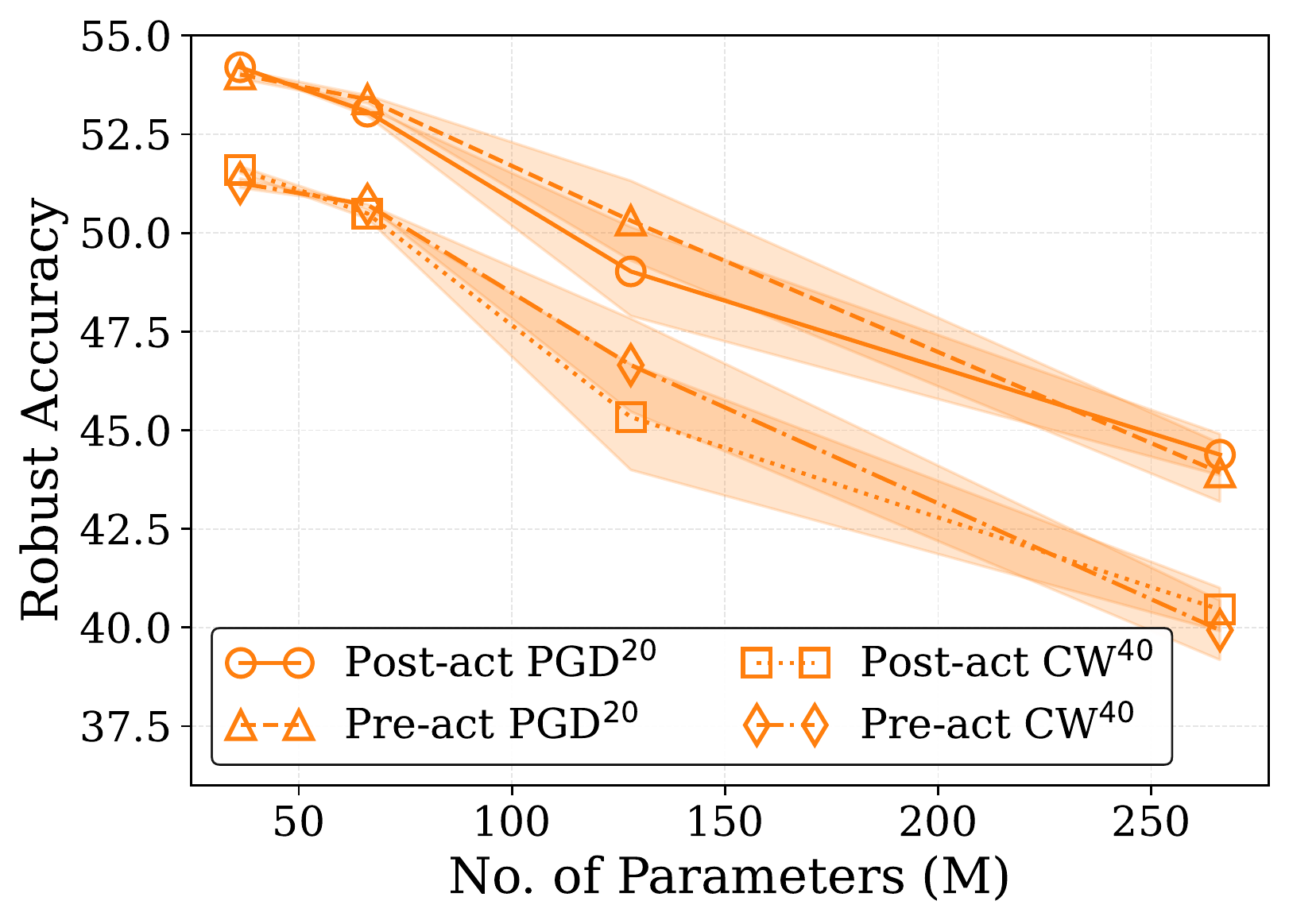}
    \caption{Non-residual block \label{fig:abl_topology_c10_vgg}}
    \end{subfigure} \hfill
    \begin{subfigure}[b]{0.235\textwidth}
    \centering
    \includegraphics[width=\textwidth]{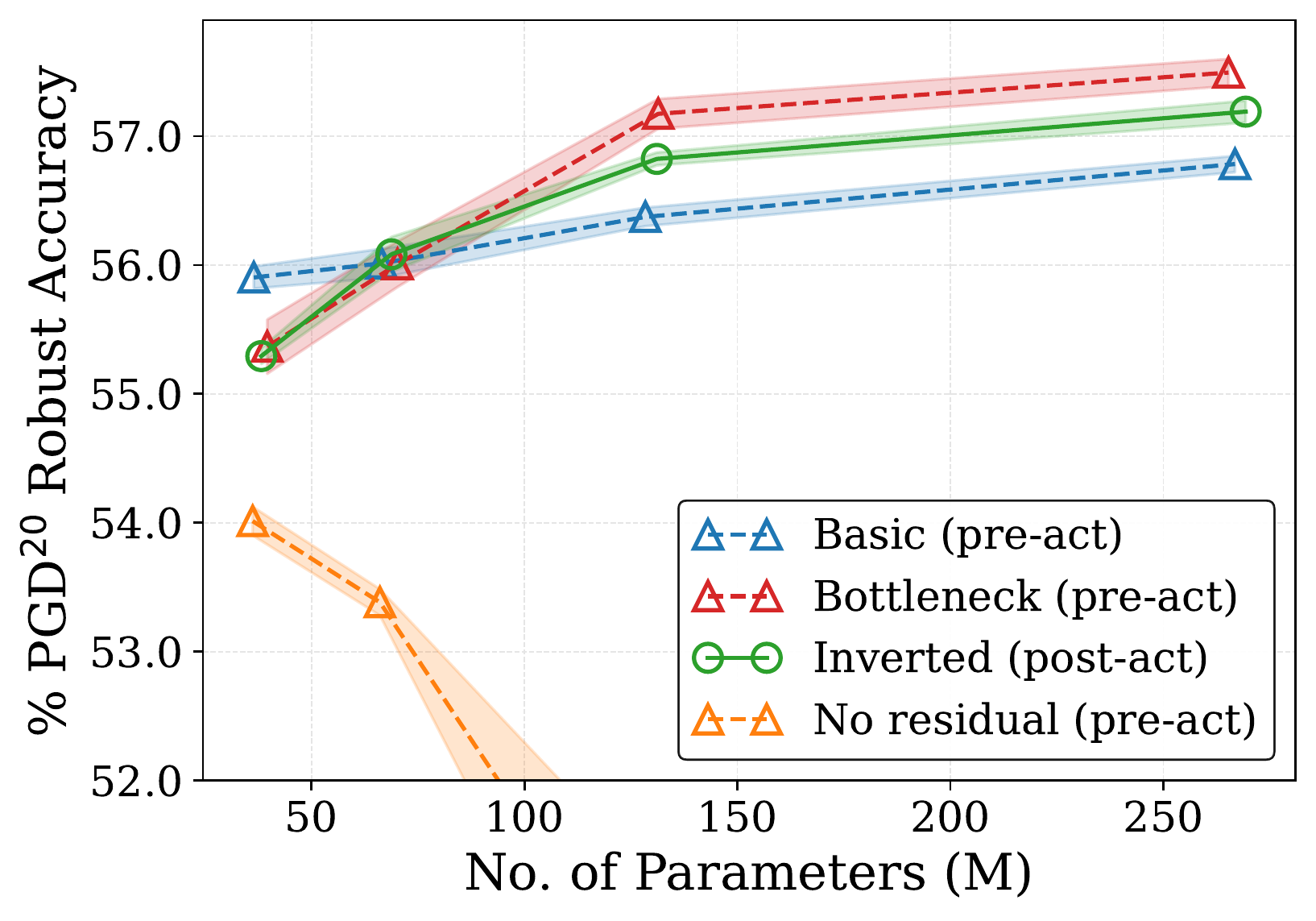}
    \caption{Comparison under PGD$^{20}$ \label{fig:abl_topology_c10_pgd20}}
    \end{subfigure}
    \caption{(a) A pictorial illustration of the standard post-activation (\emph{Top}) and pre-activation arrangements (\emph{Bottom}). (b) Comparing post- and pre-activation for a non-residual block (i.e., VGG block) on CIFAR-10. (c) Comparison among variants of a residual block with pre-activation on CIFAR-10 against PGD$^{20}$ attack. \label{fig:app_block_topology}}
\end{figure}

In this section, we first provide a visual comparison between post-activation and pre-activation in Figure~\ref{fig:post_pre_activation}, where the standard post-activation~\cite{he2016deep} places the activation function after the weights. In contrast, the pre-activation proposed by~\cite{he2016identity} places the activation function before the weights. Then, we compare the effectiveness of these two arrangements of activation for a non-residual block (i.e., VGG block) on CIFAR-10 in Figure \ref{fig:abl_topology_c10_vgg}, followed by comparison over variants of residual blocks with pre-activation on CIFAR-10 against PGD$^{20}$ attack in Figure \ref{fig:abl_topology_c10_pgd20}. 

% -------------------------------------------------------------------------------------
% -------------------------------------------------------------------------------------

\subsection{Additional Results of Aggregated and Hierarchical Convolutions \label{sec:app_block_connection}}

\begin{figure}[ht]
    \begin{subfigure}[b]{0.23\textwidth}
    \centering
    \includegraphics[width=0.95\textwidth]{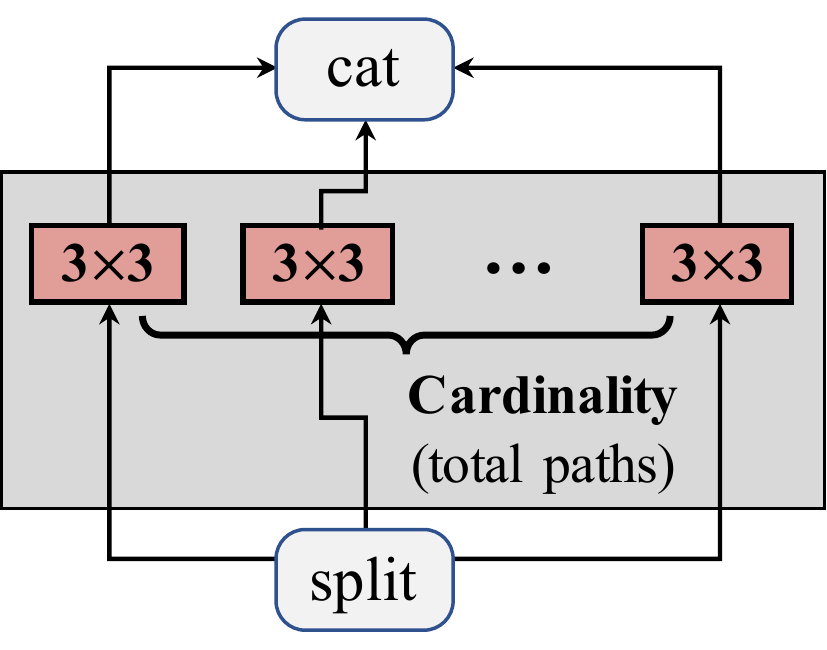}
    \caption{\scriptsize Aggregated convolution \label{fig:aggregated}}
    \end{subfigure}\hfill
    \begin{subfigure}[b]{0.235\textwidth}
    \centering
    \includegraphics[width=0.95\textwidth]{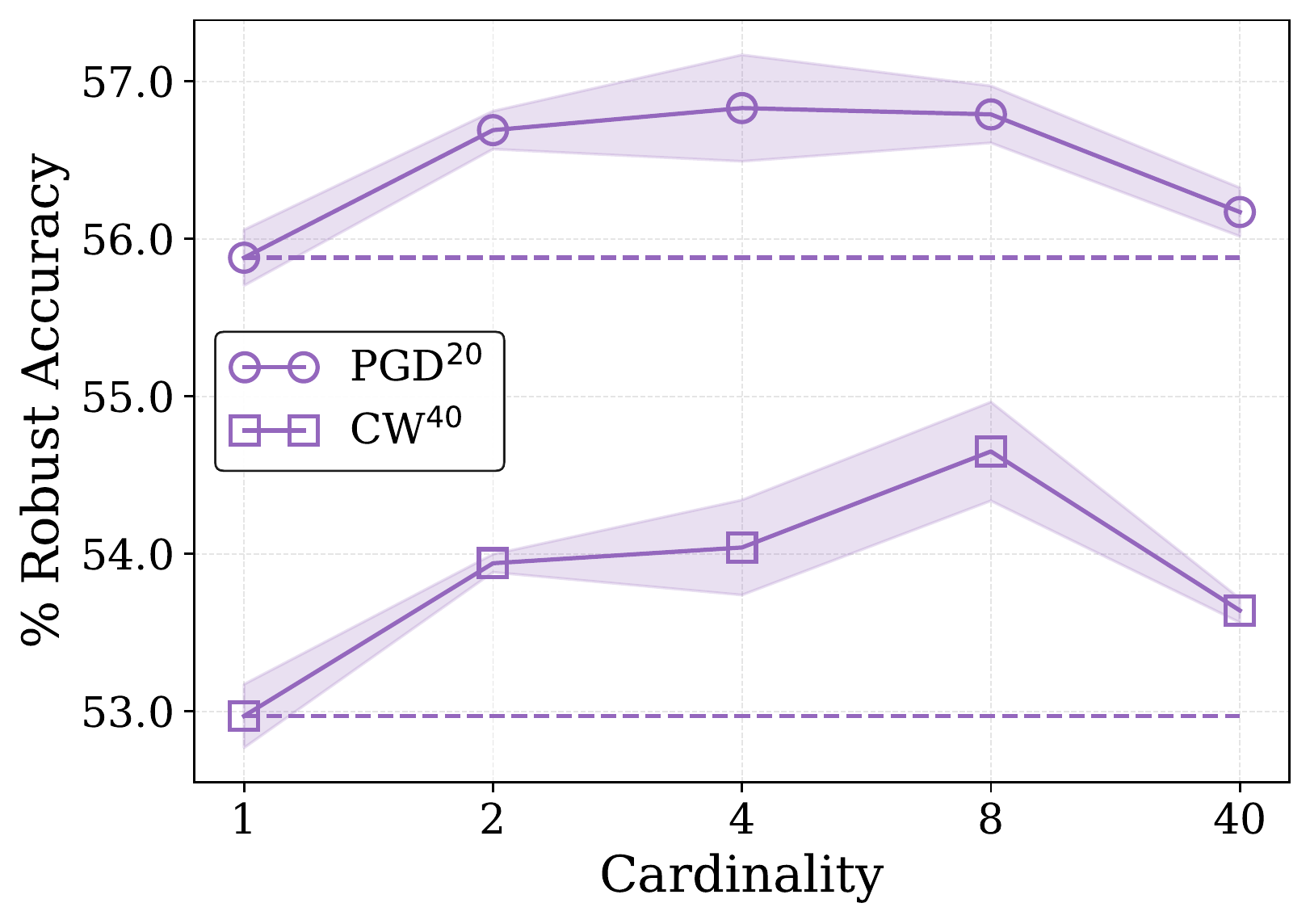}
    \caption{\scriptsize $D_i=5, W_i=12$ \label{fig:abl_aggre_d34_w12_c10}}
    \end{subfigure} \\
    \begin{subfigure}[b]{0.235\textwidth}
    \centering
    \includegraphics[width=0.95\textwidth]{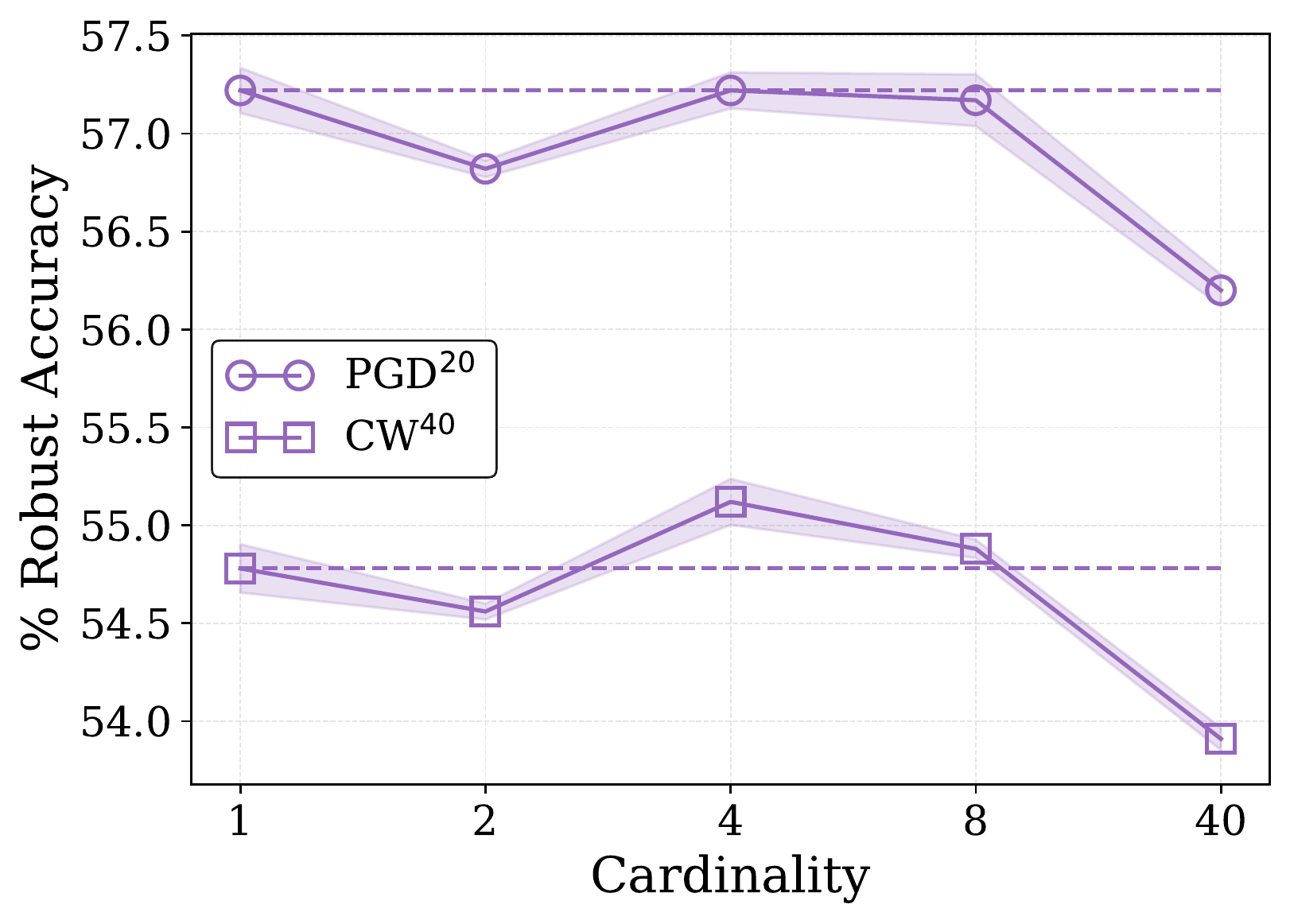}
    \caption{\scriptsize $D_i=7, W_i=14$  \label{fig:abl_aggre_d46_w14_c10}}
    \end{subfigure}\hfill
    \begin{subfigure}[b]{0.235\textwidth}
    \centering
    \includegraphics[width=0.95\textwidth]{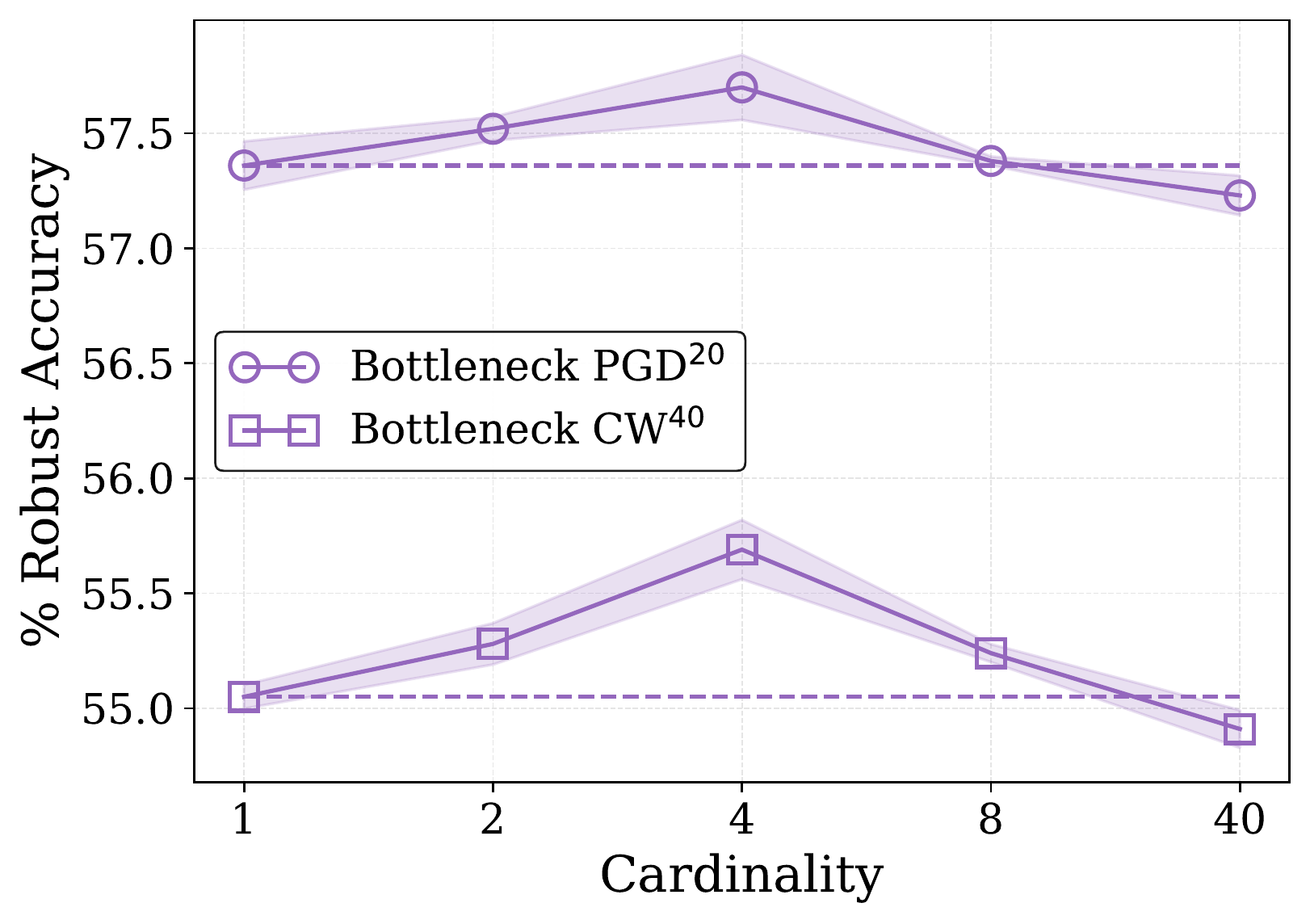}
    \caption{\scriptsize $D_i=11, W_i=16$  \label{fig:abl_aggre_d70_w16_c10}}
    \end{subfigure}
    \vspace{-8pt}
    \caption{
    (a) Aggregated convolution that splits a regular convolution into multiple parallel convolutions (cardinality). Results are then concatenated. (b, c, d) show the robustness of models from three different capacity regions. \label{fig:app_abl_aggre} \vspace{-1em}}
\end{figure}

\begin{figure}[ht]
    \begin{subfigure}[b]{0.23\textwidth}
    \centering
    \includegraphics[width=0.95\textwidth]{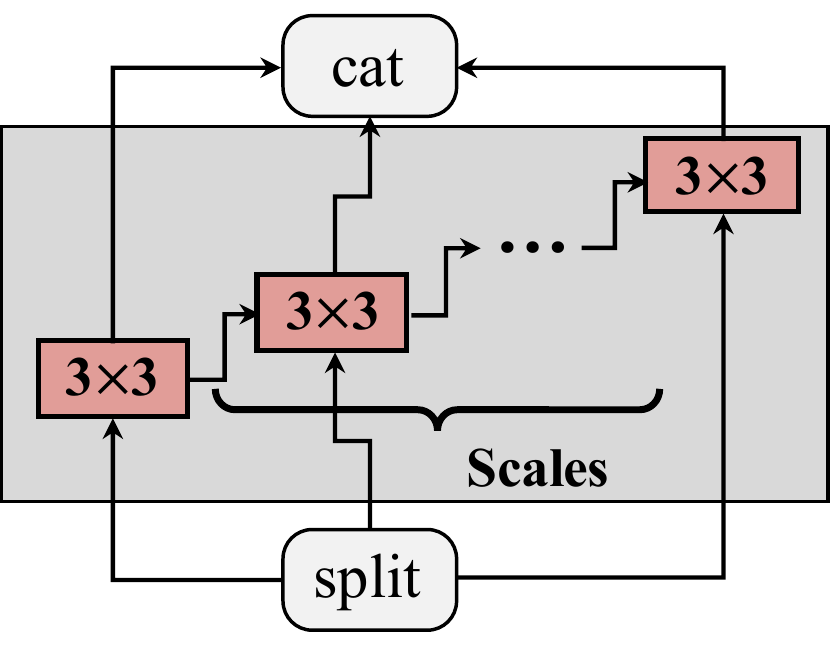}
    \caption{\scriptsize Hierarchical convolution \label{fig:hierarchical}}
    \end{subfigure}\hfill
    \begin{subfigure}[b]{0.235\textwidth}
    \centering
    \includegraphics[width=0.95\textwidth]{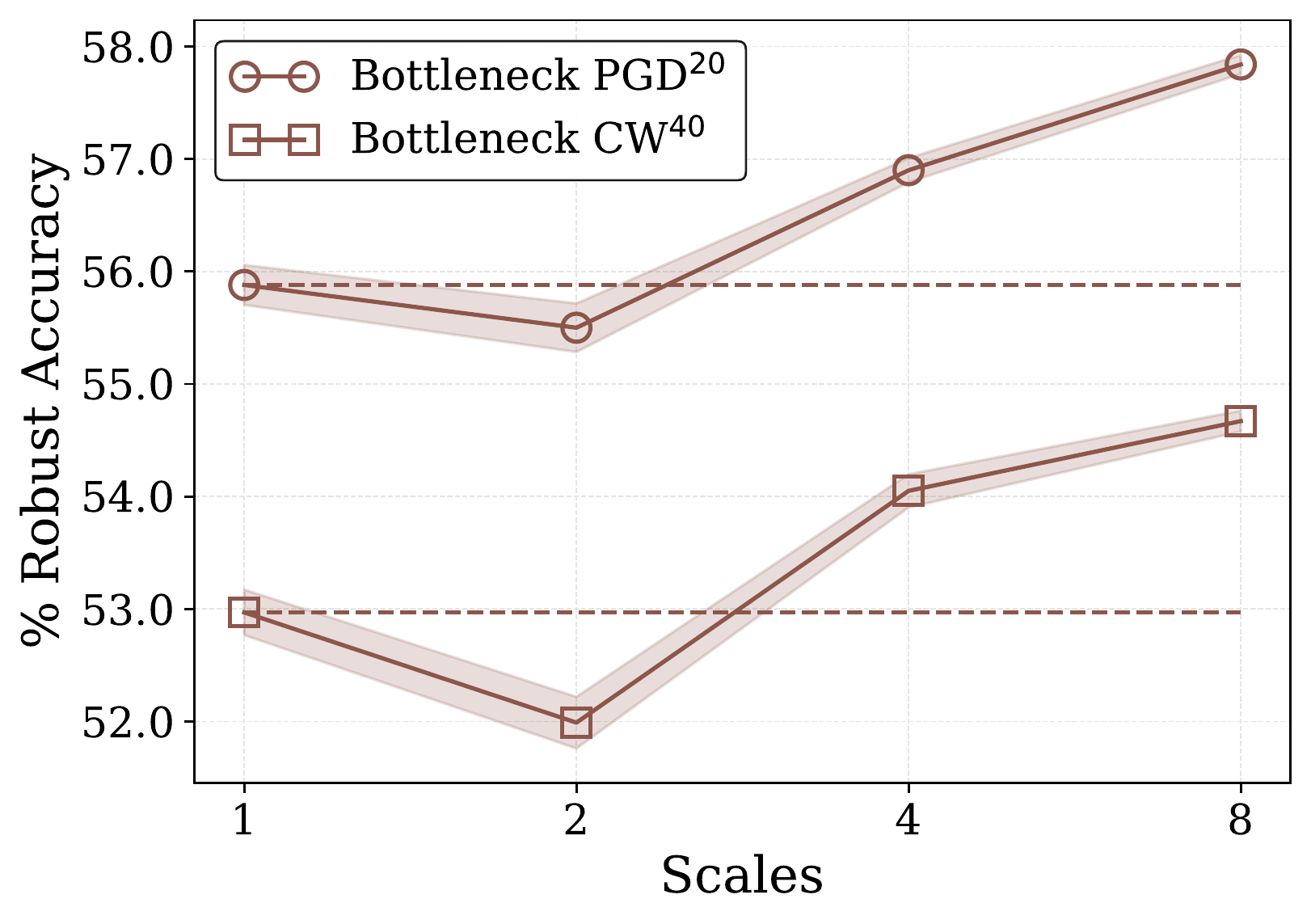}
    \caption{\scriptsize $D_i=5, W_i=12$ \label{fig:abl_hier_d34_w12_c10}}
    \end{subfigure} \\
    \begin{subfigure}[b]{0.235\textwidth}
    \centering
    \includegraphics[width=0.95\textwidth]{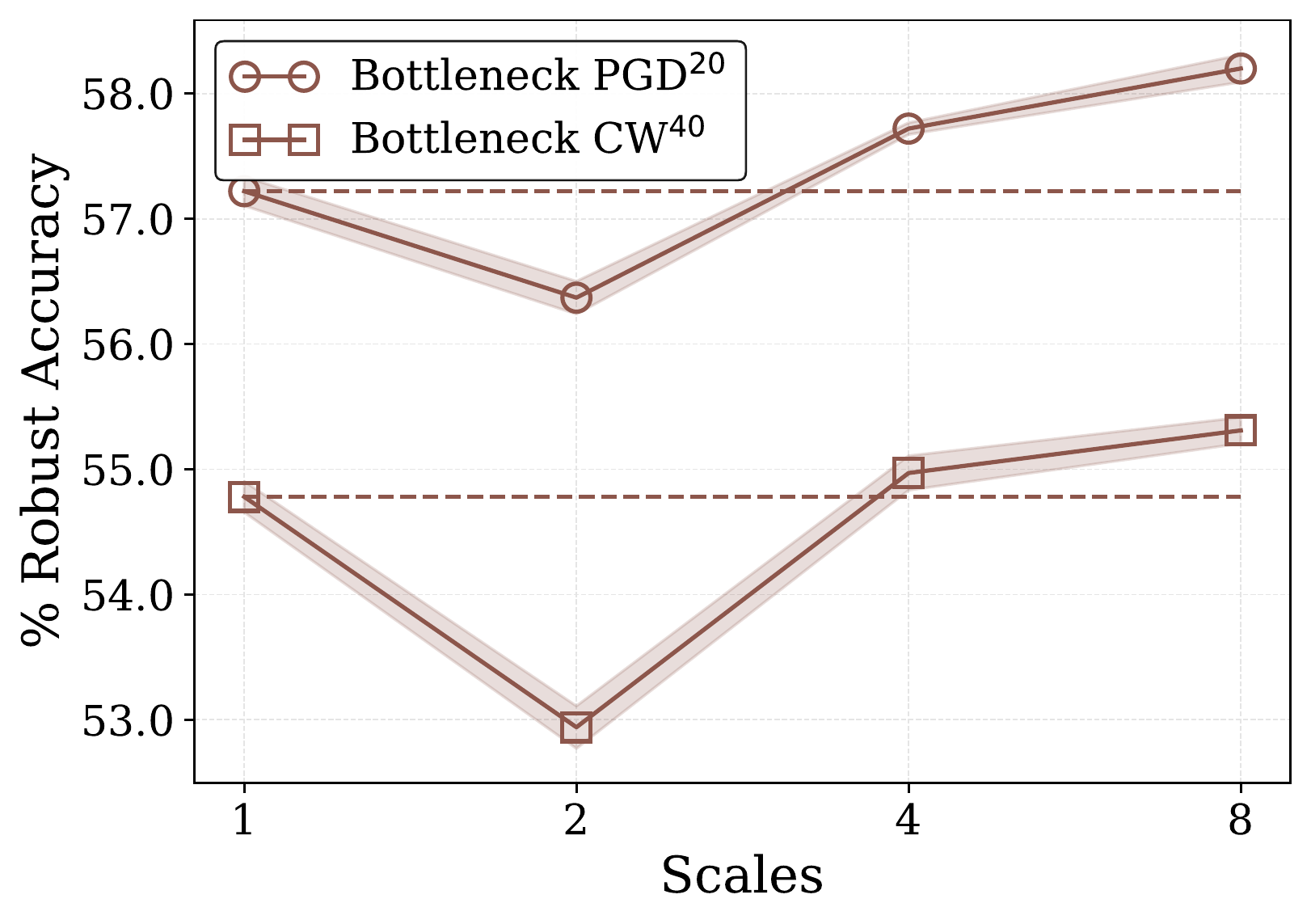}
    \caption{\scriptsize $D_i=7, W_i=14$  \label{fig:abl_hier_d46_w14_c10}}
    \end{subfigure}\hfill
    \begin{subfigure}[b]{0.235\textwidth}
    \centering
    \includegraphics[width=0.95\textwidth]{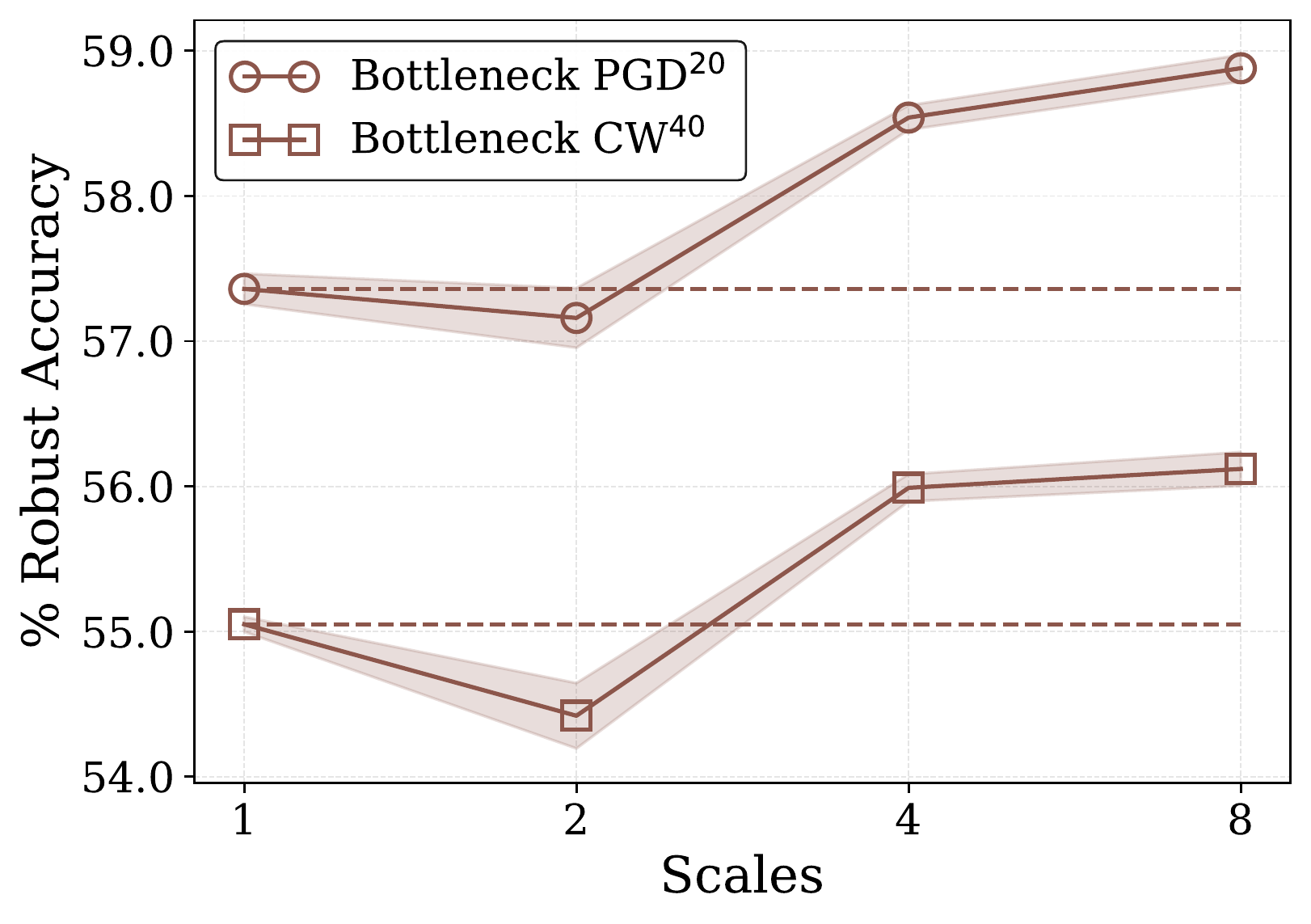}
    \caption{\scriptsize $D_i=11, W_i=16$  \label{fig:abl_hier_d70_w16_c10}}
    \end{subfigure}
    \vspace{-8pt}
    \caption{
    (a) Hierarchical convolution that splits a regular convolution into multiple hierarchically connected convolutions (scales). Results are then concatenated. (b, c, d) show the robustness of models from three different capacity regions. \label{fig:app_abl_hier} \vspace{-1em}}
\end{figure}

\begin{figure}[!h]
    \begin{subfigure}[b]{0.23\textwidth}
    \centering
    \includegraphics[width=0.95\textwidth]{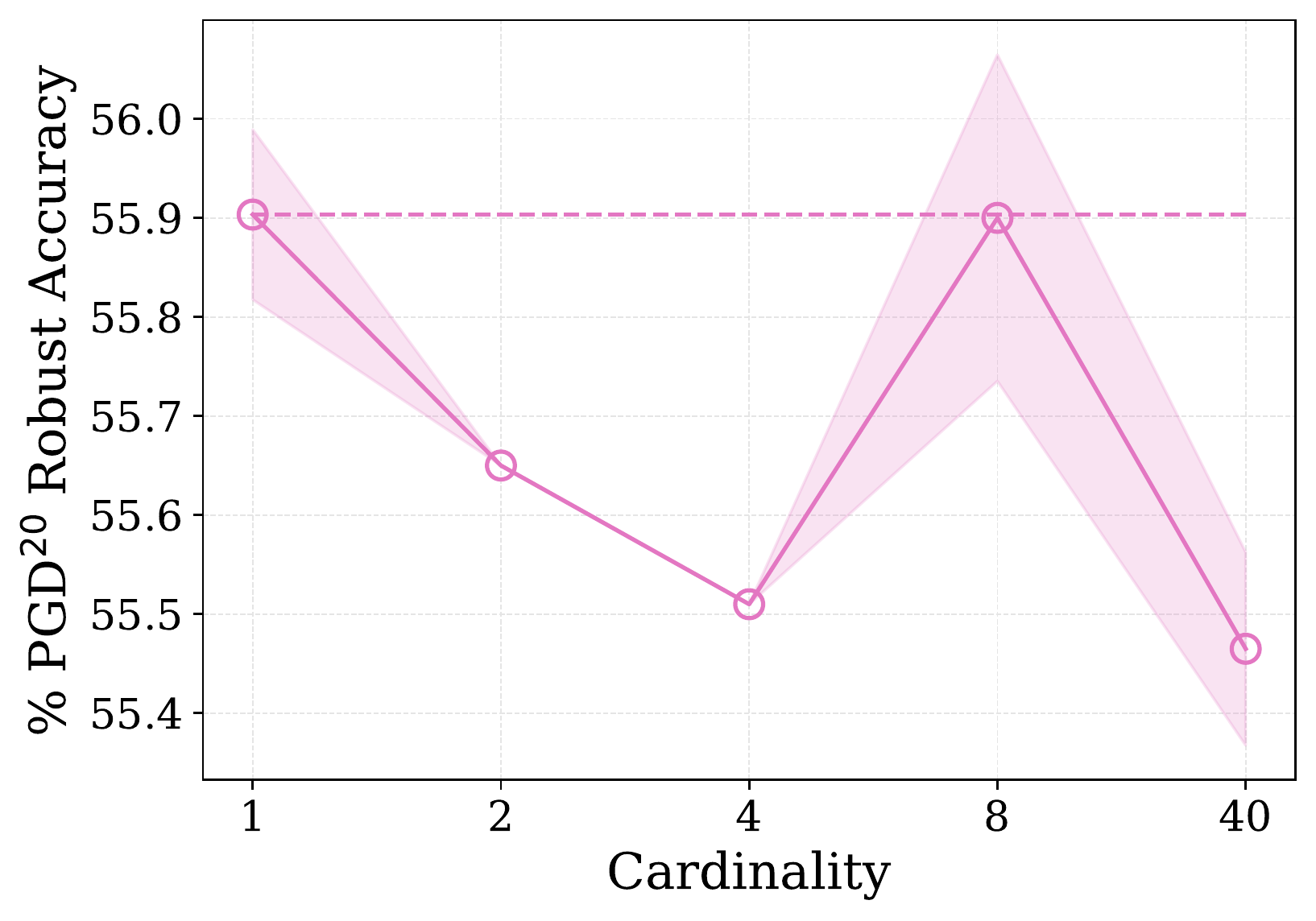}
    \caption{\scriptsize PGD$^{20}$ \label{fig:abl_basic_aggre_d28_w10_c10_pgd20}}
    \end{subfigure}\hfill
    \begin{subfigure}[b]{0.235\textwidth}
    \centering
    \includegraphics[width=0.95\textwidth]{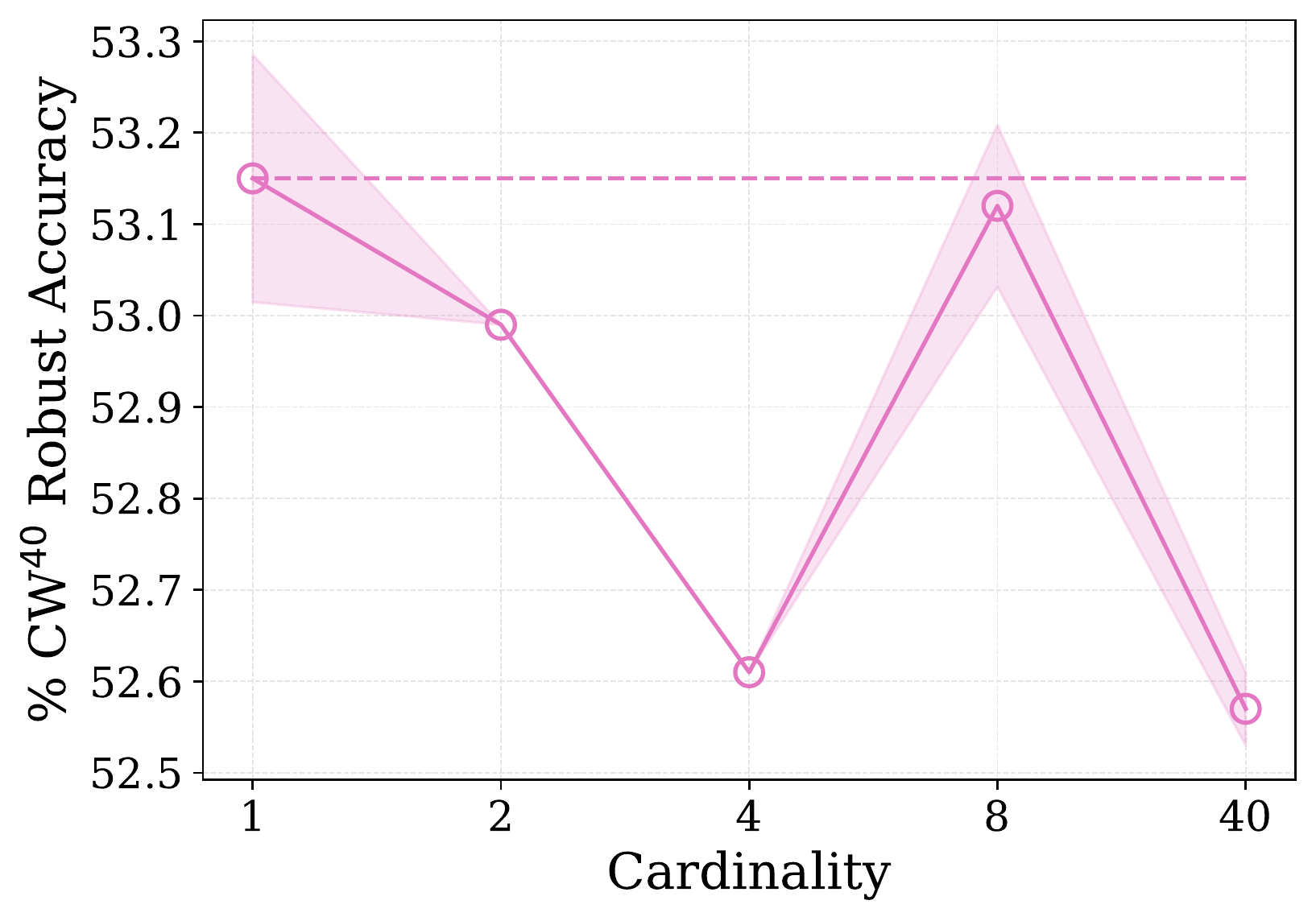}
    \caption{\scriptsize CW$^{40}$ \label{fig:abl_basic_aggre_d28_w10_c10_cw40}}
    \end{subfigure}
    \vspace{-8pt}
    \caption{
    The impact of aggregated convolution for the basic block. Results show the robustness of the model with $D_i=4, W_i=10$. \label{fig:app_abl_aggre_basic} \vspace{-1em}}
\end{figure}

This section presents pictorial illustrations of aggregated and hierarchical convolutions in Figures \ref{fig:aggregated} and \ref{fig:hierarchical}, respectively. Additional results showing the effects of hyperparameters cardinality (for aggregated convolution) and scales (for hierarchical convolution) are presented in Figures \ref{fig:app_abl_aggre} (b, c, d) and \ref{fig:app_abl_hier} (b, c, d). Finally, we show the impact of aggregated convolution for the basic block in Figure~\ref{fig:app_abl_aggre_basic}, where we observe that aggregated convolution adversely affects the robustness of the basic block. 

% -------------------------------------------------------------------------------------
% -------------------------------------------------------------------------------------
% -------------------------------------------------------------------------------------
% -------------------------------------------------------------------------------------
% -------------------------------------------------------------------------------------
% -------------------------------------------------------------------------------------

\subsection{Impact of Normalization \label{sec:app_normalization}}
This section investigates the relationship between normalization methods and adversarial robustness. In addition to the baseline of Batch Normalization (BN), we consider three other normalization methods, i.e., Group Normalization (GN) \cite{wu2018group}, and Instance Normalization (IN) \cite{ulyanov2016instance}. We also confine all blocks in a DNN model to use a single choice of normalization method and repeat the experiment for each technique three times. The experimental results are summarized in Table~\ref{tab:abl_norm}. The baseline normalization method (i.e., BN) outperforms all other alternative normalization methods, particularly on Tiny-ImageNet. 

\begin{table}[ht]
\centering
\caption{The adversarial robustness of the considered normalization methods. We highlight the best results of each section in bold. \label{tab:abl_norm} \vspace{-1em}}
\resizebox{.49\textwidth}{!}{%
\begin{tabular}{@{\hspace{2mm}}l|cccc|cccc|ccc|c@{\hspace{2mm}}}
\toprule
{\multirow{2}{*}{}} & \multicolumn{4}{c|}{CIFAR-10} & \multicolumn{4}{c|}{CIFAR-100} & \multicolumn{3}{c|}{Tiny-ImageNet} & \multirow{2}{*}{\begin{tabular}[c]{@{}c@{}}Ave. \\ Rank\end{tabular}} \\ \cmidrule(lr){2-5} \cmidrule(lr){6-9} \cmidrule(lr){10-12}
& Nat. & PGD$^{20}$ & CW$^{40}$ & AA & Nat. & PGD$^{20}$ & CW$^{40}$ & AA & Nat. & PGD$^{20}$ & AA & \\ \midrule
BN & 85.11 & {55.36} & \textbf{53.02} & {\textbf{51.43}} & 55.77 & {\textbf{29.91}} & 26.23 & {\textbf{25.35}} & {\textbf{42.09}} & {\textbf{20.68}} & {\textbf{16.25}} & \textbf{1.5} \\ 
GN & {85.28} & {\textbf{55.82}} & 52.76 & {51.23} & {\textbf{56.60}} & {29.86} & \textbf{26.26} & {25.09} & {30.99} & {16.87} & {13.01} & 1.7 \\
IN & {\textbf{85.34}} & 54.49 & 50.82 & 49.34 & {56.56} & 28.41 & 24.17 & 22.68 & 17.25 & 10.69 & 8.18 & 2.7 \\
% & LN & - & - & - & - & - & - & - & - & - & - & - & 4.0 \\
\bottomrule
\end{tabular}%
}
\end{table}

% -------------------------------------------------------------------------------------
\begin{figure*}[!ht]
    \begin{subfigure}[b]{0.235\textwidth}
    \centering
    \includegraphics[width=0.95\textwidth]{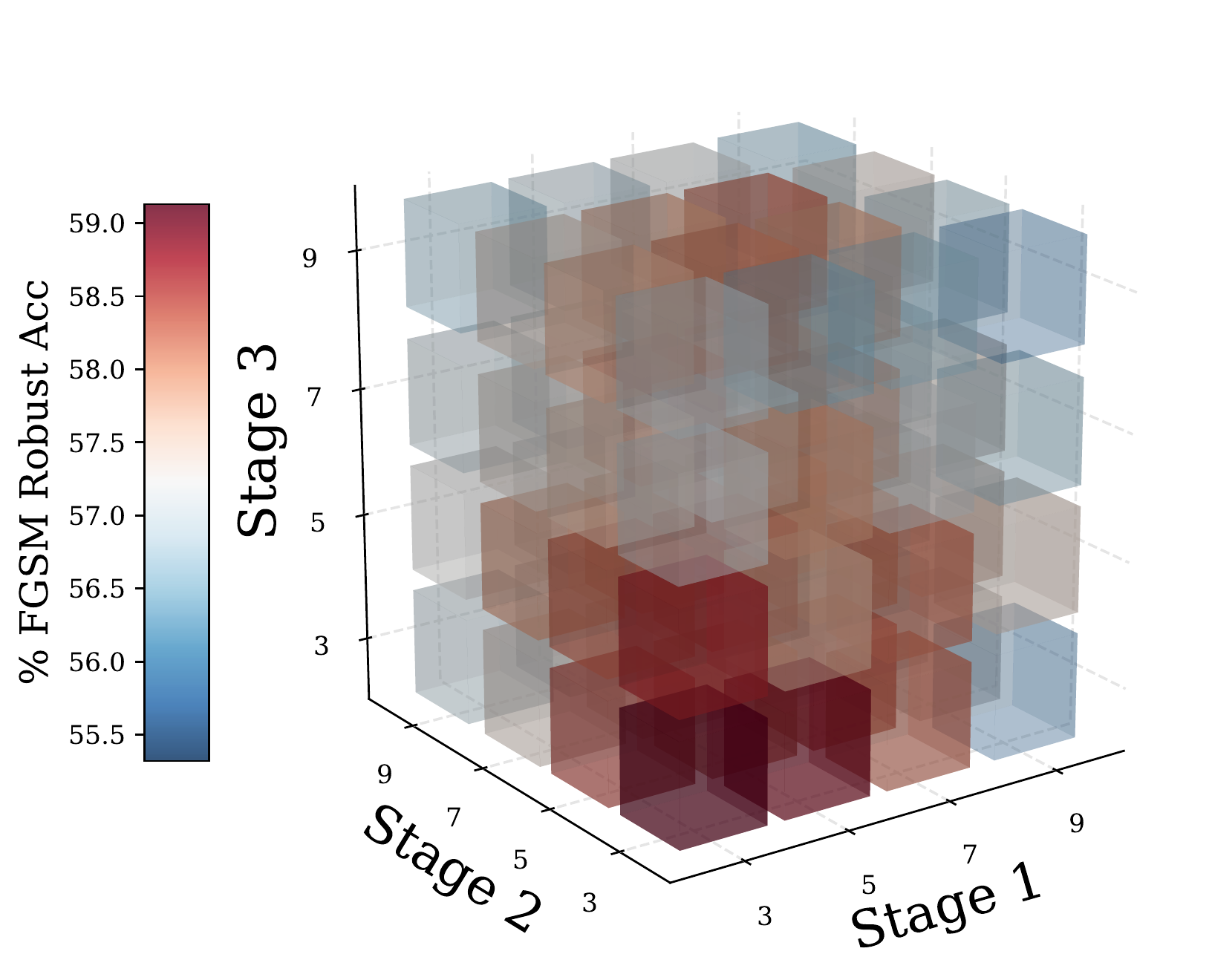}
    \caption{\scriptsize CIFAR-10, FGSM \vspace{-5pt}}
    \end{subfigure}\hfill
    \begin{subfigure}[b]{0.235\textwidth}
    \centering
    \includegraphics[width=0.95\textwidth]{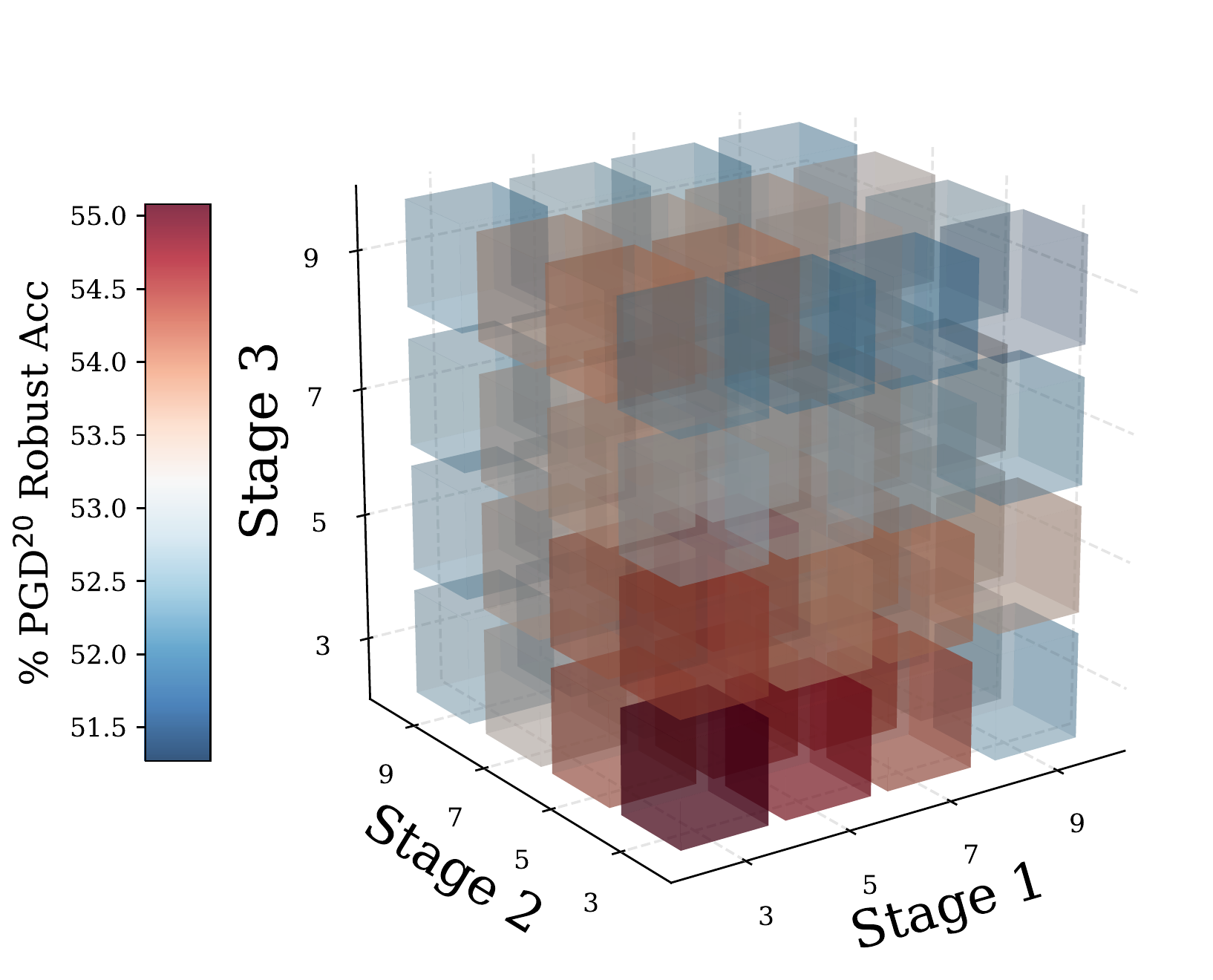}
    \caption{\scriptsize CIFAR-10, PGD$^{20}$ \vspace{-5pt}}
    \end{subfigure}\hfill
    \begin{subfigure}[b]{0.235\textwidth}
    \centering
    \includegraphics[width=0.95\textwidth]{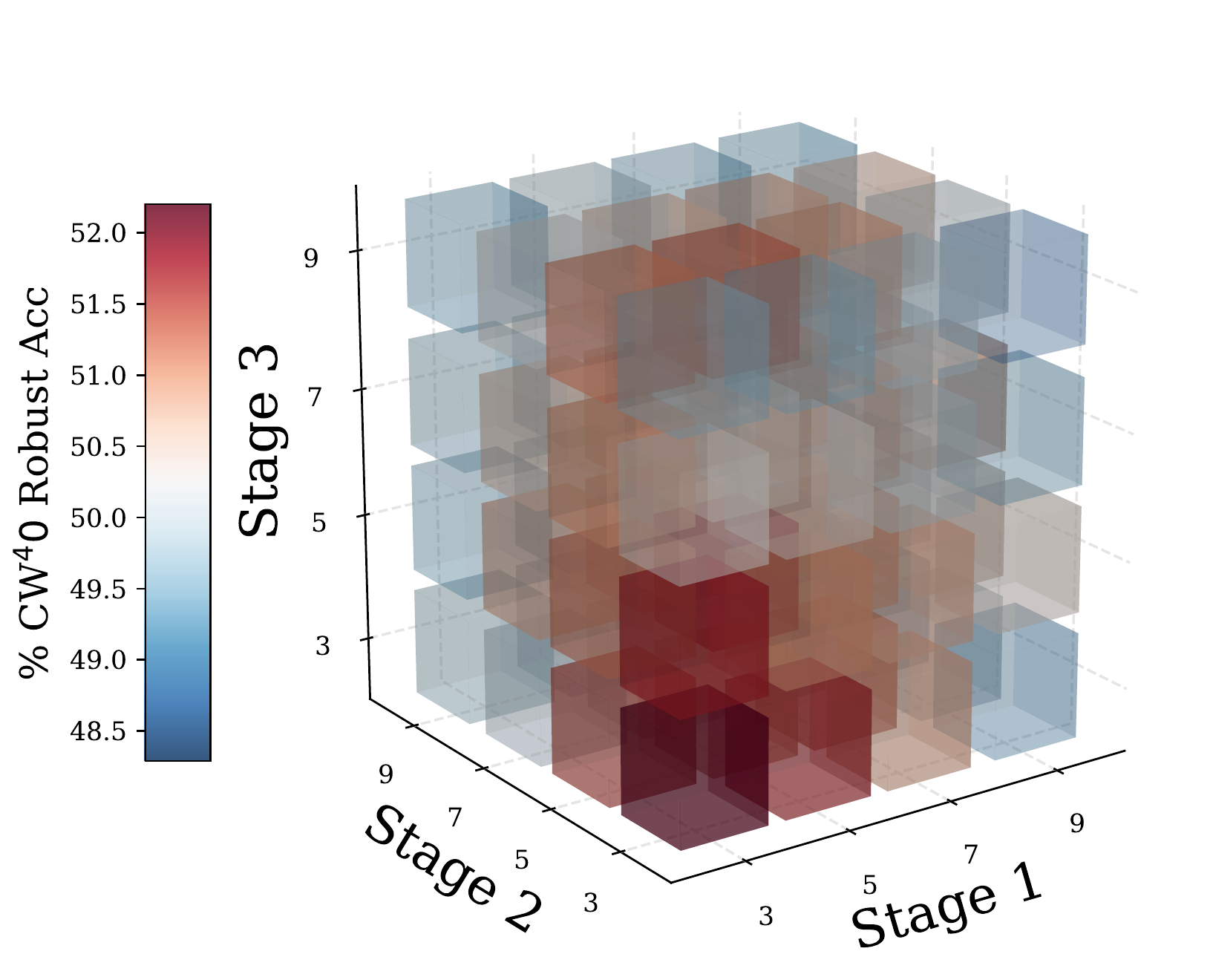}
    \caption{\scriptsize CIFAR-10, CW$^{40}$ \vspace{-5pt}}
    \end{subfigure}\hfill
    \begin{subfigure}[b]{0.235\textwidth}
    \centering
    \includegraphics[width=0.95\textwidth]{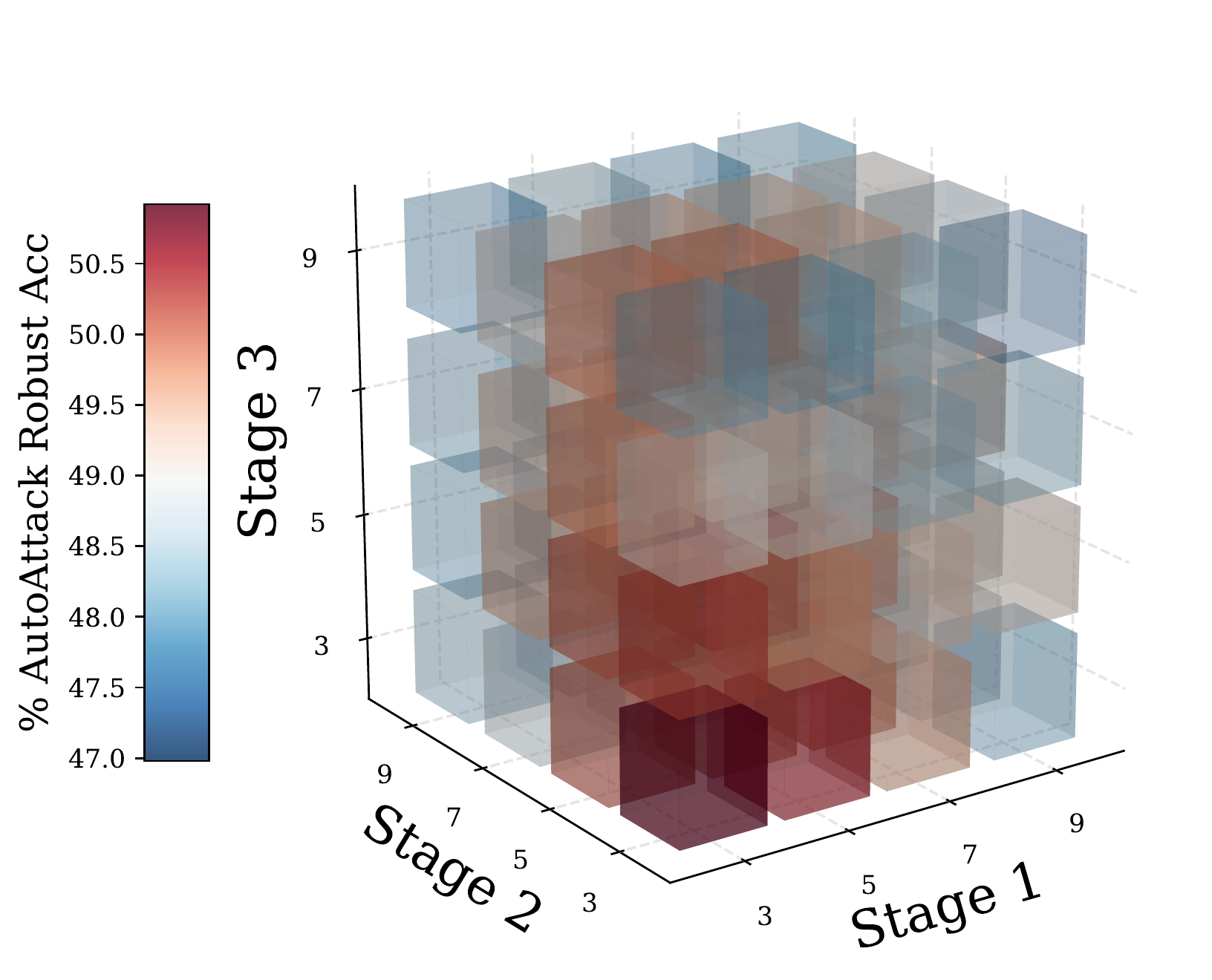}
    \caption{\scriptsize CIFAR-10, AutoAttack \vspace{-5pt}}
    \end{subfigure}\\
    \begin{subfigure}[b]{0.235\textwidth}
    \centering
    \includegraphics[width=0.95\textwidth]{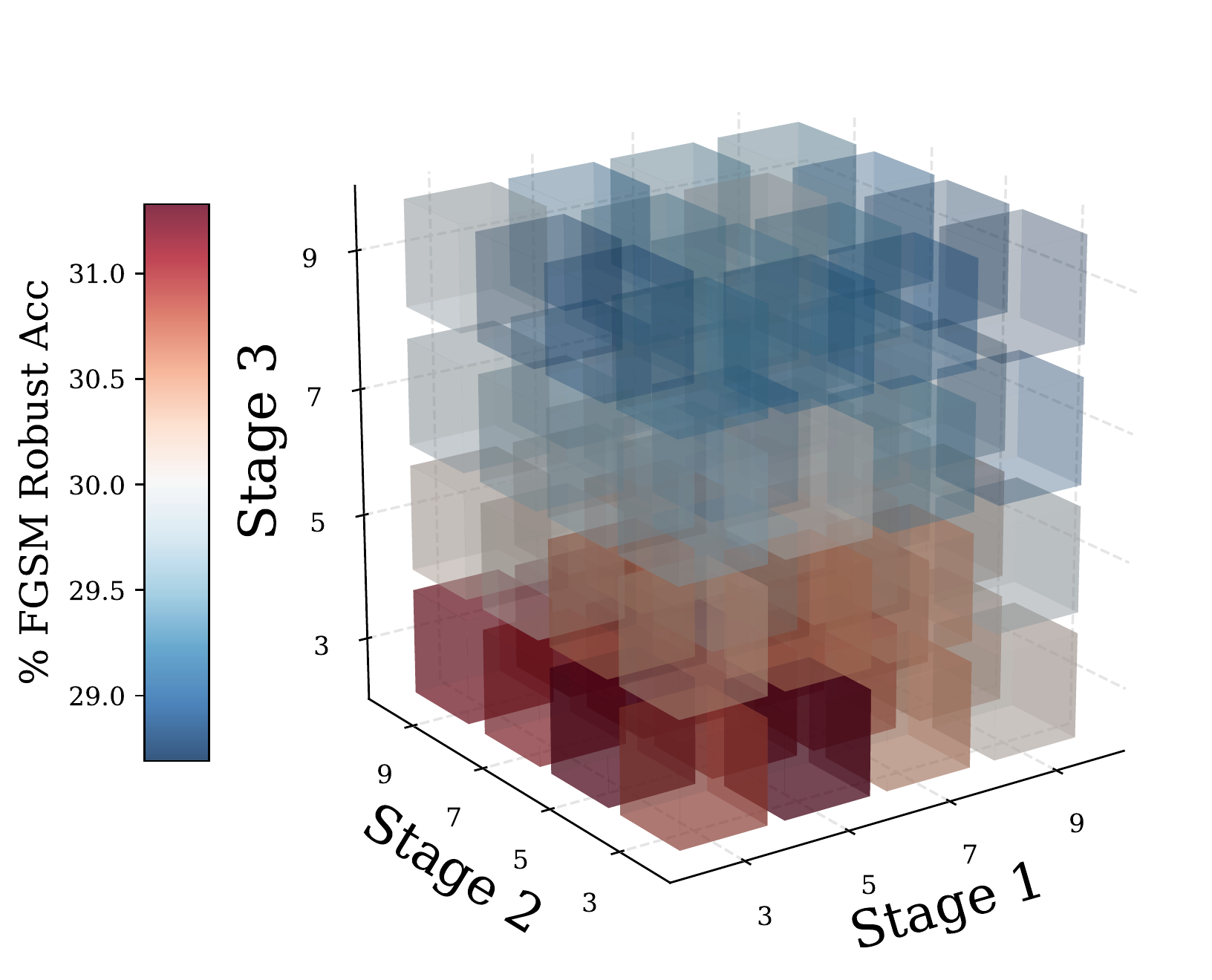}
    \caption{\scriptsize CIFAR-100, FGSM \vspace{-5pt}}
    \end{subfigure}\hfill
    \begin{subfigure}[b]{0.235\textwidth}
    \centering
    \includegraphics[width=0.95\textwidth]{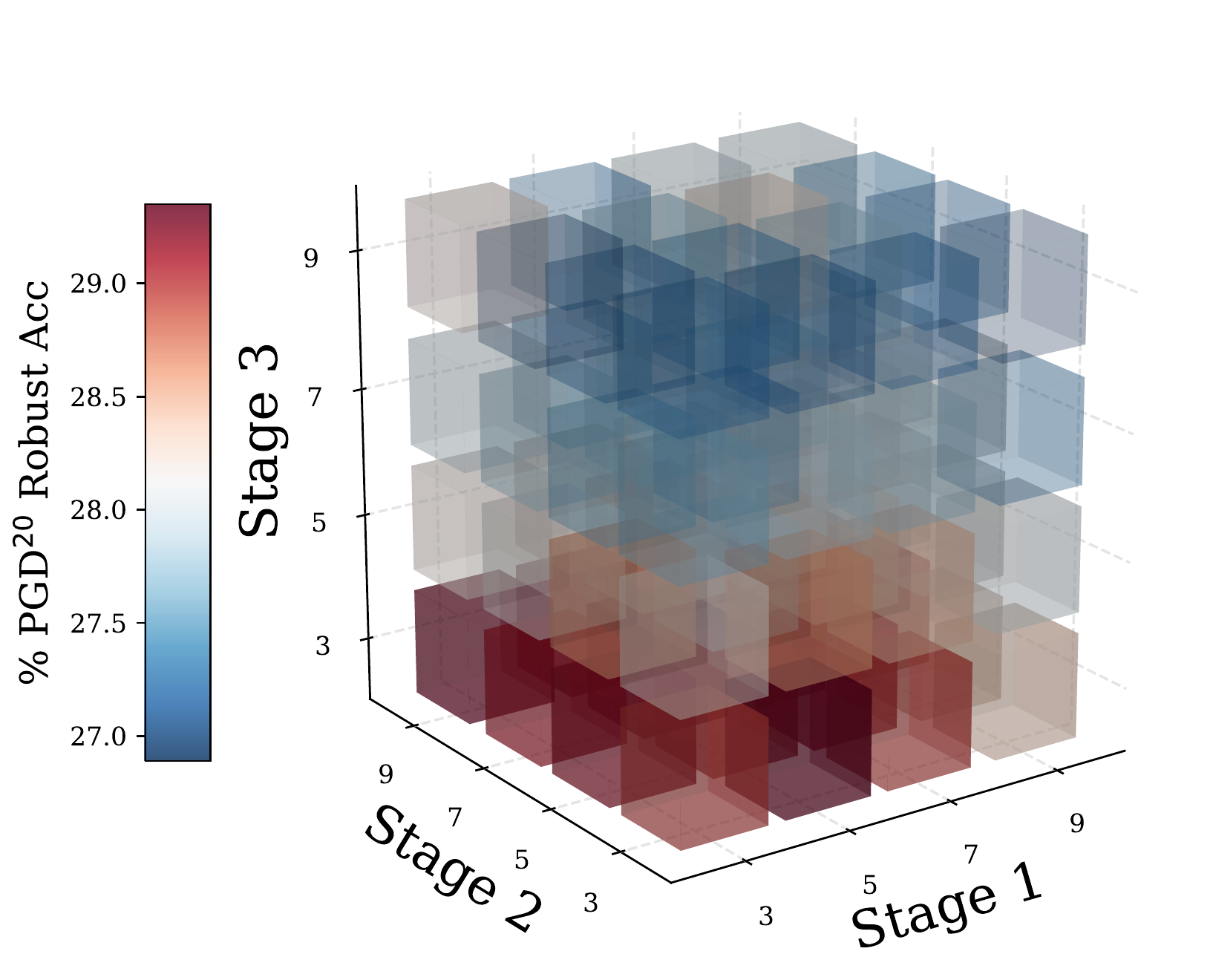}
    \caption{\scriptsize CIFAR-100, PGD$^{20}$ \vspace{-5pt}}
    \end{subfigure}\hfill
    \begin{subfigure}[b]{0.235\textwidth}
    \centering
    \includegraphics[width=0.95\textwidth]{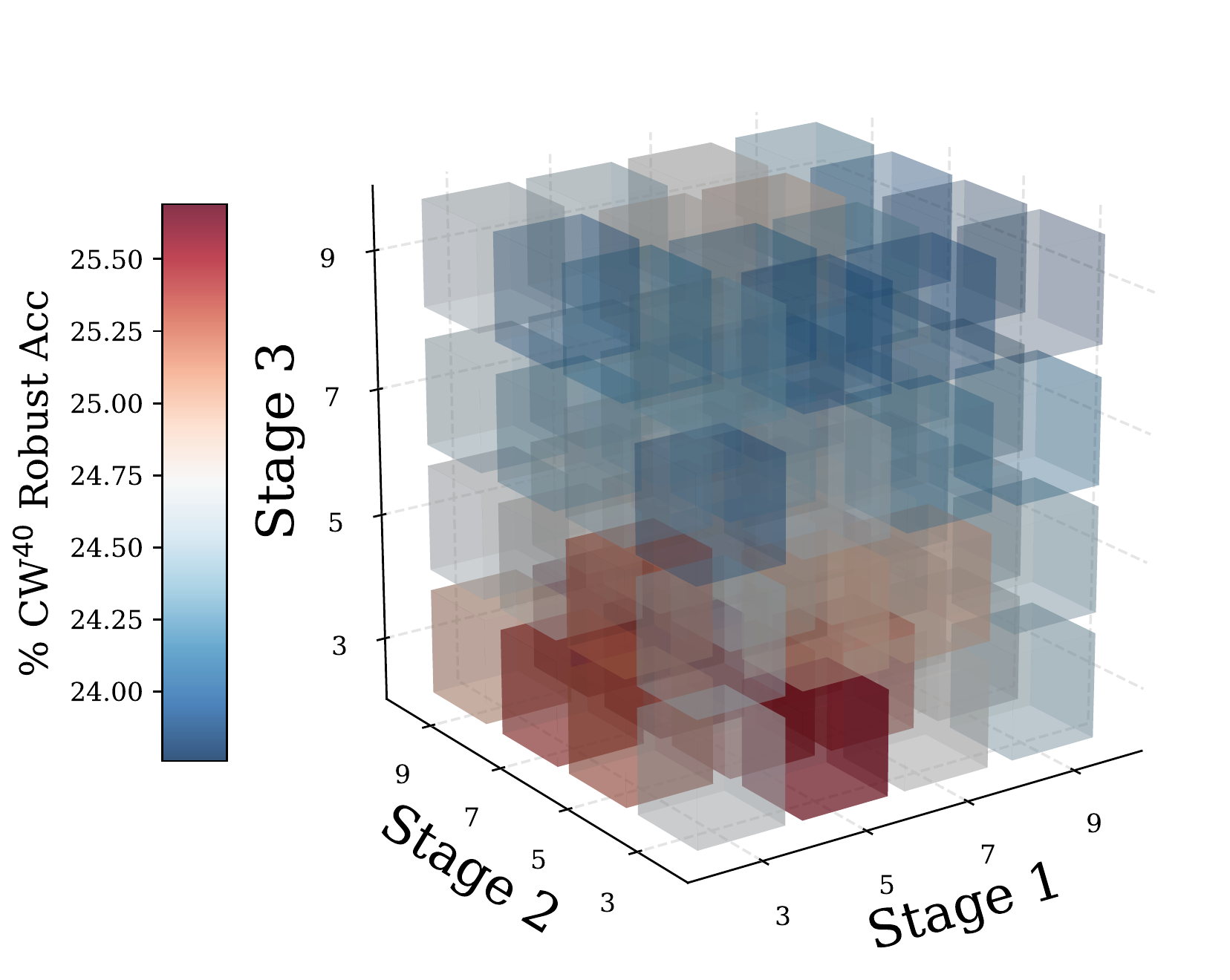}
    \caption{\scriptsize CIFAR-100, CW$^{40}$ \vspace{-5pt}}
    \end{subfigure}\hfill
    \begin{subfigure}[b]{0.235\textwidth}
    \centering
    \includegraphics[width=0.95\textwidth]{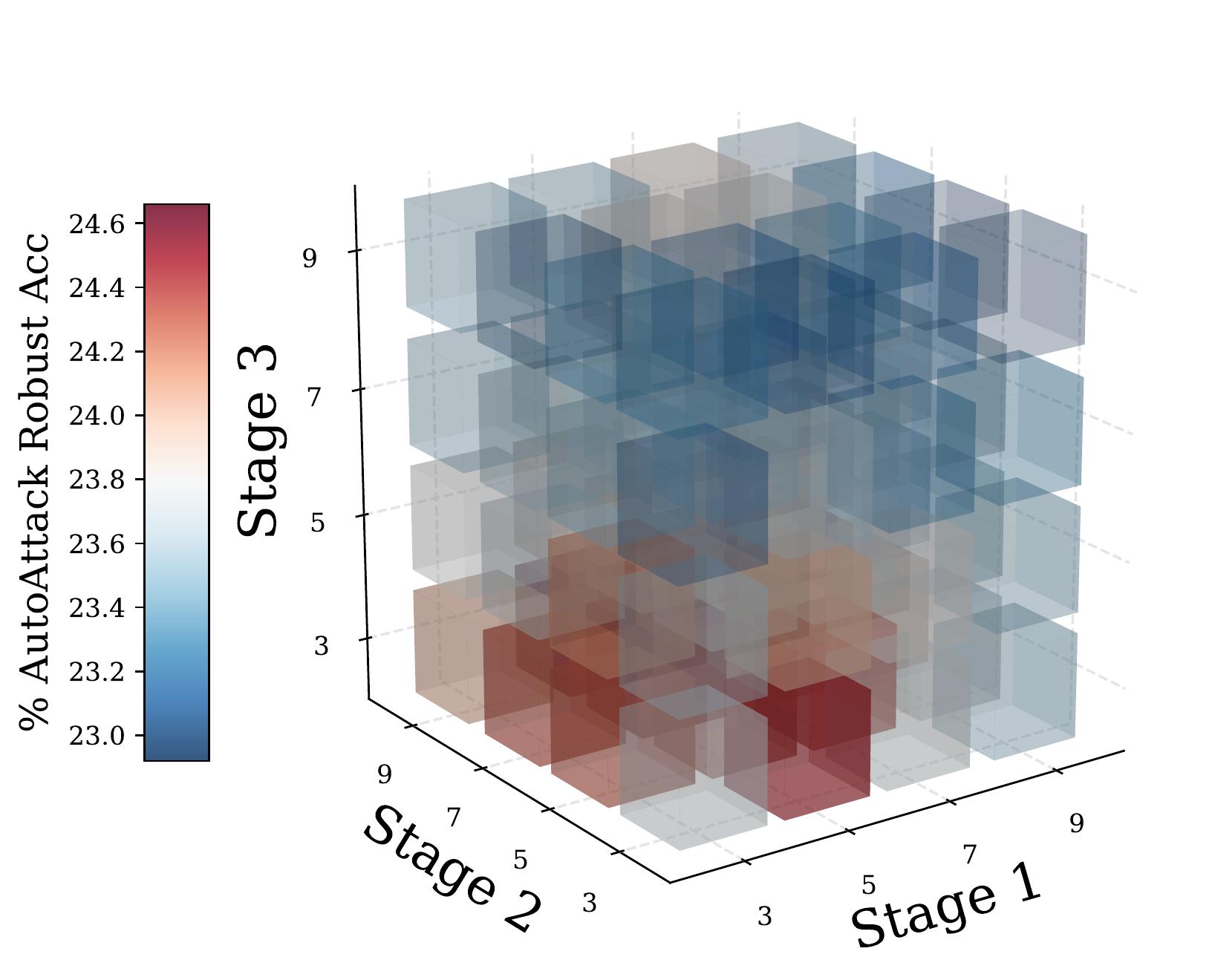}
    \caption{\scriptsize CIFAR-100, AutoAttack \vspace{-5pt}}
    \end{subfigure}\\
    \begin{subfigure}[b]{0.235\textwidth}
    \centering
    \includegraphics[width=0.95\textwidth]{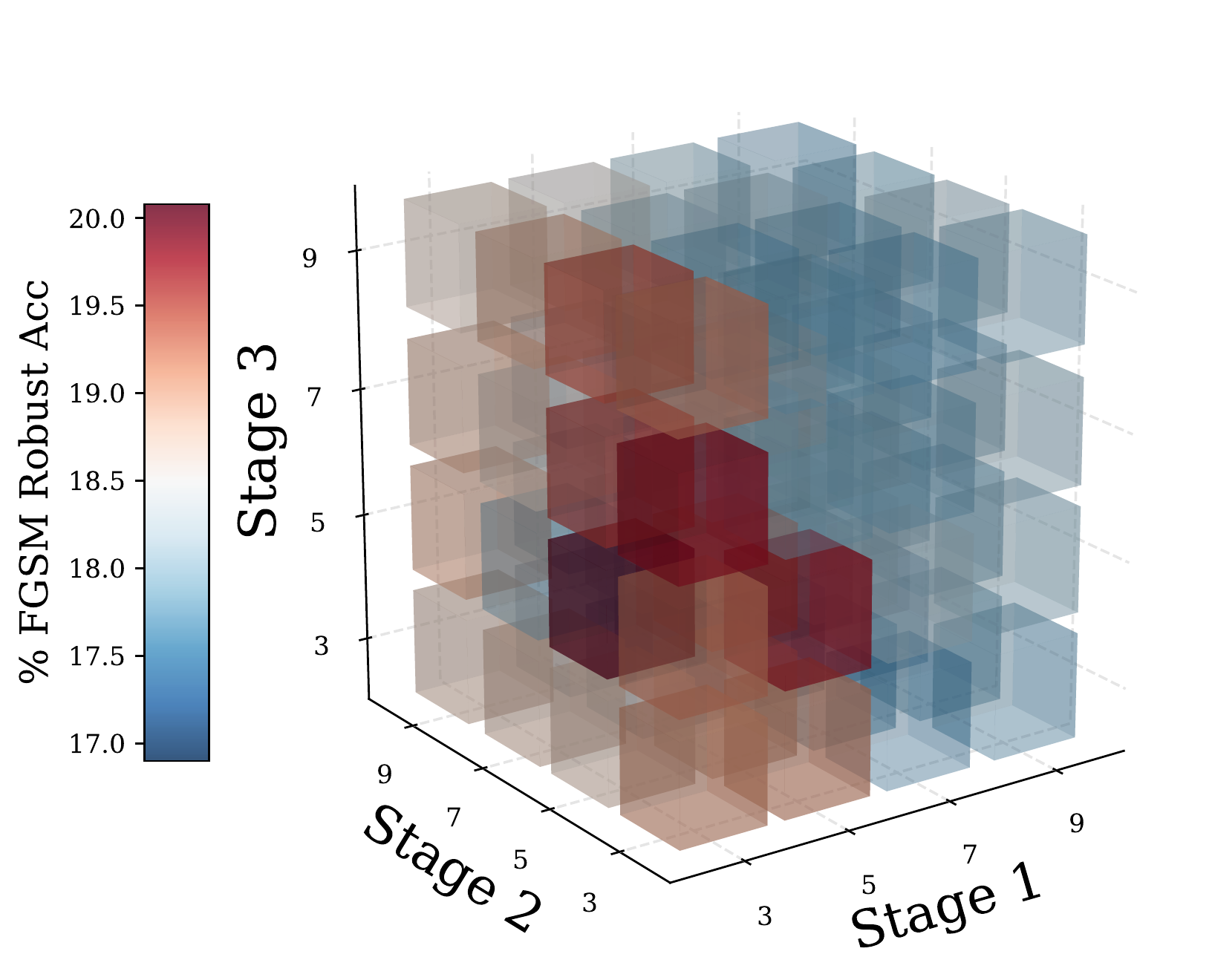}
    \caption{\scriptsize Tiny-ImageNet, FGSM \vspace{-5pt}}
    \end{subfigure}\hfill
    \begin{subfigure}[b]{0.235\textwidth}
    \centering
    \includegraphics[width=0.95\textwidth]{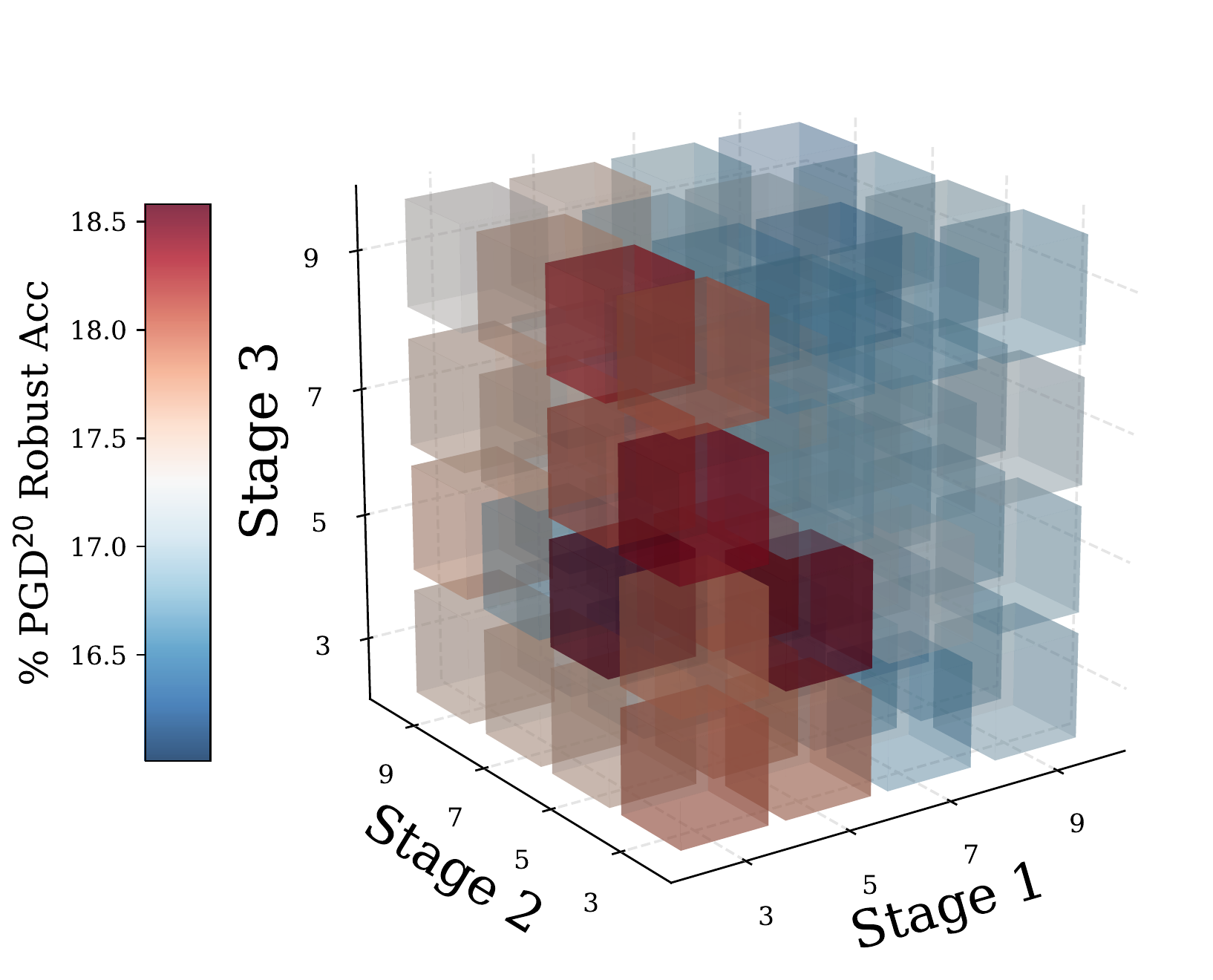}
    \caption{\scriptsize Tiny-ImageNet, PGD$^{20}$ \vspace{-5pt}}
    \end{subfigure}\hfill
    \begin{subfigure}[b]{0.235\textwidth}
    \centering
    \includegraphics[width=0.95\textwidth]{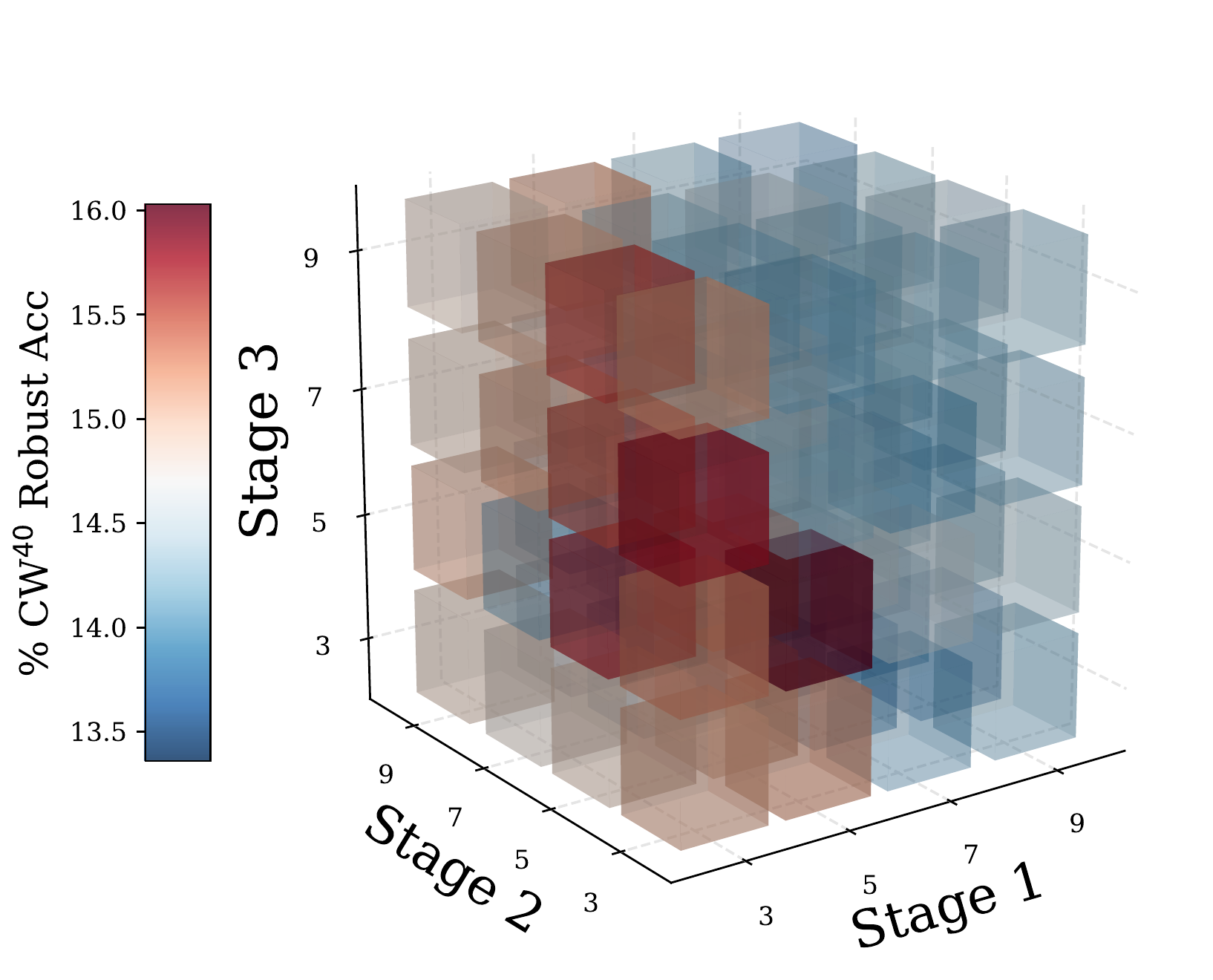}
    \caption{\scriptsize Tiny-ImageNet, CW$^{40}$ \vspace{-5pt}}
    \end{subfigure}\hfill
    \begin{subfigure}[b]{0.235\textwidth}
    \centering
    \includegraphics[width=0.95\textwidth]{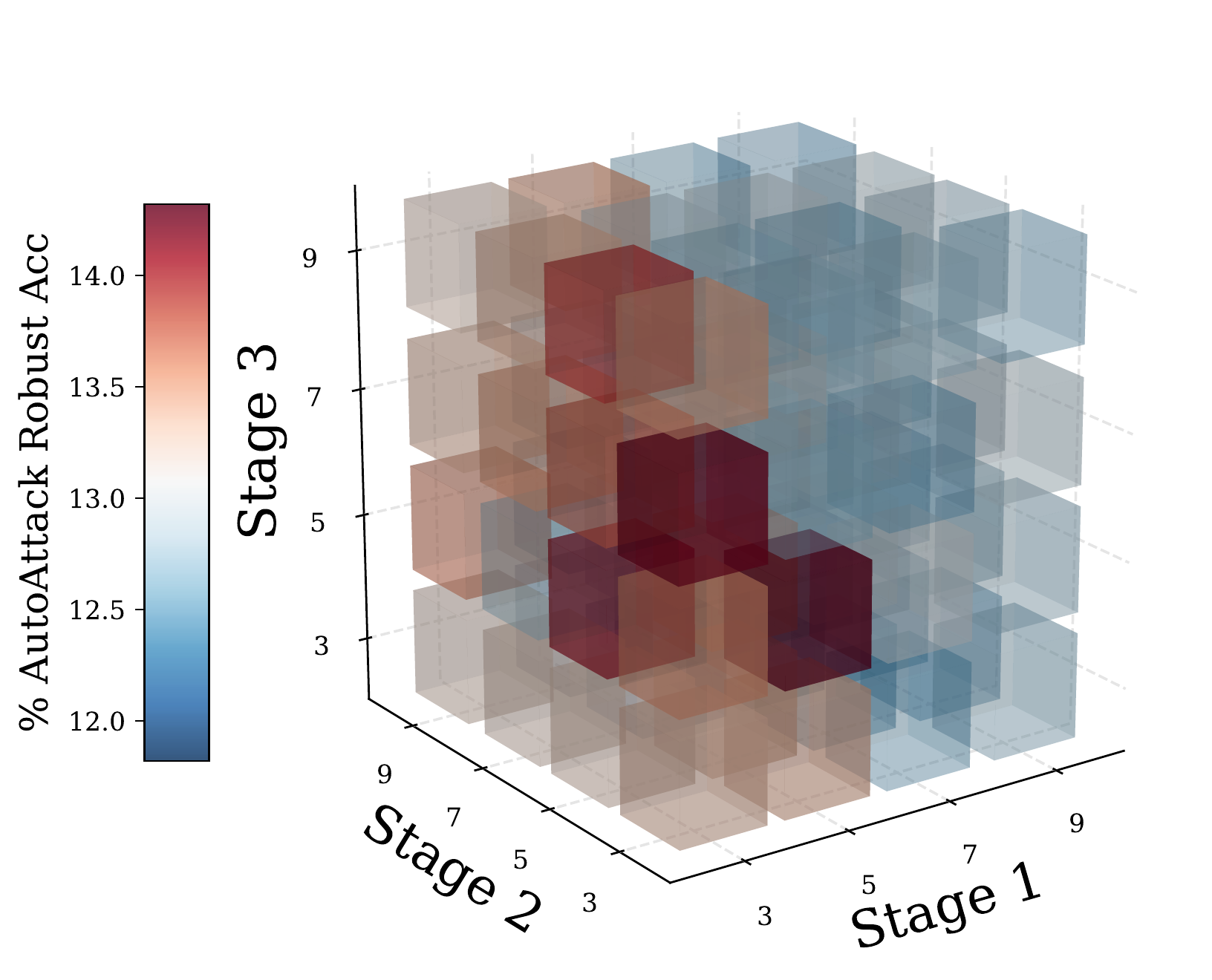}
    \caption{\scriptsize Tiny-ImageNet, AutoAttack \vspace{-5pt}}
    \end{subfigure}
    \caption{Heat maps visualizing the relationship between kernel sizes and adversarial robustness on CIFAR-10, CIFAR-100, and Tiny-ImageNet (from top to bottom) against FGSM, PGD$^{20}$, CW$^{40}$, and AutoAttack (from left to right). \label{fig:abl_kernel_resolution} \vspace{-1em}}
\end{figure*}
% -------------------------------------------------------------------------------------
% -------------------------------------------------------------------------------------
\begin{figure*}[ht]
    \begin{subfigure}[b]{0.235\textwidth}
    \centering
    \includegraphics[width=0.95\textwidth]{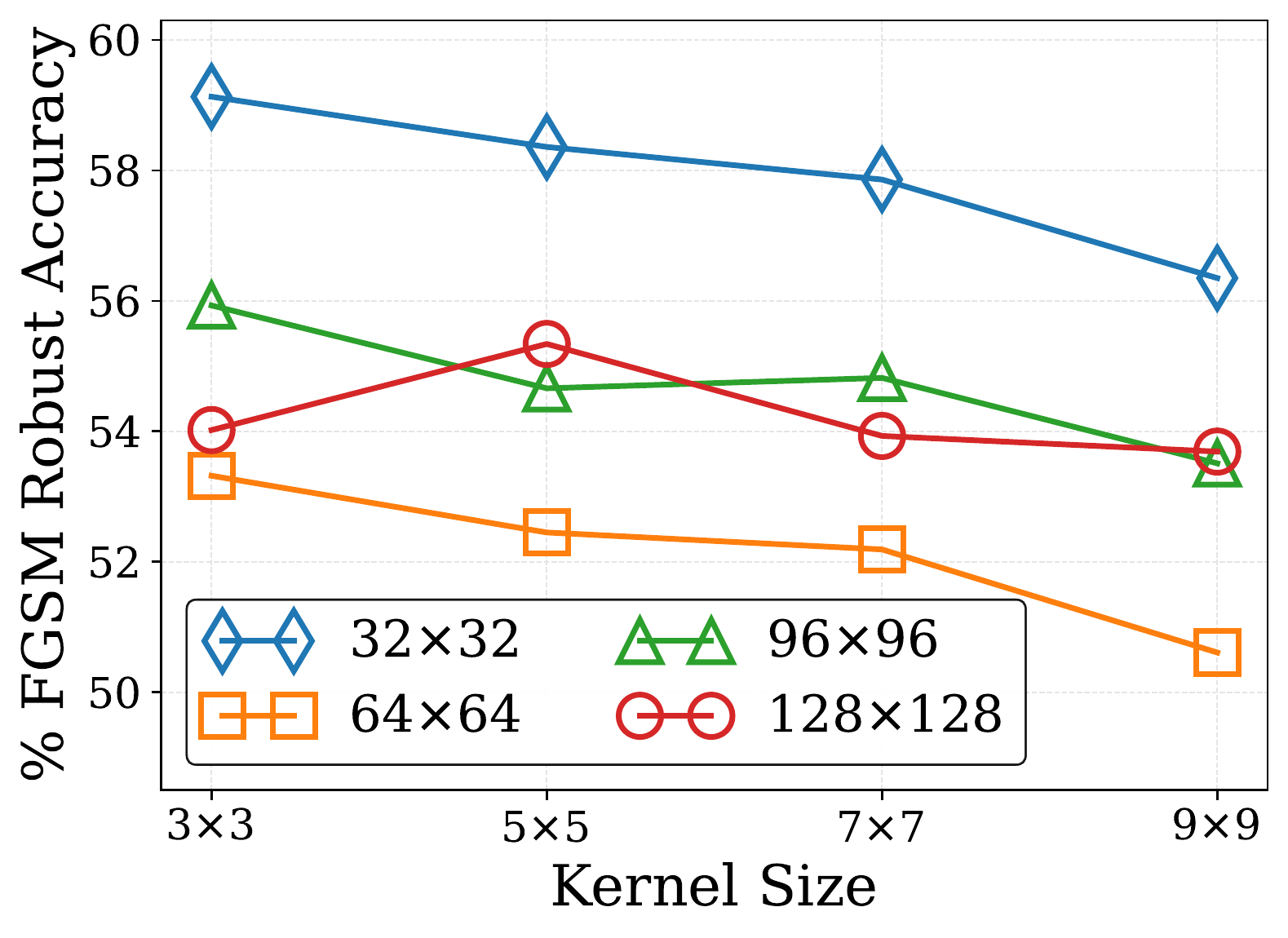}
    \caption{\scriptsize CIFAR-10, FGSM \vspace{-8pt}}
    \end{subfigure}\hfill
    \begin{subfigure}[b]{0.235\textwidth}
    \centering
    \includegraphics[width=0.95\textwidth]{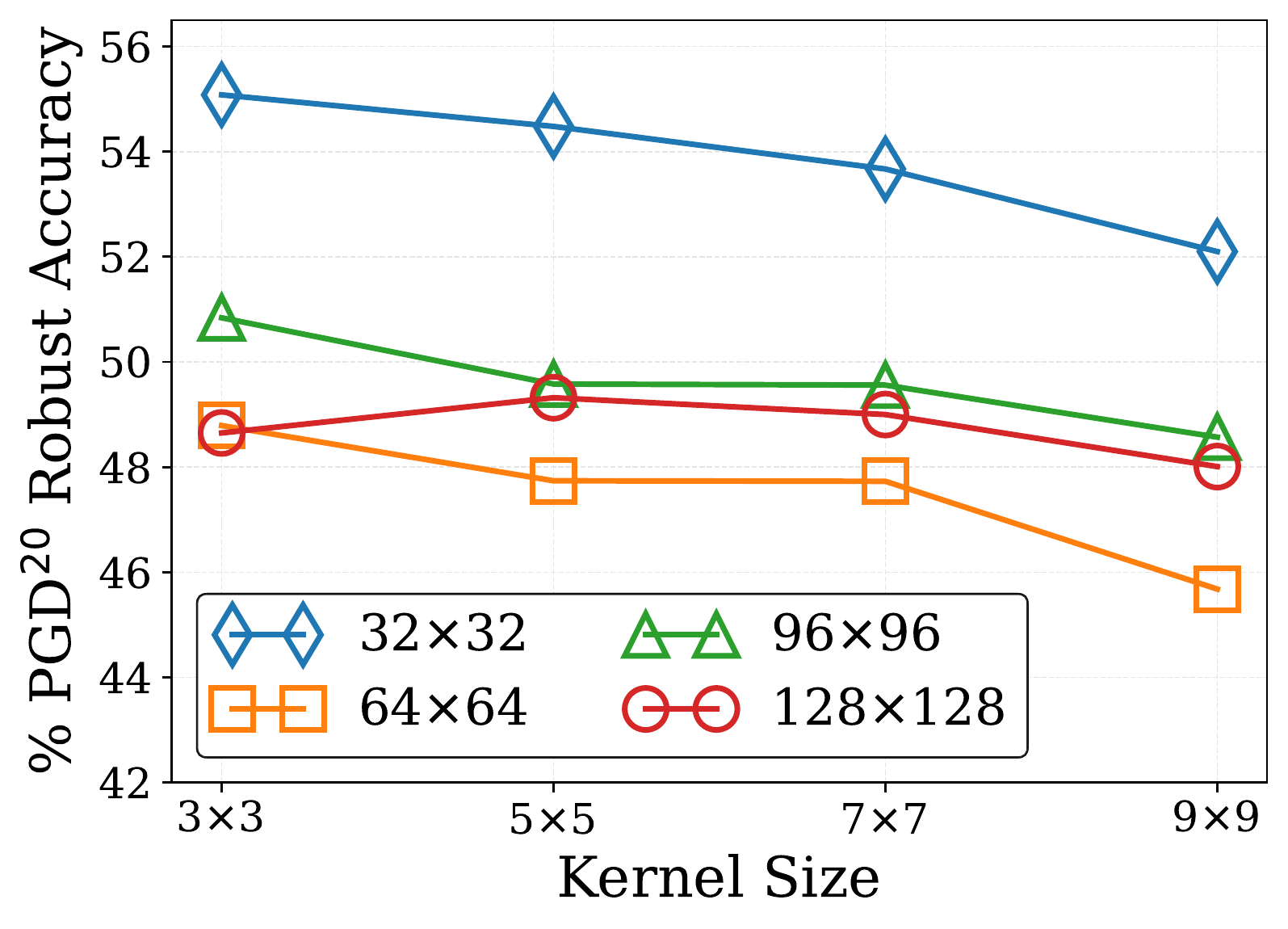}
    \caption{\scriptsize CIFAR-10, PGD$^{20}$ \vspace{-8pt}}
    \end{subfigure}\hfill
    \begin{subfigure}[b]{0.235\textwidth}
    \centering
    \includegraphics[width=0.95\textwidth]{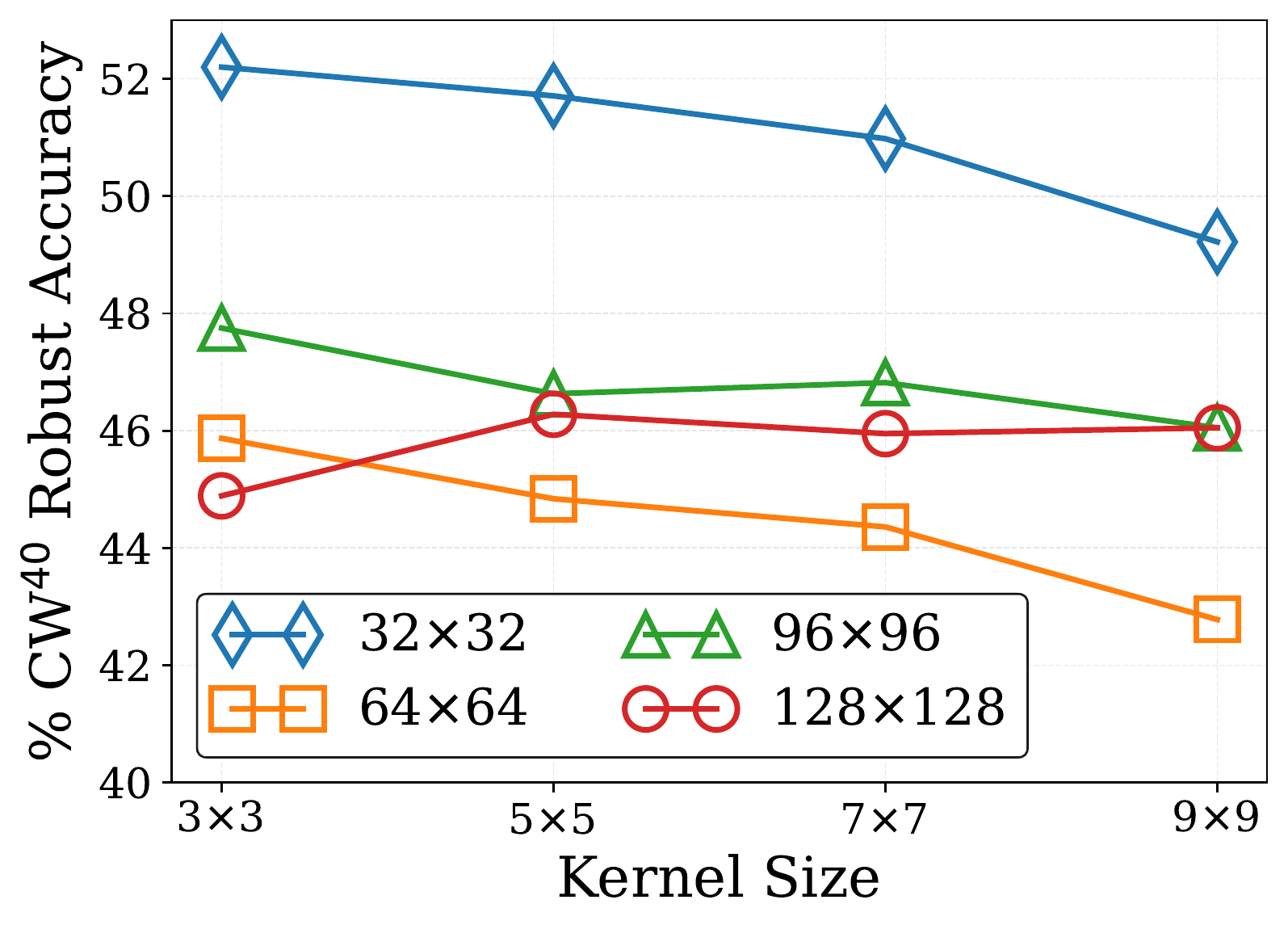}
    \caption{\scriptsize CIFAR-10, CW$^{40}$ \vspace{-8pt}}
    \end{subfigure}\hfill
    \begin{subfigure}[b]{0.235\textwidth}
    \centering
    \includegraphics[width=0.95\textwidth]{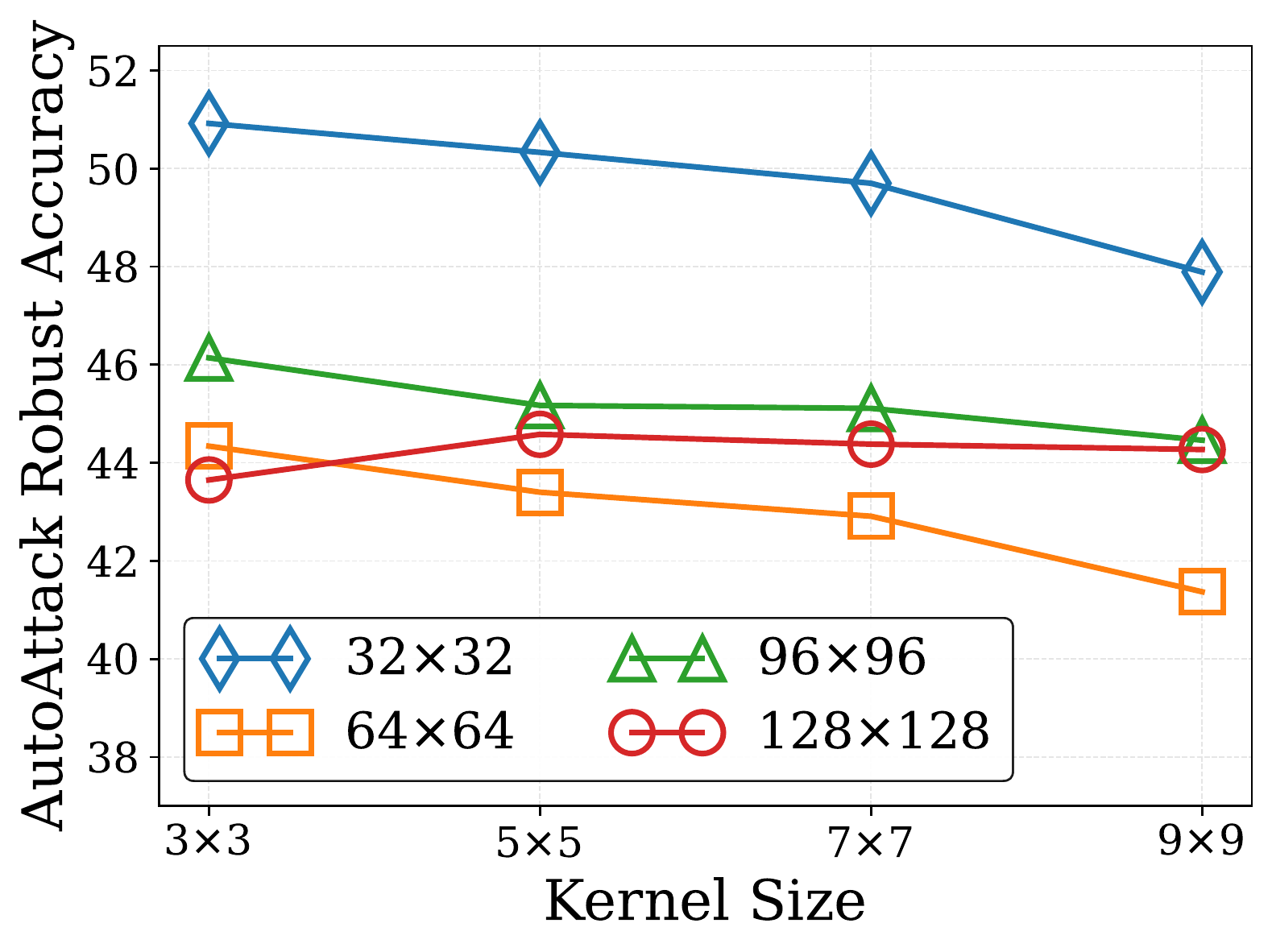}
    \caption{\scriptsize CIFAR-10, AutoAttack \vspace{-8pt}}
    \end{subfigure}\\
    \begin{subfigure}[b]{0.235\textwidth}
    \centering
    \includegraphics[width=0.95\textwidth]{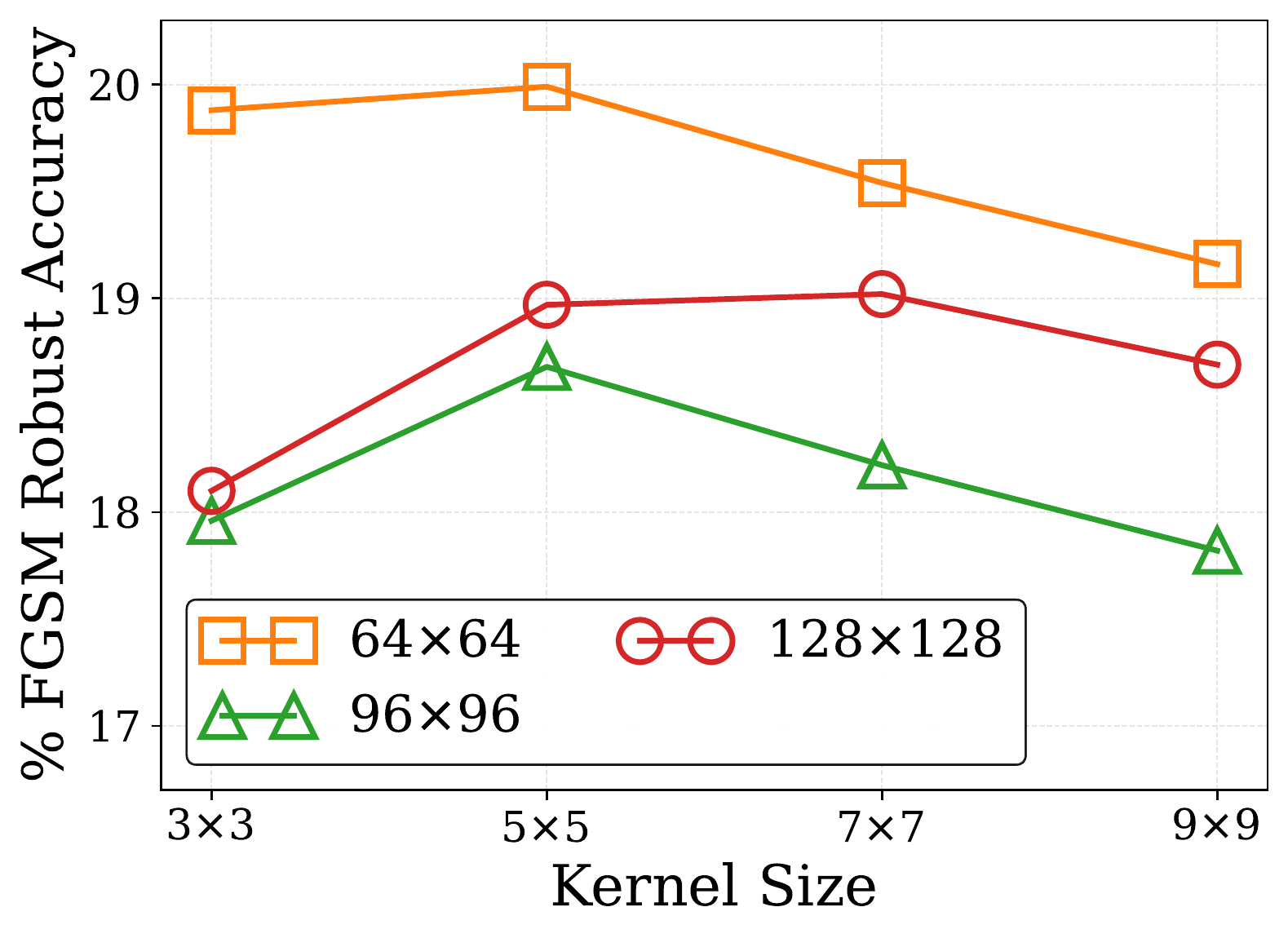}
    \caption{\scriptsize Tiny-ImageNet, FGSM \vspace{-8pt}}
    \end{subfigure}\hfill
    \begin{subfigure}[b]{0.235\textwidth}
    \centering
    \includegraphics[width=0.95\textwidth]{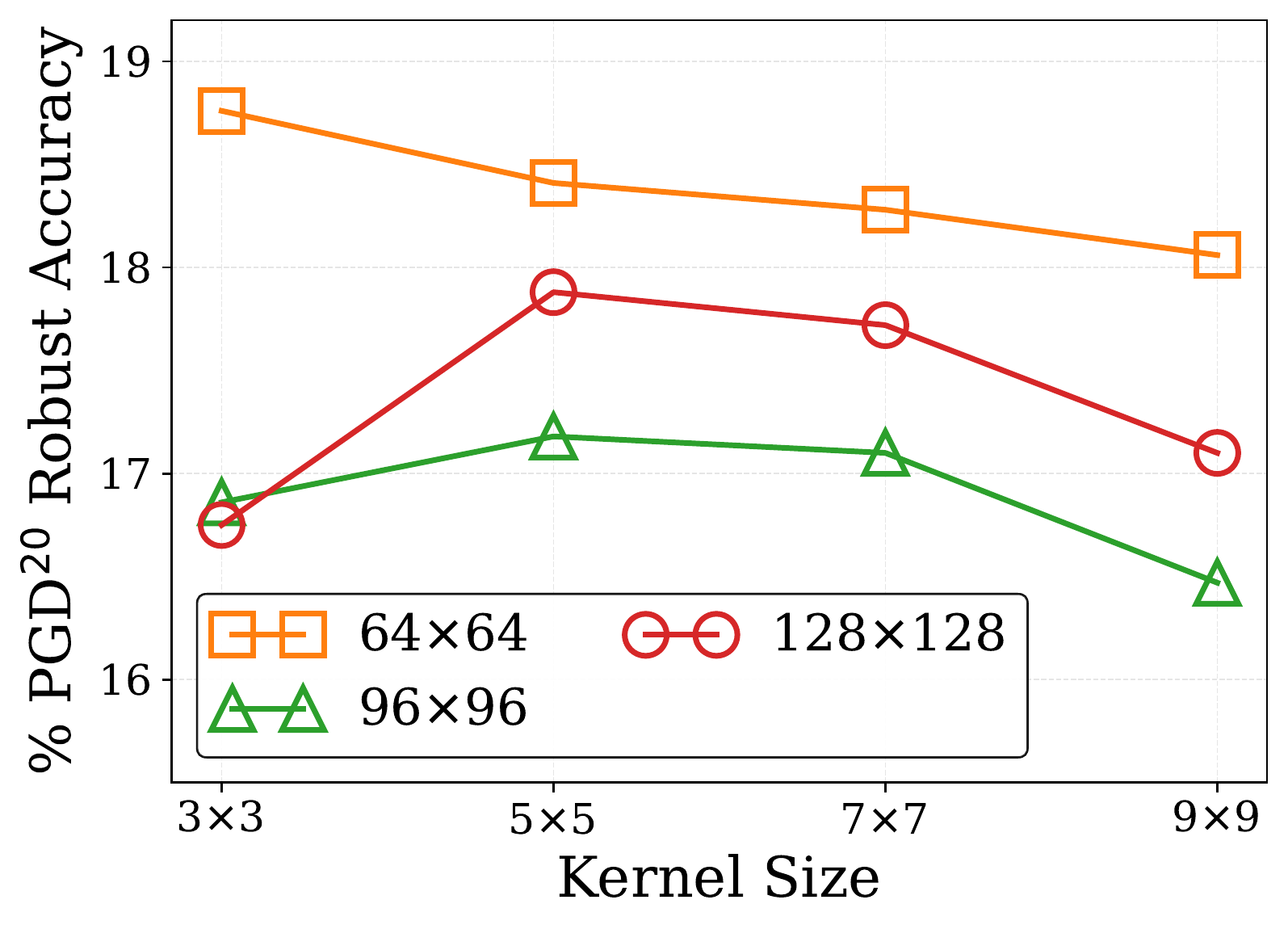}
    \caption{\scriptsize Tiny-ImageNet, PGD$^{20}$ \vspace{-8pt}}
    \end{subfigure}\hfill
    \begin{subfigure}[b]{0.235\textwidth}
    \centering
    \includegraphics[width=0.95\textwidth]{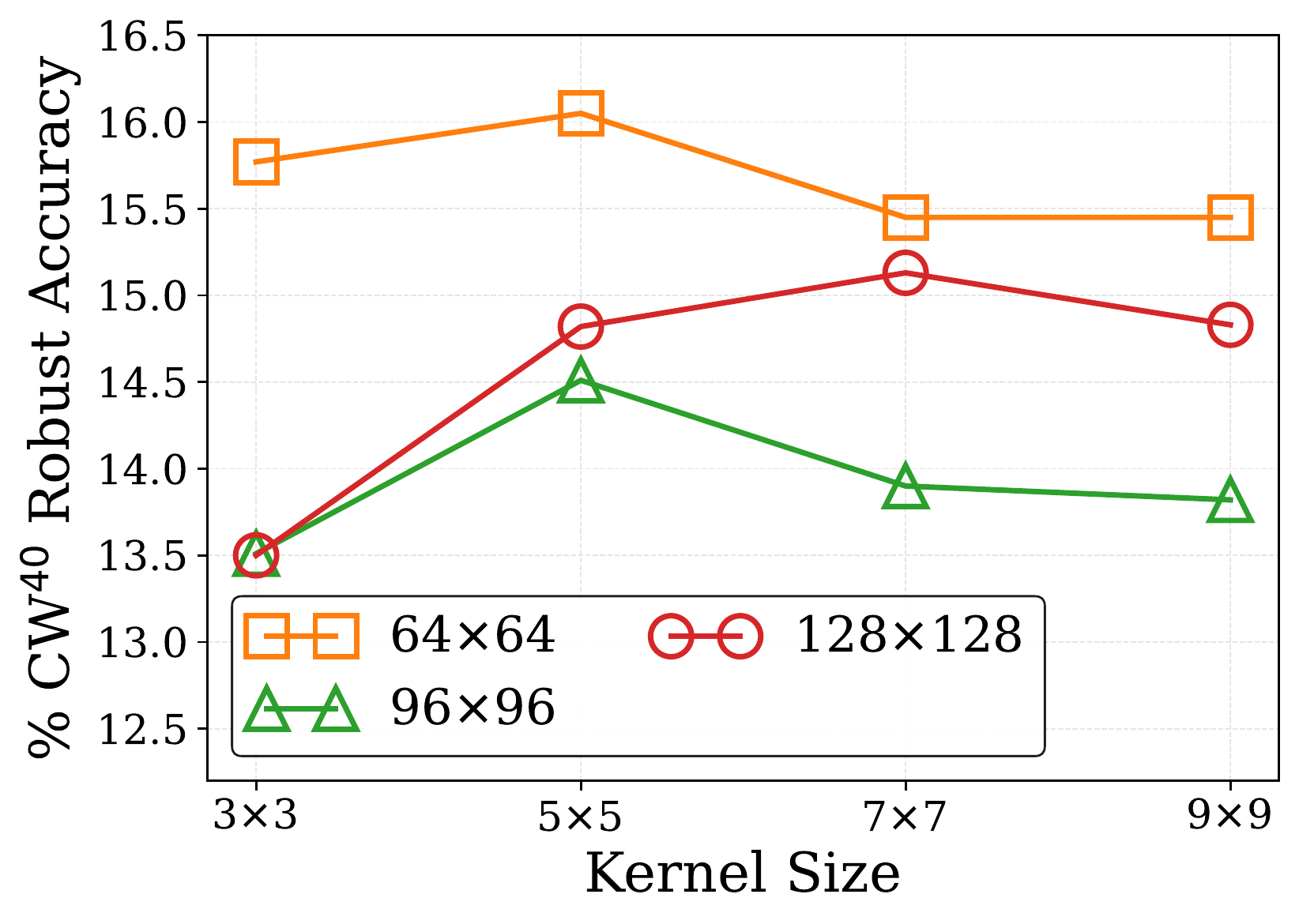}
    \caption{\scriptsize Tiny-ImageNet, CW$^{40}$ \vspace{-8pt}}
    \end{subfigure}\hfill
    \begin{subfigure}[b]{0.235\textwidth}
    \centering
    \includegraphics[width=0.95\textwidth]{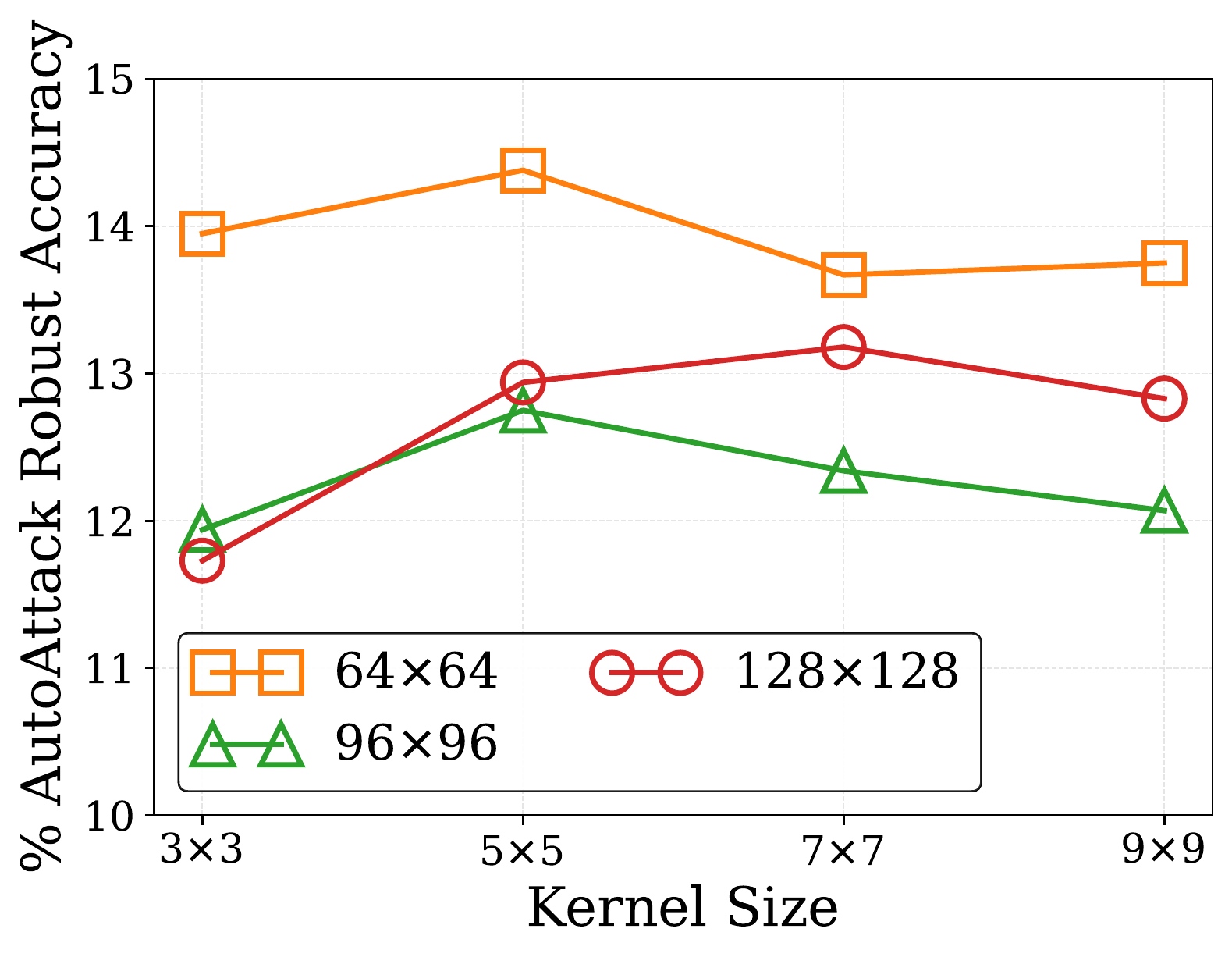}
    \caption{\scriptsize Tiny-ImageNet, AutoAttack \vspace{-8pt}}
    \end{subfigure}
    \caption{The adversarial robustness of different kernel sizes for higher resolution images on CIFAR-10 (Top) and Tiny-ImageNet (Bottom) against FGSM, PGD$^{20}$, CW$^{40}$, and AutoAttack (from left to right). \label{fig:abl_kernel_higher_resolution} \vspace{-1em}}
\end{figure*}
% -------------------------------------------------------------------------------------
% -------------------------------------------------------------------------------------
\begin{table*}[!ht]
\centering
\caption{The specifications of \ournetwork{}s. The stage wise setting is presented using $\begin{bmatrix}{k} \times k\mbox{, \#Ch}\end{bmatrix}$, where $k$ denotes the convolution filter size, \#Ch denotes the number of output channels, and $\begin{bmatrix} \cdot \end{bmatrix}$ indicates our \ourblock{} identified in \S\ref{sec:block}.\label{tab:network_spec} \vspace{-1em}}
\resizebox{.8\textwidth}{!}{%
\begin{tabular}{@{\hspace{2mm}}c|c|c|c|c|c@{\hspace{2mm}}}
\toprule
\multicolumn{1}{l|}{} & \multicolumn{1}{c|}{Output scale} & \multicolumn{1}{c|}{\ournetwork{}-A1} & \multicolumn{1}{c|}{\ournetwork{}-A2} & \multicolumn{1}{c|}{\ournetwork{}-A3} & \multicolumn{1}{c}{\ournetwork{}-A4} \\ \midrule
{Stem} & $32\times32$ & \multicolumn{4}{c}{3 $\times$ 3, 16, stride 1}  \\ \midrule
\multirow{1}{*}{Stage 1} & $32\times32$ & $\begin{bmatrix} \mbox{1} \times \mbox{1, 160}\\ \mbox{3} \times \mbox{3, 80}\\ \mbox{1} \times \mbox{1, 320}\end{bmatrix}$ $\times$ 14 & $\begin{bmatrix} \mbox{1} \times \mbox{1, 224}\\ \mbox{3} \times \mbox{3, 224}\\ \mbox{1} \times \mbox{1, 448}\end{bmatrix}$ $\times$ 17 & $\begin{bmatrix} \mbox{1} \times \mbox{1, 256}\\ \mbox{3} \times \mbox{3, 256}\\ \mbox{1} \times \mbox{1, 512}\end{bmatrix}$ $\times$ 22 & $\begin{bmatrix} \mbox{1} \times \mbox{1, 320}\\ \mbox{3} \times \mbox{3, 320}\\ \mbox{1} \times \mbox{1, 640}\end{bmatrix}$ $\times$ 27 \\ \midrule
\multirow{1}{*}{Stage 2} & $16\times16$ & $\begin{bmatrix} \mbox{1} \times \mbox{1, 448}\\ \mbox{3} \times \mbox{3, 448}\\ \mbox{1} \times \mbox{1, 896}\end{bmatrix}$ $\times$ 14 & $\begin{bmatrix} \mbox{1} \times \mbox{1, 576}\\ \mbox{3} \times \mbox{3, 576}\\ \mbox{1} \times \mbox{1, 1152}\end{bmatrix}$ $\times$ 17 & $\begin{bmatrix} \mbox{1} \times \mbox{1, 704}\\ \mbox{3} \times \mbox{3, 704}\\ \mbox{1} \times \mbox{1, 1408}\end{bmatrix}$ $\times$ 22 & $\begin{bmatrix} \mbox{1} \times \mbox{1, 896}\\ \mbox{3} \times \mbox{3, 896}\\ \mbox{1} \times \mbox{1, 1792}\end{bmatrix}$ $\times$ 28 \\ \midrule
\multirow{1}{*}{Stage 3} & $8\times8$ & $\begin{bmatrix} \mbox{1} \times \mbox{1, 384}\\ \mbox{3} \times \mbox{3, 384}\\ \mbox{1} \times \mbox{1, 768}\end{bmatrix}$ $\times$ 7 & $\begin{bmatrix} \mbox{1} \times \mbox{1, 512}\\ \mbox{3} \times \mbox{3, 512}\\ \mbox{1} \times \mbox{1, 1024}\end{bmatrix}$ $\times$ 8 & $\begin{bmatrix} \mbox{1} \times \mbox{1, 640}\\ \mbox{3} \times \mbox{3, 640}\\ \mbox{1} \times \mbox{1, 1280}\end{bmatrix}$ $\times$ 11 & $\begin{bmatrix} \mbox{1} \times \mbox{1, 768}\\ \mbox{3} \times \mbox{3, 768}\\ \mbox{1} \times \mbox{1, 1536}\end{bmatrix}$ $\times$ 13 \\ \midrule
\multirow{1}{*}{Tail} & $1\times1$ & \multicolumn{4}{c}{Global average pool} \\
\bottomrule
\end{tabular}%
}
\end{table*}
% -------------------------------------------------------------------------------------

\subsection{Impact of Convolution Kernel Size}~\label{sec:app_kernel_size}
Larger kernel sizes have been shown to be beneficial on standard problems \cite{tan2019mixconv,liu2022convnet,ding2022scaling} under standard ERM training. We evaluate large kernel sizes for adversarial robustness. Specifically, we allow the kernel size $K_{i\in\{1,2,3\}}$ for each stage to be among \{3$\times$3, 5$\times$5, 7$\times$7, 9$\times$9\} while using the default options for all other settings as described in \S\ref{sec:preliminaries}. We evaluate all the $4^3=64$ possible networks with all possible settings for the kernel size. Figure~\ref{fig:abl_kernel_resolution} shows our results. We observe that, in general, \emph{a larger kernel size does not necessarily lead to better adversarial robustness}. We repeat the experiment at higher image resolutions to verify if this observation is specific to low-resolution images. Specifically, we upsample the images to the following sizes: \{$64\times64$, $96\times96$, $128\times128$\}. We constrain all stages to use a canonical kernel size and use a stride of two in the first block of the first stage when the image resolution is higher than $64\times64$. Figure~\ref{fig:abl_kernel_higher_resolution} presents these results. Empirically, we observe that larger kernels start to improve adversarial robustness noticeably when the image size increases to $128\times128$, particularly on Tiny-ImageNet. However, adversarial robustness on upsampled images is consistently worse than that of smaller images. Thus, we argue that \emph{a kernel size of $3\times3$ remains the preferred choice for adversarial robustness}.

\subsection{Extended discussion of \ournetwork{}s}~\label{sec:app_robustresnet}
In this section, we provide detailed specifications of \ournetwork{}A1 - A4 in Table~\ref{tab:network_spec}.

\end{document}